\documentclass{article} 
\usepackage{iclr2022_conference,times}

\usepackage{amsmath,amsfonts,bm}

\def\eqref#1{equation~\ref{#1}}

\def\1{\bm{1}}

\DeclareMathAlphabet{\mathsfit}{\encodingdefault}{\sfdefault}{m}{sl}
\SetMathAlphabet{\mathsfit}{bold}{\encodingdefault}{\sfdefault}{bx}{n}

\usepackage{hyperref}
\usepackage{url}

\usepackage[pdftex]{graphicx}
\graphicspath{{../pdf/}{../jpeg/}}
\DeclareGraphicsExtensions{.pdf,.jpeg,.png}

\usepackage[ruled,vlined]{algorithm2e}
\usepackage{wrapfig}
\usepackage{multicol}
\usepackage{multirow}
\usepackage{booktabs}
 \usepackage{blindtext}
 \usepackage{floatrow}
 \usepackage{colortbl}
 \usepackage{arydshln}
 \usepackage{amsmath}
 \usepackage[toc,page,header]{appendix}
 \usepackage{minitoc}
 
 \newfloatcommand{capbtabbox}{table}[][\FBwidth]

\newcommand{\ts}{\textsuperscript}

\title{Accelerating Guided Diffusion Sampling with Splitting Numerical Methods}

\author{Antiquus S.~Hippocampus, Natalia Cerebro \& Amelie P. Amygdale \thanks{ Use footnote for providing further information
about author (webpage, alternative address)---\emph{not} for acknowledging
funding agencies.  Funding acknowledgements go at the end of the paper.} \\
Department of Computer Science\\
Cranberry-Lemon University\\
Pittsburgh, PA 15213, USA \\
\texttt{\{hippo,brain,jen\}@cs.cranberry-lemon.edu} \\
\And
Ji Q. Ren \& Yevgeny LeNet \\
Department of Computational Neuroscience \\
University of the Witwatersrand \\
Joburg, South Africa \\
\texttt{\{robot,net\}@wits.ac.za} \\
\AND
Coauthor \\
Affiliation \\
Address \\
\texttt{email}
}
\author{Suttisak Wizadwongsa, Supasorn Suwajanakorn
\\VISTEC, Thailand\\
{\texttt \{suttisak.w\_s19, supasorn.s\}@vistec.ac.th}
}

\iclrfinalcopy 
\begin{document}

\maketitle

\begin{abstract}

\emph{Guided diffusion} is a technique for conditioning the output of a diffusion model at sampling time without retraining the network for each specific task. One drawback of diffusion models, however, is their slow sampling process. Recent techniques can accelerate unguided sampling by applying high-order numerical methods to the sampling process when viewed as differential equations. On the contrary, we discover that the same techniques do not work for guided sampling, and little has been explored about its acceleration. This paper explores the culprit of this problem and provides a solution based on operator splitting methods, motivated by our key finding that classical high-order numerical methods are unsuitable for the conditional function. Our proposed method can re-utilize the high-order methods for guided sampling and can generate images with the same quality as a 250-step DDIM baseline using 32-58\% less sampling time on ImageNet256. 
We also demonstrate usage on a wide variety of conditional generation tasks, such as text-to-image generation, colorization, inpainting, and super-resolution.

\end{abstract}
\section{Submission of conference papers to ICLR 2022}

ICLR requires electronic submissions, processed by
\url{https://openreview.net/}. See ICLR's website for more instructions.

If your paper is ultimately accepted, the statement {\tt
  {\textbackslash}iclrfinalcopy} should be inserted to adjust the
format to the camera ready requirements.

The format for the submissions is a variant of the NeurIPS format.
Please read carefully the instructions below, and follow them
faithfully.

\subsection{Style}

Papers to be submitted to ICLR 2022 must be prepared according to the
instructions presented here.


Authors are required to use the ICLR \LaTeX{} style files obtainable at the
ICLR website. Please make sure you use the current files and
not previous versions. Tweaking the style files may be grounds for rejection.

\subsection{Retrieval of style files}

The style files for ICLR and other conference information are available online at:
\begin{center}
   \url{http://www.iclr.cc/}
\end{center}
The file \verb+iclr2022_conference.pdf+ contains these
instructions and illustrates the
various formatting requirements your ICLR paper must satisfy.
Submissions must be made using \LaTeX{} and the style files
\verb+iclr2022_conference.sty+ and \verb+iclr2022_conference.bst+ (to be used with \LaTeX{}2e). The file
\verb+iclr2022_conference.tex+ may be used as a ``shell'' for writing your paper. All you
have to do is replace the author, title, abstract, and text of the paper with
your own.

The formatting instructions contained in these style files are summarized in
sections \ref{gen_inst}, \ref{headings}, and \ref{others} below.

\section{General formatting instructions}
\label{gen_inst}

The text must be confined within a rectangle 5.5~inches (33~picas) wide and
9~inches (54~picas) long. The left margin is 1.5~inch (9~picas).
Use 10~point type with a vertical spacing of 11~points. Times New Roman is the
preferred typeface throughout. Paragraphs are separated by 1/2~line space,
with no indentation.

Paper title is 17~point, in small caps and left-aligned.
All pages should start at 1~inch (6~picas) from the top of the page.

Authors' names are
set in boldface, and each name is placed above its corresponding
address. The lead author's name is to be listed first, and
the co-authors' names are set to follow. Authors sharing the
same address can be on the same line.

Please pay special attention to the instructions in section \ref{others}
regarding figures, tables, acknowledgments, and references.

There will be a strict upper limit of 9 pages for the main text of the initial submission, with unlimited additional pages for citations. 

\section{Headings: first level}
\label{headings}

First level headings are in small caps,
flush left and in point size 12. One line space before the first level
heading and 1/2~line space after the first level heading.

\subsection{Headings: second level}

Second level headings are in small caps,
flush left and in point size 10. One line space before the second level
heading and 1/2~line space after the second level heading.

\subsubsection{Headings: third level}

Third level headings are in small caps,
flush left and in point size 10. One line space before the third level
heading and 1/2~line space after the third level heading.
\section{Background}
This section provides a high-level summary of the theoretical foundation of diffusion models as well as numerical methods that have been used for diffusion models.
Here we briefly explain a few that contribute to our method.

\subsection{Diffusion Models}	
Assuming that $x_0$ is a random variable from the data distribution we wish to reproduce,
diffusion models define a sequence of Gaussian noise degradation of $x_0$ as random variables $x_1, x_2,...,x_T$, where $x_{t} \sim \mathcal{N}(\sqrt{1-\beta_t} x_{t-1}, \beta_t \mathbf{I})$ and $\beta_t\in[0,1]$ are parameters that control the noise levels.
With a property of Gaussian distribution, we can express $x_t$ directly as a function of $x_0$ and noise  $\epsilon \sim \mathcal{N}(0,\mathbf{I})$ by $x_t=\sqrt{\bar{\alpha}_t}x_0 + \sqrt{1-\bar{\alpha}_t}\epsilon$, where $\bar{\alpha}_t=\prod^t_{i=1}(1-\beta_i)$.
By picking a sufficiently large $T$ (e.g., 1,000) and an appropriate set of $\beta_t$, we can assume $x_T$ is a standard Gaussian distribution.
The main idea of diffusion model generation is to sample a Gaussian noise $x_T$ and use it to reversely sample $x_{T-1}$, $x_{T-2},...$ until we obtain $x_0$, which belongs to our data distribution.

\cite{ho2020denoising} propose Denoising Diffusion Probabilistic Model (DDPM) and explain how to employ a neural network $\epsilon_\theta(x_t, t)$ to predict the noise $\epsilon$ that is used to compute $x_t$.
To train the network, we sample a training image $x_0$, $t$, and $\epsilon$ to compute $x_t$ using the above relationship.
Then, we optimize our network $\epsilon_\theta$ to minimize the difference between the predicted and real noise,
 i.e., $\|\epsilon - \epsilon_\theta (x_t,t)\|^2$.
 
\cite{song2020denoising} introduce Denoising Diffusion Implicit Model (DDIM), which uses the network $\epsilon_\theta$ to deterministically obtain $x_{t-1}$ given $x_t$. The DDIM generative process can be written as
 \begin{equation} \label{ddim}
     x_{t-1} = \sqrt{\frac{\bar{\alpha}_{t-1}}{\bar{\alpha}_{t}} }
     \left( x_t - \sqrt{1-\bar{\alpha}_{t}} \epsilon_\theta (x_t,t)\right)
     +\sqrt{1-\bar{\alpha}_{t-1}} \epsilon_\theta (x_t,t).
 \end{equation}
 
This formulation could be used to skip many sampling steps and boost sampling speed.
To turn this into an ODE, we rewrite Equation \ref{ddim} as:
\begin{equation}
    \frac{x_{t-\Delta t}}{\sqrt{\bar{\alpha}_{t-\Delta t}}} = \frac{x_{t}}{\sqrt{\bar{\alpha}_{t}}} 
    + \left(\sqrt{\frac{1-\bar{\alpha}_{t-\Delta t}}{\bar{\alpha}_{t-\Delta t}}} - \sqrt{\frac{1-\bar{\alpha}_{t}}{\bar{\alpha}_{t}}} \right) \epsilon_\theta (x_t,t),
\end{equation}
which is now equivalent to a numerical step in solving an ODE.
To derive the corresponding ODE, we can re-parameterize 
$\sigma_t = \sqrt{1-\bar{\alpha}_{t}}/ \sqrt{\bar{\alpha}_{t}}, \: \bar{x}(t) = x_t/\sqrt{\bar{\alpha}_{t}}$ and $\bar{\epsilon}_\sigma(\bar{x}) = \epsilon_\theta (x_t,t)$, yielding $\bar{x}({t-\Delta t}) - \bar{x}({t}) = (\sigma_{t-\Delta t} - \sigma_t) \bar{\epsilon}_\sigma (\bar{x})$.
By letting $(\sigma_{t-\Delta t} - \sigma_t) \rightarrow 0$, the ODE becomes:

\begin{equation} \label{eq:ode}
    \frac{d\bar{x}}{d\sigma} = \bar{\epsilon}_\sigma (\bar{x}).
\end{equation}

Note that this change of variables is equivalent to an exponential integrator technique described in both \cite{zhang2022fast} and \cite{lu2022dpm}.
Since $x_t$ and $\bar{x}(t)$ have the same value at $t=0$, our work can focus on solving $\bar{x}(t)$ rather than $x_t$.
Many numerical methods can be applied to the ODE Equation \ref{eq:ode} to accelerate diffusion sampling.
We next discuss some of them that are relevant.

\subsection{Numerical Methods}	
\textbf{Euler's Method} is  
the most basic numerical method. 
A forward Euler step is given by
$ \bar{x}_{n+1} = \bar{x}_n +\Delta \sigma \bar{\epsilon}_\sigma (\bar{x}_n)$. 
When we apply the forward Euler step to the ODE Equation \ref{eq:ode}, we get the DDIM formulation \citep{song2020denoising}.


\textbf{Heun’s Method}, also known as the trapezoid rule or improved Euler, is given by:
$ \bar{x}_{n+1} = \bar{x}_n +\frac{\Delta \sigma}{2} (e_1 + e_2)$, where $e_1 = \bar{\epsilon}_\sigma (\bar{x}_{n})$ and $ e_2 = \bar{\epsilon}_\sigma (\bar{x}_{n} + \Delta \sigma e_1).$
This method modifies Euler's method into a two-step method to improve accuracy.
Many papers have used this method on diffusion models, including Algorithm 1 in \cite{karras2022elucidating} and DPM-Solver-2 in \cite{lu2022dpm}.
This method is also the simplest case of Predictor-Corrector methods used in \cite{song2020score}.

\textbf{Runge-Kutta Methods} represent a class of numerical methods that integrate information from multiple hidden steps and provide high accuracy results. Heun's method also belongs to a family of 2\ts{nd}-order Runge-Kutta methods (RK2).
The most well-known 
variant
is the 4\ts{th}-order Runge-Kutta method (RK4), which is written as follows:
$$ e_1 = \bar{\epsilon}_\sigma (\bar{x}_{n}), \quad e_2 = \bar{\epsilon}_\sigma \left(\bar{x}_{n} + \frac{\Delta \sigma}{2} e_1\right),
\quad e_3 = \bar{\epsilon}_\sigma \left(\bar{x}_{n} + \frac{\Delta \sigma}{2} e_2\right), \quad e_4 = \bar{\epsilon}_\sigma \left(\bar{x}_{n} + \Delta \sigma e_3\right),$$
\begin{equation}
 \bar{x}_{n+1} = \bar{x}_n +\frac{\Delta \sigma}{6} (e_1 + 2 e_2 + 2 e_3 + e_4). 
\end{equation}
This method has been tested on diffusion models in \cite{liu2022pseudo} and \cite{salimans2022progressive}, but it has not been used as the main proposed method in any paper.

\textbf{Linear Multi-Step Method}, 
similar to the Runge-Kutta methods, aims to combine information from several steps; however, rather than evaluating new hidden steps, this method uses the previous steps to estimate the new step. The 1\ts{st}-order formulation is the same as Euler's method. 
The 2\ts{nd}-order formulation is given by 
\begin{equation}
    \bar{x}_{n+1} = \bar{x}_n +\frac{\Delta \sigma}{2} \left(3 e_0 - e_1\right),
\end{equation}
while the 4\ts{th}-order formulation is given by 
\begin{equation}
    \bar{x}_{n+1} = \bar{x}_n +\frac{\Delta \sigma}{24} (55 e_0 - 59e_1 + 37 e_2 - 9 e_3),
\end{equation}
where $e_k = \bar{\epsilon}_\sigma (\bar{x}_{n-k})$.
These formulations are designed for a constant $\Delta \sigma$ in each step. However, our experiments and previous work that uses this method (e.g., \cite{liu2022pseudo, zhang2022fast}) still show good results when this assumption is not strictly satisfied, i.e., when $\Delta \sigma$ is not constant.
We will refer to these formulations as PLMS (Pseudo Linear Multi-Step) for the rest of the paper, like in \cite{liu2022pseudo}.
A similar linear multi-step method for non-constant $\Delta \sigma$ can also be derived using a technique used in  \cite{zhang2022fast}, which we detail in Appendix  \ref{iPLMS}. The method can improve upon PLMS slightly, but it is not as flexible because we have to re-derive the update rule every time the $\sigma$ schedule changes.

\section{Splitting Methods for Guided Diffusion Models}
This section introduces our technique that uses splitting numerical methods to accelerate guided diffusion sampling. We first focus our investigation on \emph{classifier-guided} diffusion models for class-conditional generation and later demonstrate how this technique can be used for other conditional generation tasks in Section \ref{sec:splitting_for_other_tasks}. Like any guided diffusion models, classifier-guided models \citep{dhariwal2021diffusion} share the same training objective with regular unguided models with no modifications to the training procedure; but the sampling process is guided by an additional gradient signal from an external classifier to generate class-specific output images. Specifically, the sampling process is given by
\begin{equation}
    \hat{\epsilon} = \epsilon_\theta (x_t) -\sqrt{1-\bar{\alpha}_t}  \nabla_x \log p_\phi (c | x_t),  \quad
x_{t-1} = \sqrt{\bar{\alpha}_{t-1}}
     \left( \frac{x_t - \sqrt{1-\bar{\alpha}_{t}} \hat{\epsilon}}{\sqrt{\bar{\alpha}_{t}}}\right)
     +\sqrt{1-\bar{\alpha}_{t-1}} \hat{\epsilon},
\end{equation}
where $p_\phi (c | x_t)$ is a classifier model trained to output the probability of $x_t$ belonging to class $c$.
As discussed in the previous section, we can rewrite this formulation as a ``guided ODE'':
\begin{equation} \label{eq:guide}
    \frac{ d \bar{x}}{d \sigma} = \bar{\epsilon}_\sigma (\bar{x}) - \nabla f_\sigma (\bar{x}),
\end{equation}

where $f_\sigma (\bar{x}) = \frac{\sigma}{\sqrt{\sigma^2+1}} \log p_\phi (c | x_t)$. We refer to $f_\sigma$ as the conditional function, which can be substituted with other functions for different tasks. After obtaining the ODE form, any numerical solver mentioned earlier can be readily applied to accelerate the sampling process.
However, we observe that classical high-order numerical methods (e.g., PLMS4, RK4) fail to accelerate this task (see Figure \ref{fig:class}) and even perform worse than the baseline DDIM.

We hypothesize that the two terms in the guided ODE may have different numerical behaviors with the conditional term being less suitable to classical high-order methods. We speculate that the difference could be partly attributed to how they are computed: $\nabla f_\sigma (\bar{x})$ is computed through backpropagation, whereas $\bar{\epsilon}_\sigma (\bar{x})$ is computed directly by evaluating a network. 
One possible solution to handle terms with different behaviors is the so-called operator splitting method, which divides the problem into two subproblems:
\begin{equation} \frac{dy}{d\sigma} = \bar{\epsilon}_\sigma (y), \quad \frac{dz}{d\sigma} = - \nabla f_\sigma (z). \end{equation}
We call these the \emph{diffusion} and \emph{condition} subproblems, respectively.
This method allows separating the hard-to-approximate $\nabla f_\sigma (z)$ from $\bar{\epsilon}_\sigma (y)$ and solving them separately in each time step. Importantly, this helps reintroduce the effective use of high-order methods on the diffusion subproblem as well as provides us with options to combine different specialized methods to maximize performance. 
We explore two most famous first- and second-order splitting techniques for our task:

\subsection{Lie-Trotter Splitting (LTSP)}
Our first example is the simple first-order Lie-Trotter splitting method \citep{trotter1959product}, which expresses the splitting as
\begin{align}
    \frac{dy}{d\sigma} &= \bar{\epsilon}_\sigma (y),&y(\sigma_n) &= \bar{x}_n,       & \sigma \in [\sigma_{n+1}, \sigma_{n}] \label{eq:ltts1}\\ 
    \frac{dz}{d\sigma} &= - \nabla f_\sigma (z),    &z(\sigma_n) &= y(\sigma_{n+1}), & \sigma \in [\sigma_{n+1}, \sigma_{n}] \label{eq:ltts2} 
\end{align}

with the solution of this step being $\bar{x}_{n+1} = z(\sigma_{n+1})$. 
Note that $\sigma_n$ is a decreasing sequence in sampling schedule. 
Here Equation \ref{eq:ltts1} is the same as Equation \ref{eq:ode}, which can be solved using any high-order numerical method, e.g., PLMS.
For Equation \ref{eq:ltts2}, we can use a forward Euler step:
\begin{equation}z_{n+1}=z_n - \Delta \sigma \nabla f_\sigma (z_n).\end{equation}
This is equivalent to a single iteration of standard gradient descent with a learning rate $\Delta \sigma$.
This splitting scheme is summarized by Algorithm \ref{algo:ltsp}.
We investigate different numerical methods for each subproblem in Section \ref{exp1}.

\begin{table}
\begin{minipage}{0.49\textwidth}
  \begin{algorithm}[H]
    \label{algo:ltsp}
    \SetAlgoLined
      sample $\bar{x}_0 \sim \mathcal{N}(0,\sigma^2_{\text{max}}\mathbf{I})$ \;
     \For{$n\in\{0,...,N-1\}$}{
         $y_{n+1} = \text{PLMS}(\bar{x}_n,\sigma_n,\sigma_{n+1},\bar{\epsilon}_\sigma) $\;
         $\bar{x}_{n+1} = y_{n+1} - (\sigma_{n+1}-\sigma_n) \nabla f(y_{n+1})$ \;
        }
       \KwResult{$\bar{x}_N$} 
     \caption{Lie-Trotter Splitting (LTSP)}
  \end{algorithm}
\end{minipage}
\begin{minipage}{0.47\textwidth}

\begin{algorithm}[H]
\label{algo:stsp}
\SetAlgoLined
 sample $\bar{x}_0 \sim \mathcal{N}(0,\sigma^2_{\text{max}}\mathbf{I})$ \;
 \For{$n\in\{0,...,N-1\}$}{
    $z_{n+1} = \bar{x}_n - \frac{(\sigma_{n+1}-\sigma_n)}{2} \nabla f(\bar{x}_n)$ \;
    $y_{n+1} = \text{PLMS}(z_{n+1},\sigma_n,\sigma_{n+1},\bar{\epsilon}_\sigma) $\;
    $\bar{x}_{n+1} = y_{n+1} - \frac{(\sigma_{n+1}-\sigma_n)}{2} \nabla f(y_{n+1})$ \;
    }
   \KwResult{$\bar{x}_N$} 
 \caption{Strang Splitting (STSP)}
\end{algorithm}

\end{minipage}
\end{table}

%


\subsection{Strang Splitting (STSP)}
Strang splitting (or Strang-Marchuk)
\citep{strang1968construction} is one of the most famous and widely used operator splitting methods. 
This second-order splitting works as follows:
\begin{align}
    \frac{dz}{d\sigma} &= - \nabla f_\sigma (z),    &z(\sigma_n) = \bar{x}_n,& \quad& \sigma \in \left[ \frac{1}{2}(\sigma_{n}+\sigma_{n+1}), \sigma_n\right] \label{eq:bcpf1} \\
    \frac{dy}{d\sigma} &= \bar{\epsilon}_\sigma (y),&y(\sigma_n) = z\left(\frac{1}{2}(\sigma_{n}+\sigma_{n+1})\right),&       \quad& \sigma \in [\sigma_{n+1}, \sigma_n] \label{eq:bcpf2}\\
    \frac{d \Tilde{z}}{d\sigma} &= - \nabla f_\sigma (\Tilde{z}),    &\Tilde{z}\left(\frac{1}{2}(\sigma_{n}+\sigma_{n+1})\right) = y(\sigma_{n+1}),& \:& \sigma \in \left[ \sigma_{n+1}, \frac{1}{2}(\sigma_{n}+\sigma_{n+1})\right] \label{eq:bcpf3}
\end{align}
Instead of solving each subproblem for a full step length, we solve the condition subproblem for half a step before and after solving the diffusion subproblem for a full step. In theory, we can swap the order of operations without affecting convergence, but it is practically cheaper to compute the condition term twice rather than the diffusion term twice because $f_\sigma$ is typically a smaller network compared to $\bar{\epsilon}_\sigma$. 
The Strange splitting algorithm is shown in Algorithm \ref{algo:stsp}.
This method can be proved to have better accuracy than the Lie-Trotter method using the Banker-Campbell-Hausdorff formula \citep{mark2001stat}, but it requires evaluating the condition term twice per step in exchange for improved image quality. We assess this trade-off in the experiment section.
\section{Experiments}
Extending on our observation that classical high-order methods failed on guided sampling, we conducted a series of experiments to investigate this problem and evaluate our solution.
Section \ref{exp1} uses a simple splitting method (first-order LTSP) to study the effects that high-order methods have on each subproblem, leading to our key finding that \emph{only} the conditional subproblem is less suited to classical high-order methods.
This section also determines the best combination of numerical methods for the two subproblems under LTSP splitting.
Section \ref{exp2} explores improvements from using a higher-order splitting method and compares our best scheme to previous work.
Finally, Section \ref{sec:splitting_for_other_tasks} applies our approach to a variety of conditional generation tasks with minimal changes. 


For our comparison, we use pre-trained state-of-the-art diffusion models and classifiers from  \cite{dhariwal2021diffusion}, which were trained on the ImageNet dataset \citep{russakovsky2015imagenet} with 1000 total sampling step.  We treat full-path samples from a classifier-guided DDIM at 1,000 steps as reference solutions. Then the performance of each configuration is measured by the image similarity between its generated samples using fewer steps and the reference DDIM samples, both starting from the same initial noise maps. Given the same sampling time, we expect configurations with better performance to better match the full DDIM.
We measure image similarity using Learned Perceptual Image Patch Similarity (LPIPS) \citep{zhang2018unreasonable} (lower is better) and measure sampling time using a single NVIDIA RTX 3090 and a 24-core AMD Threadripper 3960x.




\begin{figure} \vspace{-4mm}
    \centering
    \shortstack{ \includegraphics[width = 0.49\textwidth]{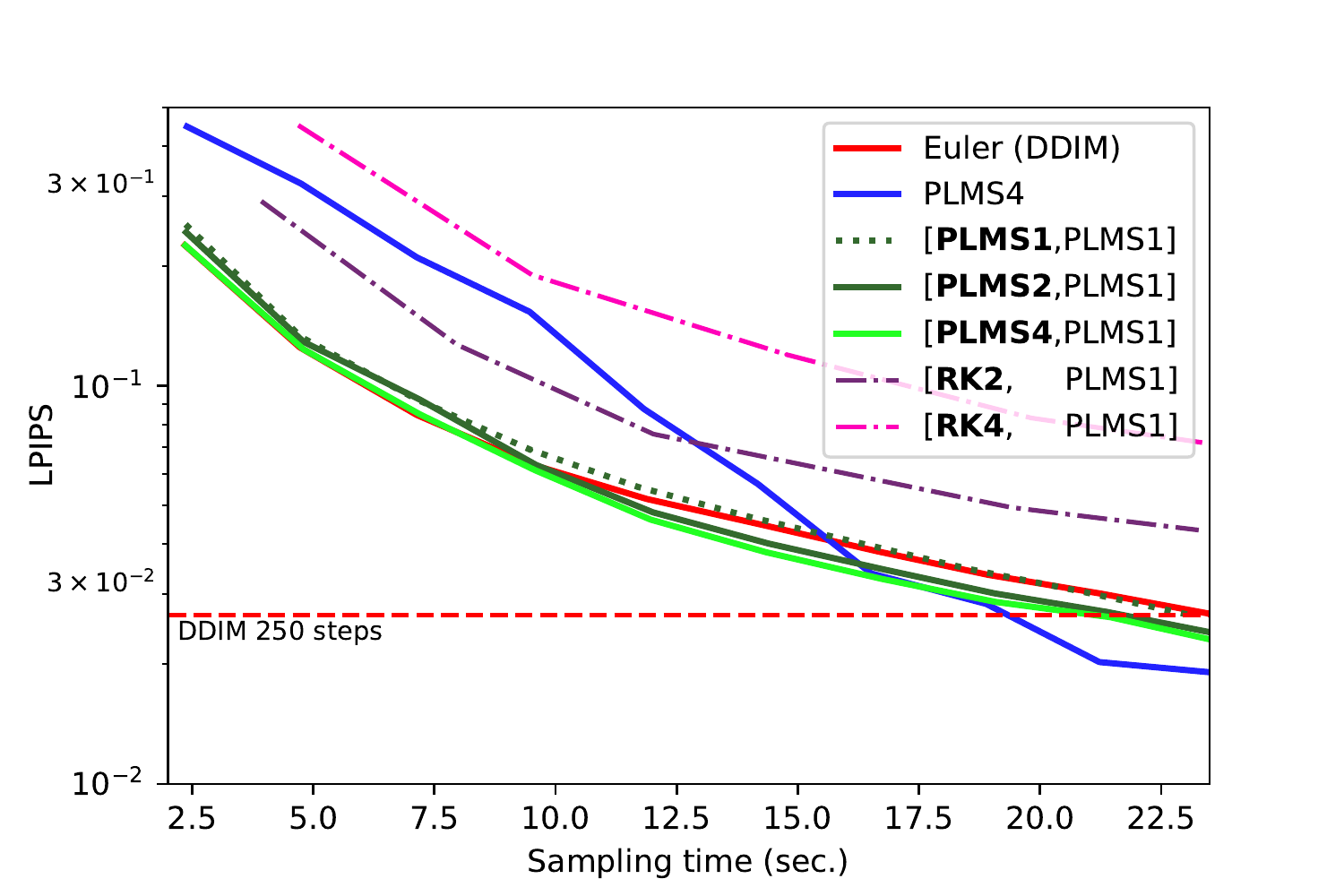}  
    \\ \footnotesize (a) Varying the method for the diffusion subproblem} 
    \shortstack{ \includegraphics[width = 0.49\textwidth]{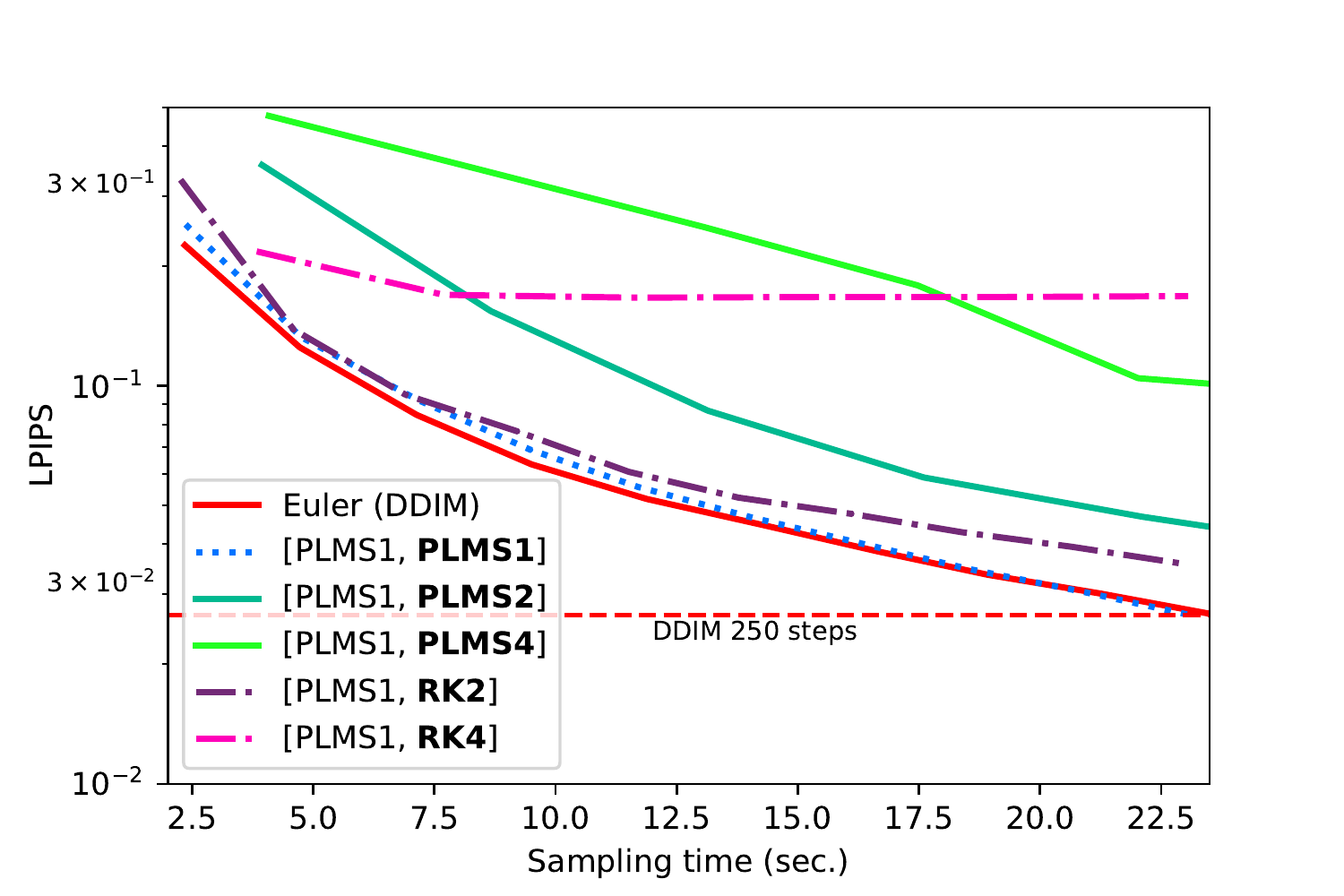} 
    \\ \footnotesize  (b) Varying the method for the condition subproblem}
    \caption{Comparison of different combinations of numerical methods under LTSP splitting for guided diffusion sampling.
    We plot LPIPS against the sampling time. [A, B] denotes the use of method A in the diffusion subproblem and method B in the condition subproblem. The red dotted lines indicate a reference DDIM score obtained from 250 sampling steps, which produce images visually close to those from 1,000 steps. 
    }
    \label{fig:exp1}
\end{figure} 

\subsection{Finding a suitable numerical method for each subproblem} \label{exp1}

To study the effects of different numerical methods on each subproblem of the guided ODE (Equation \ref{eq:guide}), we use the simplest Lie-Trotter splitting, which itself requires no additional network evaluations. This controlled experiment has two setups: a) we fix the numerical method for the condition subproblem (Equation  \ref{eq:ltts2}) to first-order PLMS1 (Euler's method) and vary the numerical method for the diffusion subproblem (Equation  \ref{eq:ltts1}), and conversely b) we fix the method for the diffusion subproblem and vary the method for the condition subproblem. The numerical method options are Euler's method (PLMS1), Heun's method (RK2), 4\ts{th} order Runge-Kutta's method (RK4), and 2\ts{nd}/4\ts{th} order pseudo linear multi-step (PLMS2/PLMS4). 
We report LPIPS vs. sampling time of various numerical combinations on a diffusion model trained on ImageNet 256$\times$256 in Figure \ref{fig:exp1}. The red dotted lines indicate a reference DDIM score obtained from 250 sampling steps, a common choice that produces good samples that are perceptually close to those from a full 1,000-step DDIM \citep{dhariwal2021diffusion, nichol2021improved}.

Given a long sampling time, non-split PLMS4 performs better than the DDIM baseline.
However, when the sampling time is reduced, the image quality of PLMS4 rapidly decreases and becomes much worse than that of DDIM, especially under 15 seconds in Figure \ref{fig:exp1}.
When we split the ODE and solve both subproblems using first-order PLMS1 (Euler), the performance is close to that of DDIM, which is also considered first-order but without any splitting. This helps verify that merely splitting the ODE does not significantly alter the sampling speed.

In the setup a), when RK2 and RK4 are used for the diffusion subproblem, they also perform worse than the DDIM baseline. This slowdown is caused by the additional evaluations of the network by these methods, which outweigh the improvement gained in each longer diffusion step. Note that if we instead measure the image quality with respect to the number of diffusion steps, RK2 and RK4 can outperform other methods (Appendix \ref{step}); however, this is not our metric of interest. On the other hand, PLMS2 and PLMS4, which require no additional network evaluations, are about 8-10\% faster than DDIM and can achieve the same LPIPS score as the DDIM that uses 250 sampling steps in 20-26 fewer steps.
Importantly, when the sampling time is reduced, their performance does not degrade rapidly like the non-split PLMS4 and remains at the same level as DDIM.


In the setup b) where we vary the numerical method for the condition subproblem, the result reveals an interesting contrast---none of the methods beats DDIM and some even make the sampling diverged [PLMS1, RK4].
These findings suggest that the gradients of conditional functions are less ``compatible'' with classical high-order methods, especially when used with a small number of steps.
This phenomenon may be related to the ``stiffness'' condition of ODEs, which we discuss further in Section \ref{sec:discussion}.
For the remainder of our experiments, we will use the combination [PLMS4, PLMS1] for the diffusion and condition subproblems, respectively.

\subsection{Improved splitting method} \label{exp2}
\begin{figure}\vspace{-4mm}
    \begin{floatrow}
    \ffigbox[.45\textwidth]{%
      \shortstack{ \includegraphics[width = 0.48\textwidth]{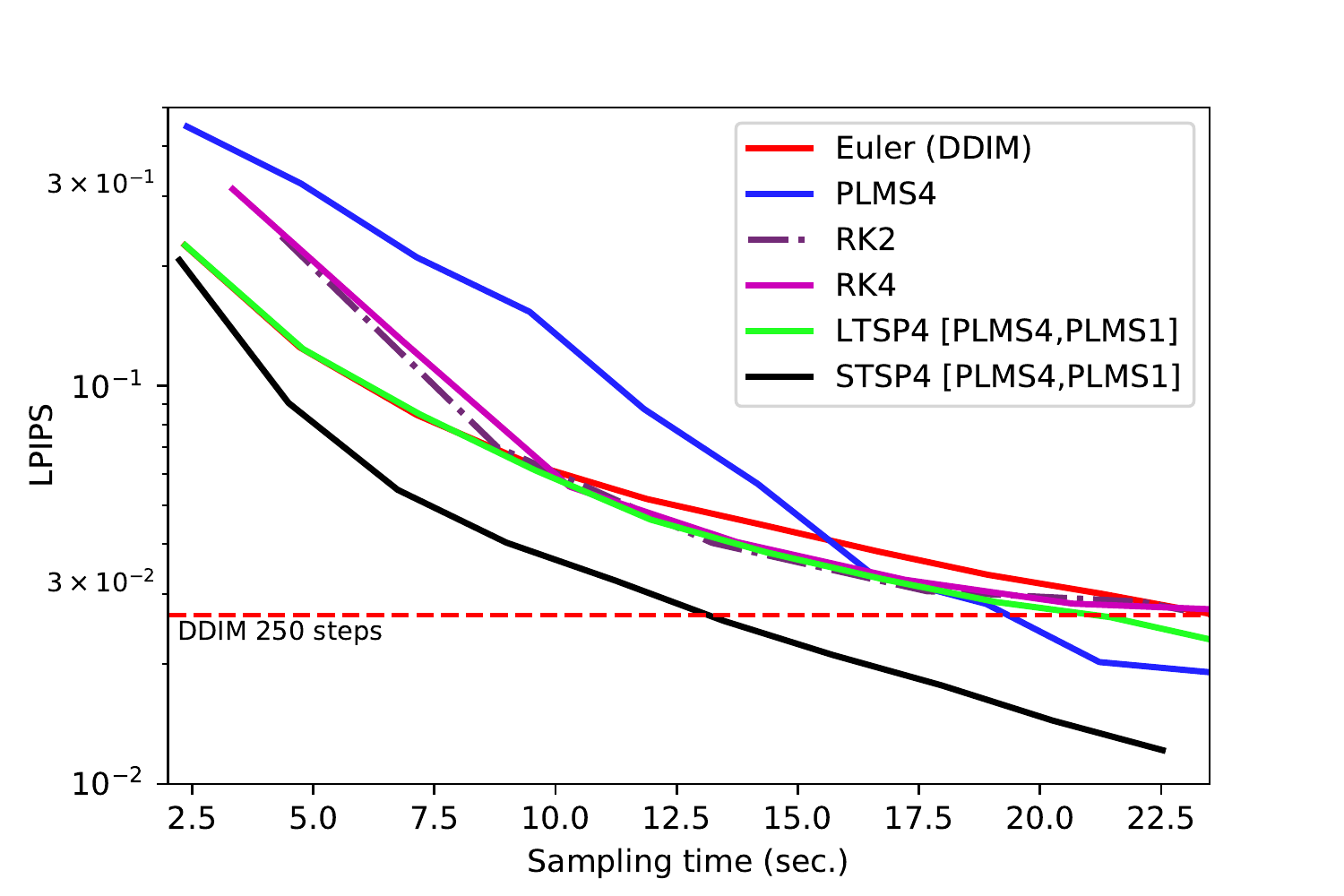}}%
    }{%
      \caption{Comparison of different numerical methods for guided diffusion sampling. }%
      \label{fig:exp2}
    }
    \capbtabbox{%
        \begin{tabular}{lcccc}
       \toprule
         &\multicolumn{4}{c}{Sampling time within} \\
         & 5 sec.  & 10 sec.  & 15 sec. & 20 sec. \\ 
       \midrule[0.08em]
       DDIM           & 0.116 & 0.062 & 0.043 & 0.033 \\
       PLMS4          & 0.278 & 0.141 & 0.057 & 0.026 \\
       RK2            & 0.193 & 0.059 & 0.036 & 0.028 \\
       RK4            & 0.216 & 0.054 & 0.039 & 0.028 \\
       \textbf{LTSP4} & 0.121 & 0.058 & 0.037 & 0.028 \\
       \textbf{STSP4} & \textbf{0.079} & \textbf{0.035} & \textbf{0.022} & \textbf{0.013} \\
       \bottomrule
       \end{tabular}
      \caption{
      Average LPIPS when the sampling time is limited to be under 5 - 20 seconds. 
      }%
      \label{tab:exp2}
    }{%
    }
    \end{floatrow}
\end{figure}

This experiment investigates improvements from using a high-order \emph{splitting} method, specifically the Strang splitting method, with the numerical combination [PLMS4, PLMS1] and compares our methods to previous work.
Note that besides DDIM \cite{dhariwal2021diffusion}, no previous work is specifically designed for accelerating \emph{guided} sampling, thus the baselines in this comparison are only adaptations of the core numerical methods used in those papers. And to our knowledge, no prior guided-diffusion work uses splitting numerical methods.
%
Non-split numerical method baselines are PLMS4, which is used in  \cite{liu2022pseudo}, RK2, which is used in \cite{karras2022elucidating, lu2022dpm}, and higher-order RK4. 
We report the LPIPS scores of these methods with respect to the sampling time in Figure \ref{fig:exp2} and Table \ref{tab:exp2}.



Without any splitting, PLMS4, RK2 and RK4 show significantly poorer image quality when used with short sampling times $<10$ seconds. The best performer is our Strang splitting (STSP4), which can reach the same quality as 250-step DDIM while using 32-58\% less sampling time. STSP4 also obtains the highest LPIPS scores for sample times of 5, 10, 15, and 20 seconds. 
More statistical details and comparison with other split combinations are in Appendix \ref{std}, \ref{str_aba}.

In addition, we perform a quantitative evaluation for class-conditional generation by sampling 50,000 images based on uniformly chosen class conditions with a small number of sampling steps and evaluating the Fenchel Inception Distance (FID) \cite{heusel2017gans} (lower is better) and the improved precision/recall \cite{kynkaanniemi2019improved} (higher is better) against an ImageNet test set.
Following \citep{dhariwal2021diffusion}, we use a 25-step DDIM as a baseline, which already produces visually reasonable results. 
As PLMS and LTSP require the same number of network evaluations as the DDIM, they are used also with 25 steps. For STSP with a longer network evaluation time, it is only allowed 20 steps, which is the highest number of steps such that its sampling time is within that of the baseline 25-step DDIM.
Here LTSP2 and STSP2 are Lie-Trotter and Strang splitting methods with the combination [PLMS2, PLMS1]. 
In Table \ref{tab:Sampling}, we report the results of three different ImageNet resolutions and the average sampling time per image in seconds.

Our STSP4 performs best on all measurements except Recall on ImageNet512.
On ImageNet512, PLMS4 has the highest Recall score but a poor FID of 16, indicating that the generated images have good distribution coverage but may poorly represent the real distribution.
On ImageNet256, STSP4 can yield 4.49 FID in 20 steps, compared to 4.59 FID in 250 steps originally reported in the paper \citep{dhariwal2021diffusion}; our STSP4 is about 9.4$\times$ faster when tested on the same machine.

\begin{table}
  \begin{minipage}{0.5\columnwidth}
    \centering
    \resizebox{0.98\columnwidth}{!}{
    \begin{tabular}{lccccc}
        \toprule
        Method & Steps & Time  & FID & Prec & Rec \\
        \midrule
        \multicolumn{4}{l}{\textbf{ImageNet128}} \\
        \arrayrulecolor[rgb]{0.7, 0.7, 0.7}\hline
       \rule{0pt}{2ex}DDIM & 25 & 0.55 & 6.69  & 0.78    & 0.49 \\
       PLMS2               & 25 & 0.57 & 5.71  & 0.80    & 0.51 \\
       PLMS4               & 25 & 0.57  & 4.97  & 0.80    & 0.53 \\
       \textbf{LTSP2}      & 25 & 0.55  & 5.14  & 0.81    & 0.51 \\
       \textbf{LTSP4}      & 25 & 0.55  & 3.85  &\textbf{0.81} & \textbf{0.54} \\
       \textbf{STSP2}      & 20 & 0.54  & 5.33  & 0.80    & 0.52 \\
       \textbf{STSP4}      & 20 & 0.54  &\textbf{3.78}& \textbf{0.81}& \textbf{0.54} \\
       [-1.2em]\\
       \hdashline[1.2pt/1.2pt]\arrayrulecolor{black}
       \rule{0pt}{2ex}\textit{ADM-G}  & \textit{250} & \textit{5.59*} & \textit{2.97} & \textit{0.78}  & \textit{0.59} \\
       \bottomrule
    \end{tabular}}
  \end{minipage}\hfill 
  \begin{minipage}{0.5\columnwidth}
    \centering
    \resizebox{0.98\columnwidth}{!}{
    \begin{tabular}{lccccc}
        \toprule
        Method & Steps & Time & FID & Prec & Rec \\
        \midrule
        \multicolumn{4}{l}{\textbf{ImageNet256}} \\
        \arrayrulecolor[rgb]{0.7, 0.7, 0.7}\hline
       \rule{0pt}{2ex}DDIM  & 25  & 1.99 & 5.47  & 0.80    & 0.47 \\
       PLMS4                & 25  & 2.05 & 4.71  & 0.82    & 0.49 \\
       \textbf{STSP4}       & 20  & 1.95 &\textbf{4.49}& \textbf{0.83}& \textbf{0.50} \\
       [-1.2em]\\
       \hdashline[1.2pt/1.2pt]\arrayrulecolor{black}
       \rule{0pt}{2ex}\textit{ADM-G}   & \textit{250} & \textit{20.9*}   & \textit{4.59}  & \textit{0.82}     & \textit{0.50} \\
        \\
        \multicolumn{4}{l}{\textbf{ImageNet512}} \\
        \arrayrulecolor[rgb]{0.7, 0.7, 0.7}\hline
       \rule{0pt}{2ex}DDIM  & 25  & 5.56 & 9.07  & {0.81}    & 0.42 \\
       PLMS4                & 25  & 5.78 & 16.00 & {0.75}    & \textbf{0.51} \\
       \textbf{STSP4}       & 20  & 5.13 &\textbf{8.24}& \textbf{0.83}& {0.45} \\
       [-1.2em]\\
       \hdashline[1.2pt/1.2pt]\arrayrulecolor{black}
       \rule{0pt}{2ex}\textit{ADM-G}   & \textit{250} & \textit{56.2*} & \textit{7.72}  & \textit{0.87}     & \textit{0.42} \\
       \bottomrule
    \end{tabular}}
  \end{minipage}
    \caption{Comparison of different numerical methods using a few steps on guided diffusion sampling. 
    Our methods and the best scores are highlighted in bold.
    We provide the reported scores from \cite{dhariwal2021diffusion} using 250 sampling steps, referred to as ADM-G in their paper. *ADM-G's sampling times are measured using our machine.}
    \label{tab:Sampling}
\end{table}

\subsection{Splitting methods in other tasks} \label{sec:splitting_for_other_tasks}
Besides class-conditional generation, our approach can also accelerate any conditional image generation as long as the gradient of the conditional function can be defined.
We test our approach on four tasks: text-to-image generation, image inpainting, colorization, and super-resolution. 


	
\textbf{Text-to-image generation:} We use a pre-trained text-to-image Disco-Diffusion \citep{Adam2021disco}  based on \cite{crowson2021clip}, which substitutes the classifier output with the dot product of the image and caption encodings from CLIP \citep{radford2021learning}. For more related experiments on Stable-Diffusion \citep{rombach2022high}, please refer to Appendix \ref{dreambooth}, \ref{clip_sd}. 

\textbf{Image inpainting \& colorization:} For these two tasks, we follow the techniques proposed in \cite{song2020score} and \cite{chung2022improving}, which improves the conditional functions of both tasks with ``manifold constraints.'' We use the same diffusion model \cite{dhariwal2021diffusion} trained on ImageNet as our earlier Experiments \ref{exp1}, \ref{exp2}.

\textbf{Super-resolution:} We follow the formulation from ILVR \citep{choi2021ilvr} combined with the manifold contraints \cite{chung2022improving}, and also use our earlier ImageNet diffusion model.

\begin{figure}
    \centering
    \setlength\tabcolsep{1.5pt}
    \begin{tabular}{cc|cccc}
        \shortstack{\scriptsize ``A beautiful painting of a singular \\
        \scriptsize lighthouse, shining its light across  \\
         \scriptsize  a tumultuous sea of blood, \\
         \scriptsize trending on artstation.''\vspace{3mm}}
       &
        \includegraphics[width=0.14\textwidth]{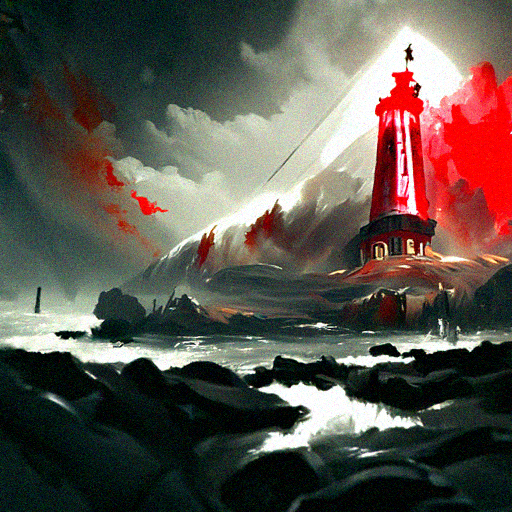}&
        \includegraphics[width=0.14\textwidth]{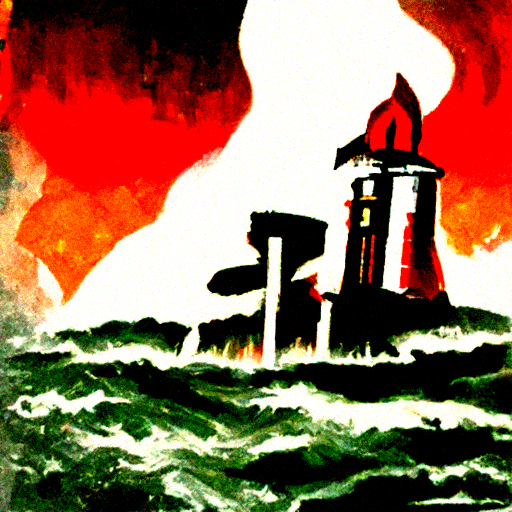}&
        \includegraphics[width=0.14\textwidth]{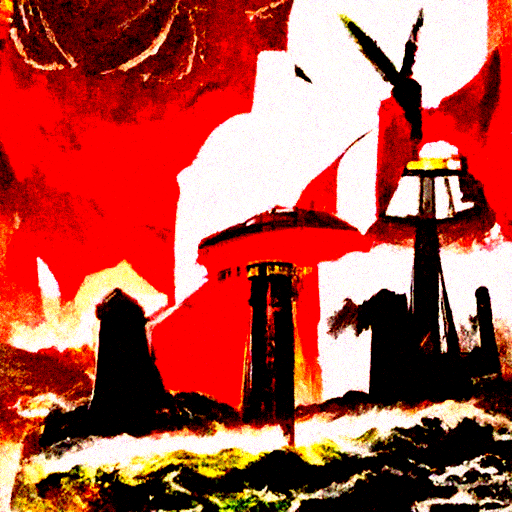}&
        \includegraphics[width=0.14\textwidth]{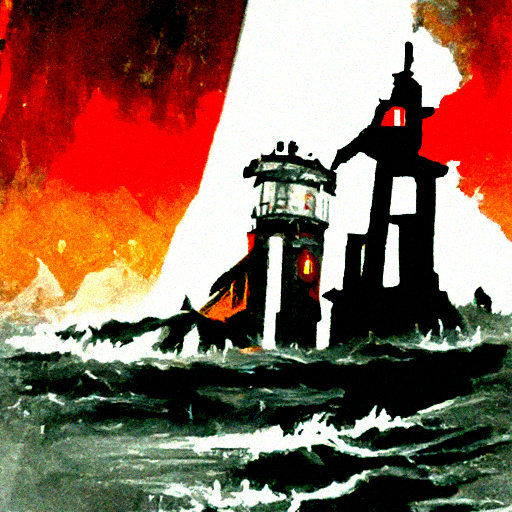}&
        \includegraphics[width=0.14\textwidth]{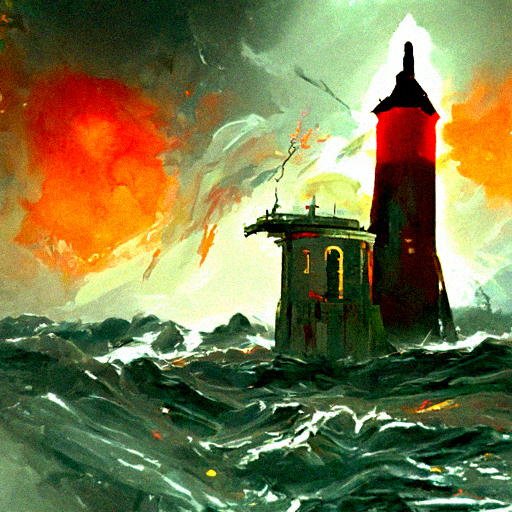}
        \\
        \shortstack{\scriptsize ``A beautiful painting of a starry \\
          \scriptsize night, over a sunflower sea,\\
          \scriptsize trending on artstation.'' \vspace{5mm}} &
         \includegraphics[width=0.14\textwidth]{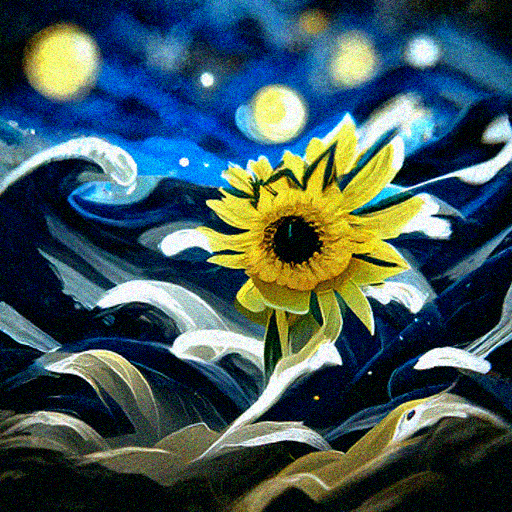} & 
         \includegraphics[width=0.14\textwidth]{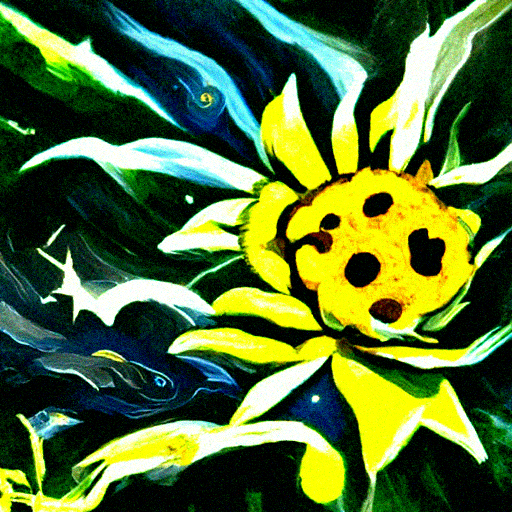} &
         \includegraphics[width=0.14\textwidth]{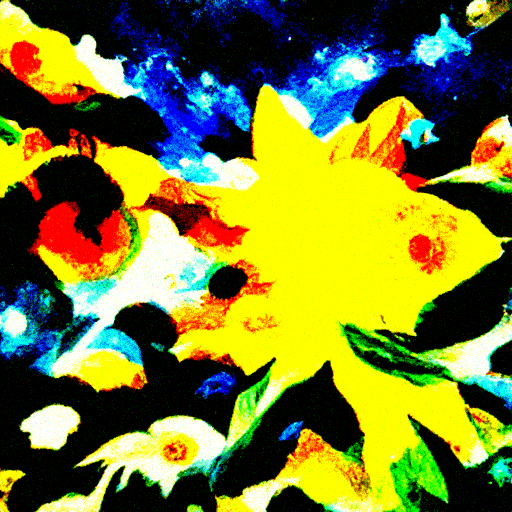} &
         \includegraphics[width=0.14\textwidth]{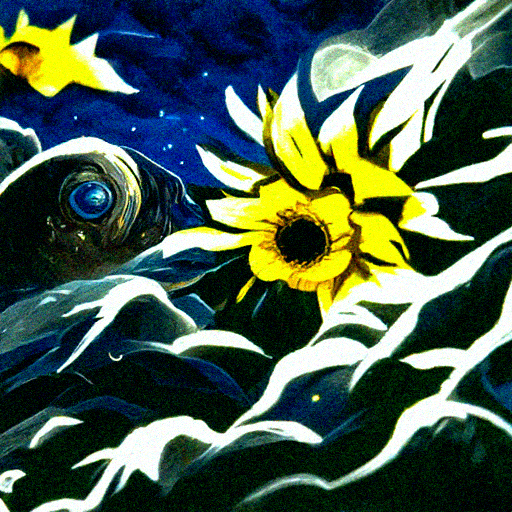} &
         \includegraphics[width=0.14\textwidth]{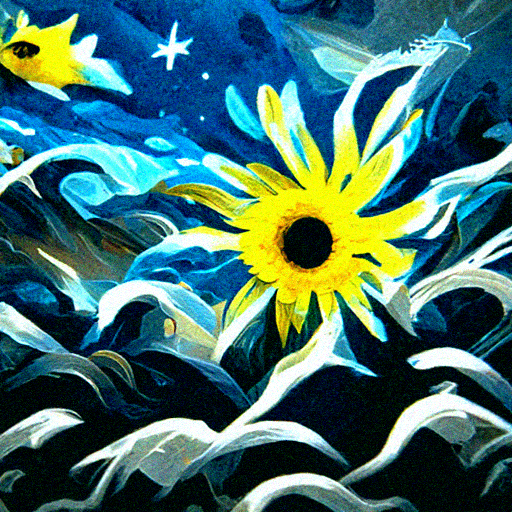} 
         \\
         &Full DDIM  & DDIM & PLMS4 & LTSP4 & STSP4 \\
        &\small (1,000 steps) & \small  (45 steps) & \small  (45 steps) & \small  (45 steps) & \small  (30 steps) \\
        & & \multicolumn{4}{c}{ \footnotesize (approximately using the same sampling time) } 
    \end{tabular}
        \caption{ Text-to-image generation using different sampling methods.}
    \label{fig_txt2img}
\end{figure}
\begin{figure*}
    \centering
    \setlength\tabcolsep{1.5pt}
    \begin{tabular}{cc|c|c@{}c@{}c|c@{}c@{}c}
        &Original & Input & \multicolumn{3}{c}{STSP4 (Ours)} & \multicolumn{3}{c}{DDIM} \\
        \multirow{2}{*}{\rotatebox{90}{\parbox[c]{1cm}{\centering Inpainting}}} &
        \includegraphics[width=0.1\textwidth]{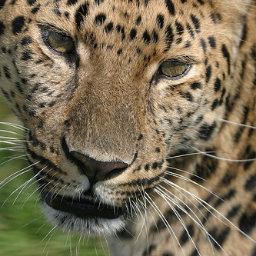} &
        \includegraphics[width=0.1\textwidth]{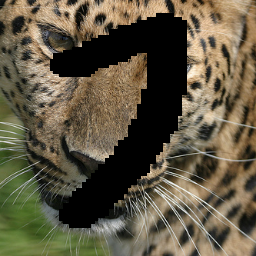}&
        \includegraphics[width=0.1\textwidth]{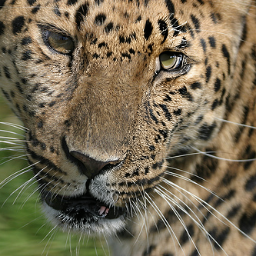}&
        \includegraphics[width=0.1\textwidth]{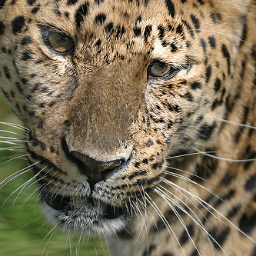}&
        \includegraphics[width=0.1\textwidth]{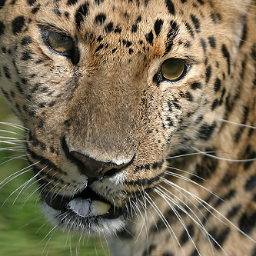}&
        \includegraphics[width=0.1\textwidth]{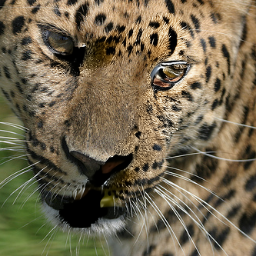}&
        \includegraphics[width=0.1\textwidth]{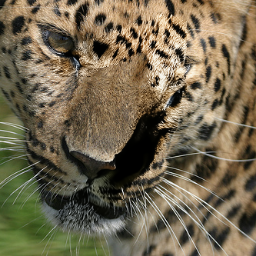}&
        \includegraphics[width=0.1\textwidth]{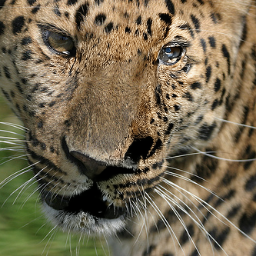}\\ &
        \includegraphics[width=0.1\textwidth]{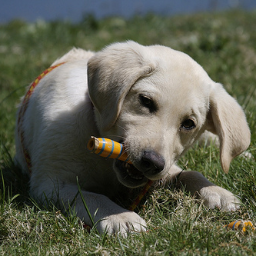} &
        \includegraphics[width=0.1\textwidth]{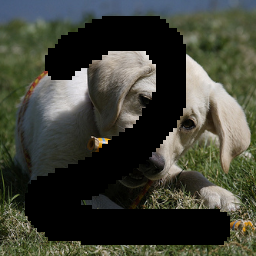}&
        \includegraphics[width=0.1\textwidth]{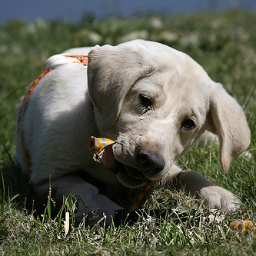}&
        \includegraphics[width=0.1\textwidth]{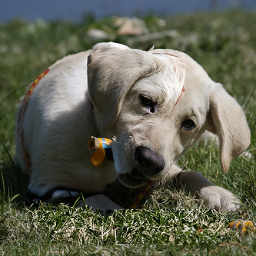}&
        \includegraphics[width=0.1\textwidth]{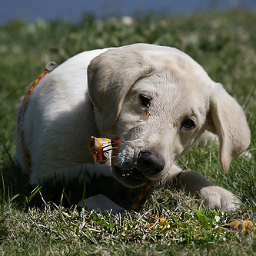}&
        \includegraphics[width=0.1\textwidth]{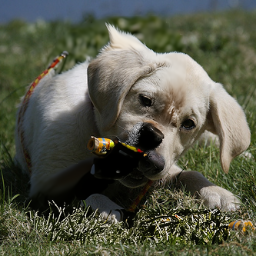}&
        \includegraphics[width=0.1\textwidth]{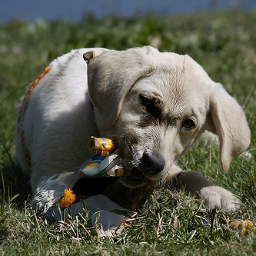}&
        \includegraphics[width=0.1\textwidth]{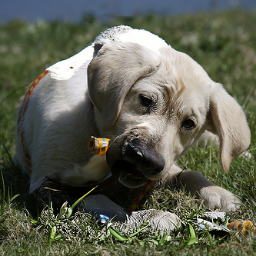}\\ 
        \hline  
        \multirow{2}{*}{\rotatebox{90}{\parbox[c]{1.25cm}{\centering Colorization}}} &
        \includegraphics[width=0.1\textwidth]{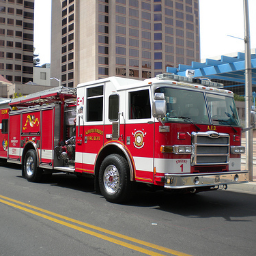} &
        \includegraphics[width=0.1\textwidth]{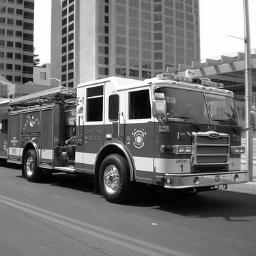}&
        \includegraphics[width=0.1\textwidth]{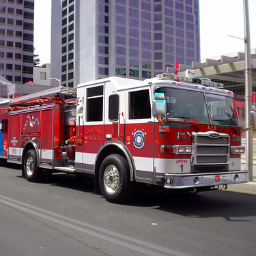}&
        \includegraphics[width=0.1\textwidth]{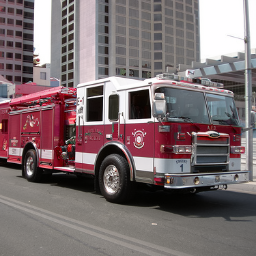}&
        \includegraphics[width=0.1\textwidth]{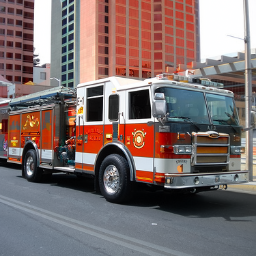}&
        \includegraphics[width=0.1\textwidth]{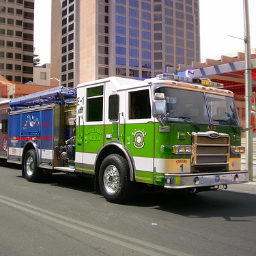}&
        \includegraphics[width=0.1\textwidth]{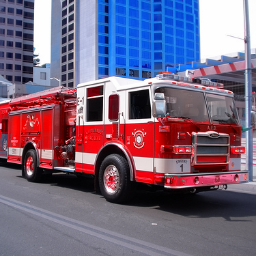}&
        \includegraphics[width=0.1\textwidth]{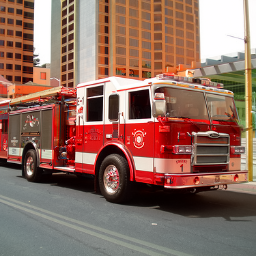} \\&
        \includegraphics[width=0.1\textwidth]{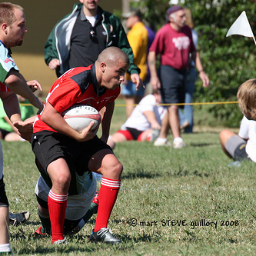} &
        \includegraphics[width=0.1\textwidth]{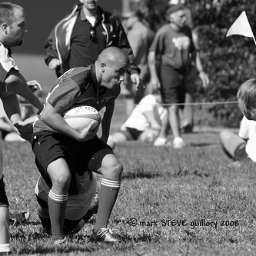}&
        \includegraphics[width=0.1\textwidth]{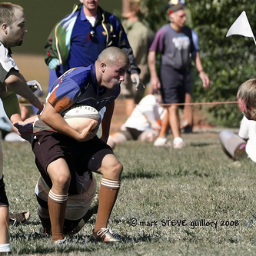}&
        \includegraphics[width=0.1\textwidth]{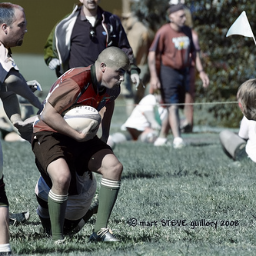}&
        \includegraphics[width=0.1\textwidth]{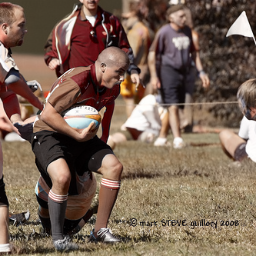}&
        \includegraphics[width=0.1\textwidth]{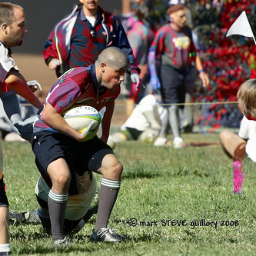}&
        \includegraphics[width=0.1\textwidth]{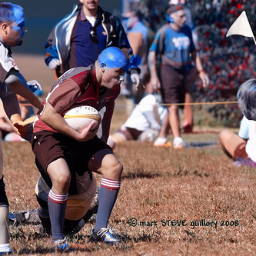}&
        \includegraphics[width=0.1\textwidth]{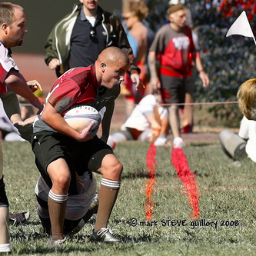} \\
        \hline  
        \multirow{2}{*}{\rotatebox{90}{\parbox[c]{1cm}{\centering 8xSuper-Resolution}}} &
        \includegraphics[width=0.1\textwidth]{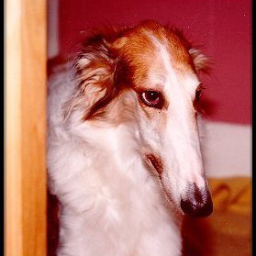} &
        \includegraphics[width=0.1\textwidth]{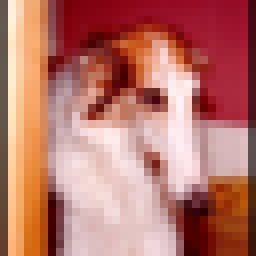}&
        \includegraphics[width=0.1\textwidth]{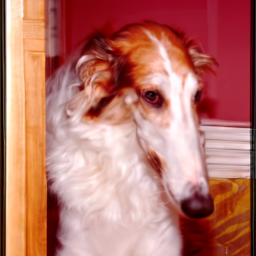}&
        \includegraphics[width=0.1\textwidth]{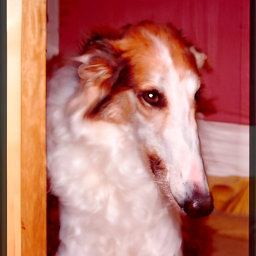}&
        \includegraphics[width=0.1\textwidth]{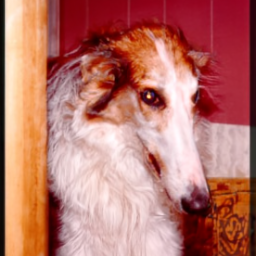}&
        \includegraphics[width=0.1\textwidth]{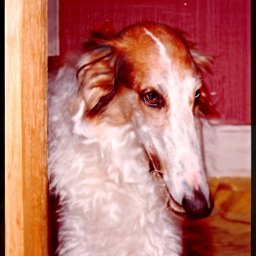}&
        \includegraphics[width=0.1\textwidth]{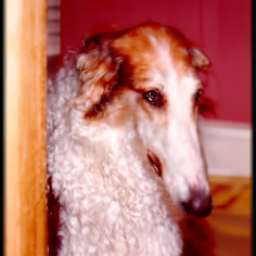}&
        \includegraphics[width=0.1\textwidth]{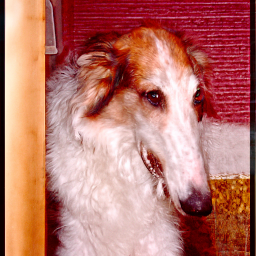}\\&
        \includegraphics[width=0.1\textwidth]{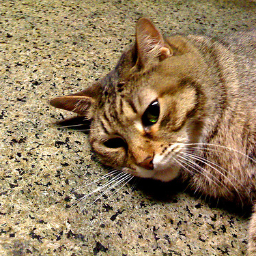} &
        \includegraphics[width=0.1\textwidth]{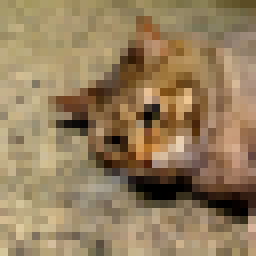}&
        \includegraphics[width=0.1\textwidth]{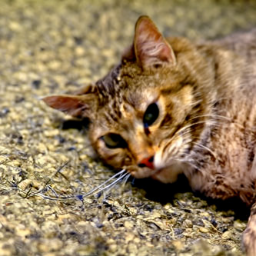}&
        \includegraphics[width=0.1\textwidth]{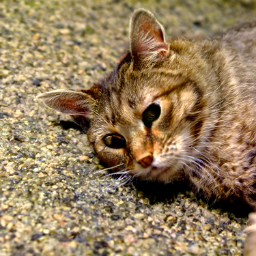}&
        \includegraphics[width=0.1\textwidth]{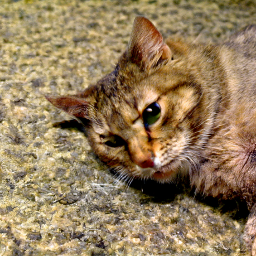}&
        \includegraphics[width=0.1\textwidth]{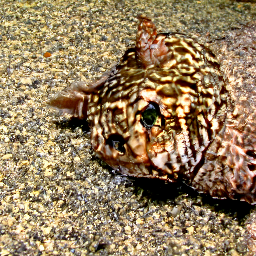}&
        \includegraphics[width=0.1\textwidth]{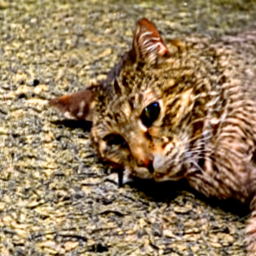}&
        \includegraphics[width=0.1\textwidth]{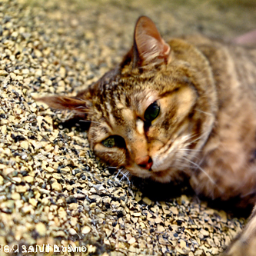}
    \end{tabular} 
    \caption{Guided-diffusion results of our STSP4 and DDIM on inpainting, colorization, and super-resolution.
    Both methods were limited to use approximately the same sampling time.
    }
    \label{fig_inpaint}
\end{figure*}

Figure \ref{fig_txt2img} compares our techniques, LTSP4 and STSP4, with the DDIM baseline and PLMS4 on text-to-image generation.
Each result is produced using a fixed sampling time of about 26 seconds.
STSP4, which uses 30 diffusion steps compared to 45 in the other methods, produces more realistic results with color contrast that is more similar to the full DDIM references'.
Figure \ref{fig_inpaint} shows that our STSP4 produces more convincing results than the DDIM baseline with fewer artifacts on the other three tasks while using the same 5 second sampling time.
Implementation details, quantitative evaluations, and more results are in Appendix \ref{txt2im}, \ref{con_gen}.

\section{Discussion}\label{sec:discussion}
Our findings show that when the sampling ODE consists of multiple terms from different networks, their numerical behaviors can be different and treating them separately can be more optimal.
Another promising direction is to improve the behavior of the gradient of the conditional function / classifier itself and study whether related properties such as adversarial robustness or gradient smoothness can induce the desirable temporal smoothness in the sampling ODE. However, it is not yet clear what specific characteristics of the behavior play an important role. This challenge may be related to a condition called ``stiffness'' in solving ODEs \cite{Hairer2010stiff}, which lacks a clear definition but describes the situation where explicit numerical methods, such as RK or PLMS, require a very small step size \emph{even in} a region with smooth curvature. 


As an alternative to the classifier-guided model, \cite{ho2022classifier} propose a classifier-free model that can perform conditional generation without a classifier while remaining a generative model.
This model can utilize high-order methods as no classifier is involved, but it requires evaluating the classifier-free network twice per step, which is typically more expensive than evaluating a normal diffusion model and a classifier. It is important to note that our accelerating technique and classifier-free models are \emph{not} mutually exclusive, and one can still apply a conditional function and our splitting technique to guide a classifier-free model in a direction it has not been trained for.



While our paper only focuses on ODEs derived from the deterministic sampling of DDIM, one can convert SDE-based diffusion models to ODEs \citep{karras2022elucidating} and still use our technique. More broadly, we can accelerate any diffusion model that can be expressed as a differential equation with a summation of two terms. When these terms behave differently, the benefit from splitting can be substantial.
Nevertheless, our findings are based on common, existing models and $\sigma$ schedule from \cite{dhariwal2021diffusion}. Further investigation into the impact of the $\sigma$ schedule or different types and architectures of diffusion models is still required.

\section{Conclusion}
In this paper, we investigate the failure to accelerate guided diffusion sampling of classical high-order numerical methods and propose a solution based on splitting numerical methods.
We found that the gradients of conditional functions are less suitable to classical high-order numerical methods and design a technique based on Strang splitting and a combination of forth- and first-order numerical methods. Our method achieves better LPIPS and FID scores than previous work given the same sampling time and is 32-58\% faster than a 250-step DDIM baseline. Our technique can successfully accelerate a variety of tasks, such as text-to-image generation, inpainting, colorization, and super-resolution.


\bibliography{iclr2022_conference}
\bibliographystyle{iclr2022_conference}

\appendix
\newpage
\appendix
\doparttoc
\faketableofcontents

\addcontentsline{toc}{section}{Appendix} 
\part{Appendix} 
\parttoc 

\section{Implementation}
Our implementation is available here
\footnote{https://github.com/sWizad/split-diffusion}.
The implementation is based on Katherine Crowson's guided-diffusion \footnote{https://github.com/crowsonkb/guided-diffusion}, which is inspired by OpenAI's guided-diffusion\footnote{https://github.com/openai/guided-diffusion}.
All of the pre-trained diffusion and classifier models are available here\footnote{https://github.com/openai/guided-diffusion/blob/main/model-card.md}. For evaluation, we use OpenAI’s measurement implementation with their reference image batch, which can be found here\footnote{https://github.com/openai/guided-diffusion/tree/main/evaluations}.

\section{Improving PLMS}
\label{iPLMS}
Initial points are required for using high-order PLMS.
The fourth-order formulation, for example, requires three initial points.
The original paper \citep{liu2022pseudo} employs the Runge-Kutta method to compute the initial points.
However, Runge-Kutta’s method has high computational costs and is inconvenient to use when the number of steps is small.
To reduce the computation costs, we compute the starting points of the higher-order PLMS using lower-order PLMS. Our PLMS can be summarized using Algorithm  \ref{algo:plms}.

\begin{algorithm}
\label{algo:plms}
\SetAlgoLined
    \textbf{input:} $\bar{x}_n$ (previous result), $\sigma_{n+1}$, $\sigma_n$, \\
    $\{e_i\}_{i<n}$ (evaluation buffer), $r$ (method order) \;
   \quad $e_n = \bar{\epsilon}_\sigma(\bar{x}_n)$ \; \quad $c= \min(r, n)$ \;
   \quad \textbf{if} $c==1$ \textbf{then}  \\ \quad \quad 
        $\hat{e} = e_n $ \;
   \quad \textbf{else if} $c==2$ \textbf{then} \\ \quad \quad 
        $\hat{e} = (3e_n - e_{n-1})/2 $ \;
   \quad \textbf{else if} $c==3$ \textbf{then} \\ \quad\quad 
        $\hat{e} = (23e_n - 16e_{n-1} + 5 e_{n-2})/12  $ \;
   \quad \textbf{else} \\ \quad\quad 
        $\hat{e} = (55e_n - 59e_{n-1} + 37e_{n-3} - 9e_{n-4})/24  $ \;
   \quad  \KwResult{ $\bar{x}_n + (\sigma_{n+1} - \sigma_n) \hat{e}$ } 
 \caption{PLMS}
 
\end{algorithm}

In Algorithm \ref{algo:plms}, the PLMS formulation is obtained by assuming a constant $\Delta \sigma$ for each step. The results in our experiments and previous work (e.g., \cite{liu2022pseudo, zhang2022fast}) are still reasonable when this assumption is not strictly satisfied, i.e., when $\Delta \sigma$ is not constant.
Inspired by \cite{zhang2022fast}, we also show how to derive another linear multi-step formulation for non-constant $\Delta \sigma$.
Let us first define a dummy variable $\tau$ in which $\Delta \tau$ is a constant in each time step.
To make it simple, let the value $\tau = 0$ when $t=T$ and when $t=0$, $\tau = N$, the total number of steps.
As a result, the discretization of $\tau$ can be defined by $\tau_n = n$ and $\Delta \tau = \tau_n - \tau_{n-1} = 1$ is a constant.
Next, we want to extrapolate the value of $\epsilon_\theta$ using a polynomial $P_\epsilon (\tau)$.

For example, the first order formulation can be produced by integrating a constant polynomial $P_\epsilon (\tau) = e_n$ :
\begin{align*}
    \frac{d\bar{x}}{d\sigma} = \bar{\epsilon}(\bar{x}) \approx&  P_\epsilon (\tau) = e_n, \\
    \int^{\bar{x}_{n+1}}_{\bar{x}_{n}} d\bar{x} = & \int^{\sigma_{n+1}}_{\sigma_{n}} e_n \: d\sigma ,\\
    \bar{x}_{n+1} - \bar{x}_{n} = & e_n \Delta \sigma
\end{align*}

which leads us to Euler's formulation.
Rather than using Lagrange's polynomial like \cite{zhang2022fast}, we use Newton's polynomial, which gives a nicer final formulation.
However, both are the same polynomial but have different expressions.
For the 2\ts{nd} order formulation, we interpolate between $(\tau_n, e_n)$ and $(\tau_{n-1},e_{n-1})$ using a Newton's polynomial $P_\epsilon (\tau) = e_n + (e_n-e_{n-1})(\tau-\tau_n)$.
$$ \int^{\bar{x}_{n+1}}_{\bar{x}_{n}} d\bar{x} = \int^{\sigma_{n+1}}_{\sigma_{n}} e_n + (e_n-e_{n-1})(\tau-\tau_n) d\sigma .$$

Every term is the same as in the 1\ts{st} order formulation, except for one term that is $\int^{\sigma_{n+1}}_{\sigma_{n}} (\tau-\tau_n) d\sigma $.
Let us use separable integration to approximate this term.
$$ \int^{\sigma_{n+1}}_{\sigma_{n}} (\tau-\tau_n) d\sigma = \int^{\tau_{n+1}}_{\tau_{n}} (\tau-\tau_n) \frac{d\sigma}{d\tau} d\tau \approx \int^{\tau_{n+1}}_{\tau_{n}} (\tau-\tau_n)  d\tau\int^{\tau_{n+1}}_{\tau_{n}} \frac{d\sigma}{d\tau} d\tau = \frac{\Delta\sigma}{2} $$
The result is $\bar{x}_{n+1} - \bar{x}_{n} =  e_n \Delta \sigma + (e_n - e_{n+1}) \frac{\Delta\sigma}{2}$, which is the PLMS2 formulation.
To compute this term more precisely, we need to know the derivation $d\sigma/d\tau$.
For this example, let us define $\sigma$ by 
\begin{equation}
 \label{sigma_sch}
    \sigma(\tau) = \exp \left(\ln{\sigma_\text{max}} + \frac{\tau}{N} (\ln{\sigma_\text{min}} - \ln{\sigma_\text{max}} ) \right) = \exp(a + \tau b),
\end{equation}
where $ a = \ln{\sigma_\text{max}} $ and $ b = (\ln{\sigma_\text{min}} - \ln{\sigma_\text{max}} ) / N$. Now, we have
$$ \int^{\tau_{n+1}}_{\tau_{n}} (\tau-\tau_n) \frac{d\sigma}{d\tau} d\tau = \Delta \sigma \left( \frac{\exp(b)}{\exp(b)-1} - \frac{1}{b} \right). $$
Consider $\frac{\exp(b)}{\exp(b)-1} - \frac{1}{b}$ when limit $N \rightarrow \infty$ (or $b \rightarrow 0$), this term is equal to $\frac{1}{2}$, which turns the formulation back to PLMS2.
When we set $N=30$ (or $b=-0.33$), the term $\frac{\exp(b)}{\exp(b)-1} - \frac{1}{b}=0.4723$, which is also close to $\frac{1}{2}$ in PLMS2 formulation.

We can continue using higher-order Newton's polynomials to obtain higher-order formulations.
We call this method GLMS (Generalized Linear Multi-Step) and summarize it with Algorithm \ref{algo:glms}.
In our comparison in Figure \ref{fig:exp3}, both PLMS and GLMS produce comparable results.
In the fourth order, GLMS4 performs slightly better than PLMS4.
However, the GLMS formulation is dependent on the $\sigma$ schedule, and the formulation must be revised if the $\sigma$ schedule changes.
As a result, we decided to use PLMS as part of our main algorithm for more flexibility.

\begin{algorithm}
\SetAlgoLined
    \textbf{input:} $\bar{x}_n$ (previous result), $\sigma_{n+1}$, $\sigma_{n}$, $\{e_i\}_{i<n}$ (evaluation buffer), $r$ (method order) \;
   \quad $b = \ln(\sigma_{n+1}) - \ln(\sigma_{n})$  \quad $e_n = \bar{\epsilon}_{\sigma}(\bar{x}_n)$ ; \quad $\{e_i\}$.append($e_n$) ; \quad $c= \min(r, n)$ \;   
   \quad \textbf{if} $c \geq 1$ \textbf{then}  \\ \quad \quad 
        $\hat{e} = e_n $ \;
   \quad \textbf{if} $c\geq 2$ \textbf{then} \\ \quad \quad 
        $\hat{e} = \hat{e} + (e_{n}-e_{n-1}) \left(\frac{\exp(b)}{\exp(b)-1}-\frac{1}{b}\right) $ \;
   \quad \textbf{if} $c \geq 3$ \textbf{then} \\ \quad \quad 
        $\hat{e} = \hat{e} + (e_{n}-2e_{n-1}+e_{n-2}) \left(\frac{\exp(b)}{\exp(b)-1}(2-\frac{2}{b})-\frac{1}{b} + \frac{2}{b^2}\right)  $ \;
   \quad \textbf{if} $c \geq 4$ \textbf{then} \\ \quad \quad 
        $\hat{e} = \hat{e} + (e_{n}-3e_{n-1}+3e_{n-2}-e_{n-3}) \left(\frac{\exp(b)}{\exp(b)-1}(6-\frac{9}{b}+\frac{6}{b^2})-\frac{2}{b} + \frac{6}{b^2} - \frac{6}{b^3}\right)  $ \;
   \quad  \KwResult{ $\bar{x}_n + (\sigma_{n+1} - \sigma_n) \hat{e}$ } 
 \caption{GLMS}
\label{algo:glms}
\end{algorithm}



\section{Numerical Methods for Unguided Diffusion}
\begin{figure}
    \centering
    \shortstack{ \includegraphics[width = 0.5\textwidth]{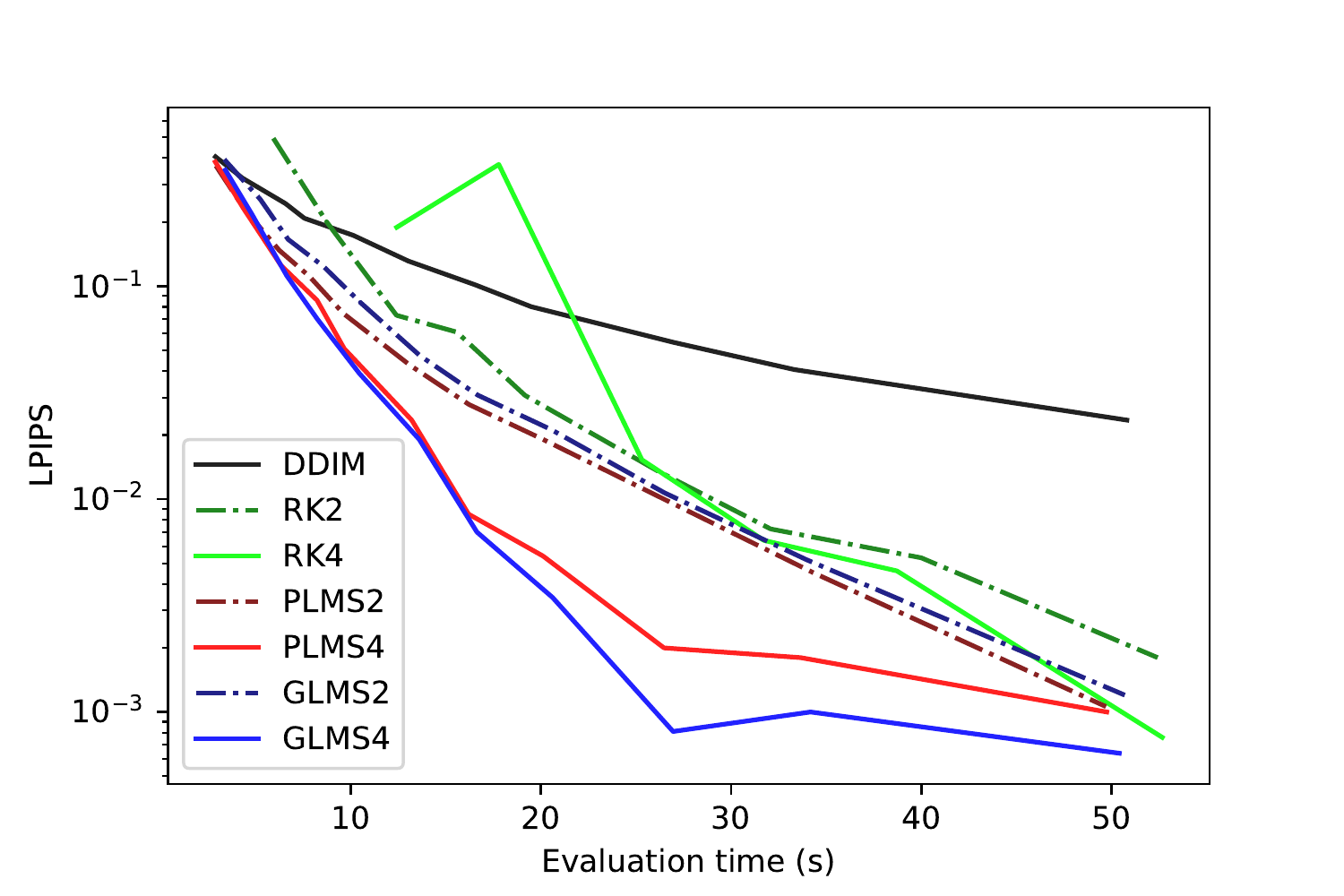}}
    \caption{Comparison of different numerical methods on an unguided diffusion model. Using samples from a 1000-step DDIM as reference solutions, we measure average LPIPS scores and plot them against the average sampling time.}
    \label{fig:exp3}
\end{figure}

This experiment evaluates the numerical methods used in our paper on \emph{unguided} sampling to see whether they may behave differently.
We perform a similar experiment as in Section \ref{exp1}, \ref{exp2} except on an unguided, unconditional diffusion model and use samples from 1,000-step DDIM as reference images. For each method, we use the same initial noise maps as the DDIM and evaluate the image similarity between its generated images and the reference images based on LPIPS. Figure \ref{fig:exp3} reports LPIPS vs sampling time for ImageNet128 using the $\sigma$ schedule in Equation \ref{sigma_sch}.

We found that RK2 and RK4 perform better than DDIM, but in our main papers these methods perform worse than DDIM when used on the split sub-problems for guided sampling.
Linear multi-step methods continue to be the best performers.
The graph also shows that higher order is generally better, especially when the number of steps is large.

\section{Comparing PLMS and DEIS} 
In Tables \ref{tab:deis_2} and \ref{tab:deis_4}, we compare the coefficients $e_n, e_{n-1}, e_{n-2}$, and $e_{n-3}$ of PLMS in Algorithm \ref{algo:plms} and the original implementation DEIS \cite{zhang2022fast} using the default linear schedule and 10 sampling steps. This shows that DEIS and PLMS are two similar methods. DEIS is analogous to PLMS with non-fixed coefficients. The coefficients can change depending on the number of steps and the noise schedule. However, the coefficients from both methods are close, and as previously stated, the DEIS coefficients will converge to the PLMS coefficients. Since DEIS's implementation relies heavily on the noise schedule, if the schedule changes, we must re-implement the method. For these reasons, we prefer PLMS over DEIS for practical purposes.
\begin{table}
  \begin{minipage}{0.5\columnwidth}
    \centering
    \shortstack{
    \begin{tabular}{c cccc}
        \toprule
         &\multicolumn{4}{c}{coefficient of } \\
       n\ts{th} step  & $e_n$  & $e_{n-1}$  & $e_{n-2}$ & $e_{n-3}$ \\
        \midrule
       1  & 1.0 & 0. & 0. & 0. \\
       2  & 1.5 & -0.5 & 0. & 0. \\
       3  & 1.5 & -0.5 & 0. & 0. \\
       4  & 1.5 & -0.5 & 0. & 0. \\
       5  & 1.5 & -0.5 & 0. & 0. \\
       6  & 1.5 & -0.5 & 0. & 0. \\
       7  & 1.5 & -0.5 & 0. & 0. \\
       8  & 1.5 & -0.5 & 0. & 0. \\
       9  & 1.5 & -0.5 & 0. & 0. \\
       10  & 1.5 & -0.5 & 0. & 0. \\
       \bottomrule
    \end{tabular}
    \\ \footnotesize (a) PLMS }
  \end{minipage}\hfill 
  \begin{minipage}{0.5\columnwidth}
    \centering
    \shortstack{
    \begin{tabular}{c cccc}
        \toprule
         &\multicolumn{4}{c}{coefficient of } \\
       n\ts{th} step  & $e_n$  & $e_{n-1}$  & $e_{n-2}$ & $e_{n-3}$ \\
        \midrule
       1  & 1.00 & 0. & 0. & 0. \\
       2  & 1.42 & -0.42 & 0. & 0. \\
       3  & 1.43 & -0.43 & 0. & 0. \\
       4  & 1.43 & -0.43 & 0. & 0. \\
       5  & 1.44 & -0.44 & 0. & 0. \\
       6  & 1.45 & -0.45 & 0. & 0. \\
       7  & 1.46 & -0.46 & 0. & 0. \\
       8  & 1.47 & -0.47 & 0. & 0. \\
       9  & 1.48 & -0.48 & 0. & 0. \\
       10  & 1.50 & -0.50 & 0. & 0. \\
       \bottomrule
    \end{tabular}
    \\ \footnotesize (b) DEIS }
  \end{minipage}
    \caption{Comparison of DEIS and PLMS on second order }
    \label{tab:deis_2}
\end{table}

\begin{table}
  \begin{minipage}{0.5\columnwidth}
    \centering
    \shortstack{
    \begin{tabular}{c cccc}
        \toprule
         &\multicolumn{4}{c}{coefficient of } \\
       n\ts{th} step  & $e_n$  & $e_{n-1}$  & $e_{n-2}$ & $e_{n-3}$ \\
        \midrule
       1  & 1.0  & 0.    & 0.   & 0. \\
       2  & 1.5  & -0.5  & 0.   & 0. \\
       3  & 1.92 & -1.33 & 0.41 & 0. \\
       4  & 2.29 & -2.46 & 1.54 & -0.38 \\
       5  & 2.29 & -2.46 & 1.54 & -0.38 \\
       6  & 2.29 & -2.46 & 1.54 & -0.38 \\
       7  & 2.29 & -2.46 & 1.54 & -0.38 \\
       8  & 2.29 & -2.46 & 1.54 & -0.38 \\
       9  & 2.29 & -2.46 & 1.54 & -0.38 \\
       10 & 2.29 & -2.46 & 1.54 & -0.38 \\
       \bottomrule
    \end{tabular}
    \\ \footnotesize (a) PLMS }
  \end{minipage}\hfill 
  \begin{minipage}{0.5\columnwidth}
    \centering
    \shortstack{
    \begin{tabular}{c cccc}
        \toprule
         &\multicolumn{4}{c}{coefficient of } \\
       n\ts{th} step  & $e_n$  & $e_{n-1}$  & $e_{n-2}$ & $e_{n-3}$ \\
        \midrule
       1  & 1.0  & 0.    & 0.   & 0. \\
       2  & 1.42 & -0.42 & 0.   & 0. \\
       3  & 1.77 & -1.12 & 0.34 & 0. \\
       4  & 2.09 & -2.07 & 1.28 & -0.34 \\
       5  & 2.11 & -2.11 & 1.31 & -0.32 \\
       6  & 2.14 & -2.16 & 1.35 & -0.33 \\
       7  & 2.16 & -2.22 & 1.38 & -0.34 \\
       8  & 2.20 & -2.28 & 1.42 & -0.35 \\
       9  & 2.23 & -2.35 & 1.47 & -0.35 \\
       10 & 2.23 & -2.28 & 1.55 & -0.38 \\
       \bottomrule
    \end{tabular}
    \\ \footnotesize (b) DEIS }
  \end{minipage}
    \caption{Comparison of DEIS and PLMS on forth order }
    \label{tab:deis_4}
\end{table}


\section{LPIPS vs. the number of sampling step} \label{step}
We report additional LPIPS results of the experiment in Section \ref{exp1} but with respect to the number of sampling steps.
The result shows that methods in the RK family can outperform other methods given the same number of sampling steps. However, once taken into account the slower evaluation time per step, these methods are overall slower to reach the same level of LPIPS of other methods.

\begin{figure}
    \centering
    \shortstack{ \includegraphics[width = 0.48\textwidth]{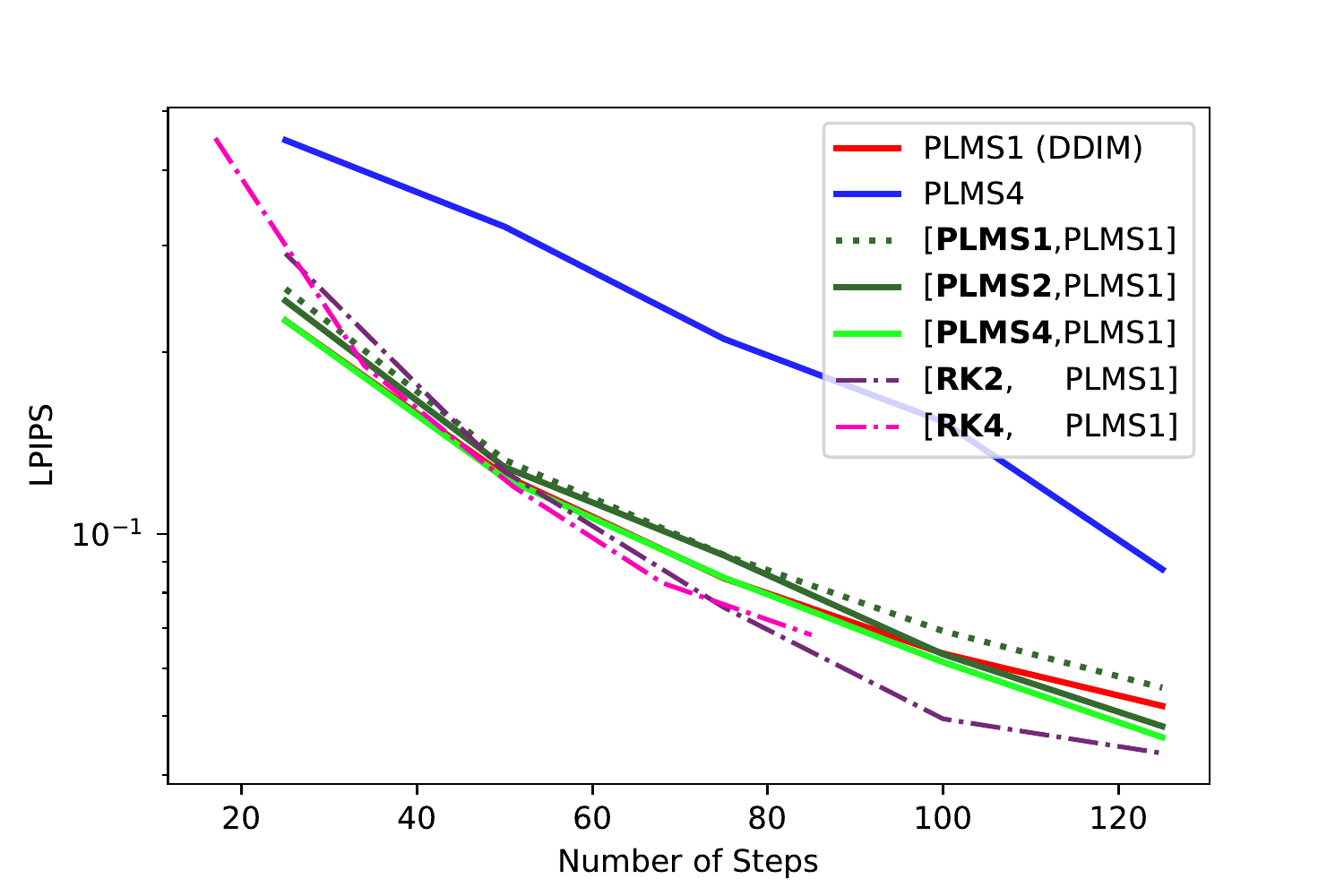}  
    \\ \footnotesize (a) Vary method for diffusion subproblem }
    \shortstack{ \includegraphics[width = 0.48\textwidth]{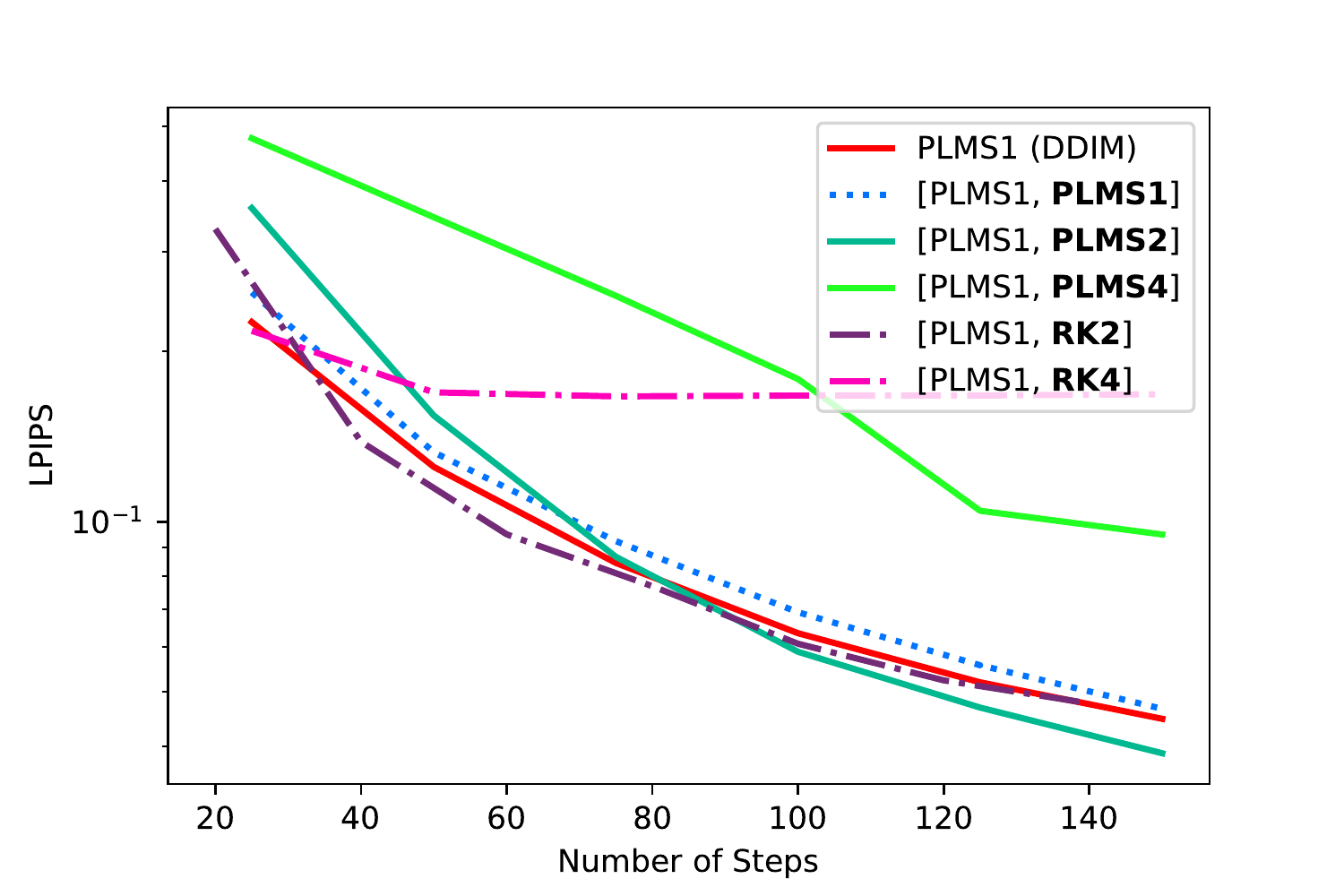}  
    \\ \footnotesize  (b) Vary method for condition subproblem}
    \caption{In addition to Figure \ref{exp1}, we plot LPIPS against their \textbf{sampling steps}. RK family can outperform other methods in many situations. However, methods in RK family took a longer time per diffusion step.}
    \label{fig:exp1_step}
\end{figure} 

\section{More Statistics for Experiement \ref{exp2}} \label{std}
We report the mean and standard deviation of each LPIPS score of the experiment in Section \ref{exp2} in Table \ref{tab:std}.
In Table \ref{tab:p-value}, we report the $p$-value for the null hypothesis that our STSP4 performs worse than other methods.
\begin{table}
    \centering
      \begin{tabular}{lcccc}
       \toprule
         &\multicolumn{4}{c}{Sampling time within} \\
          & 5 sec.  & 10 sec.  & 15 sec. & 20 sec. \\ 
       \midrule[0.08em]
       DDIM           & 0.111 $\pm$ .078 & 0.062 $\pm$ .065 & 0.042 $\pm$ .056 & 0.031 $\pm$ .048 \\
       PLMS4          & 0.240 $\pm$ .131 & 0.085 $\pm$ .112 & 0.039 $\pm$ .072 & 0.018 $\pm$ .040 \\
       RK2            & 0.152 $\pm$ .090 & 0.048 $\pm$ .051 & 0.030 $\pm$ .037 & 0.026 $\pm$ .037 \\
       RK4            & 0.190 $\pm$ .106 & 0.044 $\pm$ .022 & 0.033 $\pm$ .038 & 0.019 $\pm$ .032 \\
       \textbf{LTSP4} & 0.111 $\pm$ .092 & 0.056 $\pm$ .060 & 0.040 $\pm$ .054 & 0.028 $\pm$ .046 \\
       \textbf{STSP4} & \textbf{0.072} $\pm$ .065  & \textbf{0.033} $\pm$ .044  & \textbf{0.018} $\pm$ .028  & \textbf{0.012} $\pm$ .022  \\
       \bottomrule
    \caption{Our STSP4 has low average LPIPS scores and low standard deviations of LPIPS (N=120 samples).}
    \label{tab:std}
       \end{tabular}
\end{table}

\begin{table}
    \centering
      \begin{tabular}{lcccc}
       \toprule
         &\multicolumn{4}{c}{Sampling time within} \\
          & 5 sec.  & 10 sec.  & 15 sec. & 20 sec. \\ 
       \midrule[0.08em]
       DDIM         & $9.1\times10^{-11}$  & $9.0\times10^{-10}$  & $7.1\times10^{-9}$  & $9.2\times10^{-8}$  \\
       PLMS4        & $3.3\times10^{-41}$  & $2.1\times10^{-13}$  & $8.1\times10^{-4}$  & $3.7\times10^{-2}$  \\
       RK2          & $1.2\times10^{-30}$  & $6.4\times10^{-5}$   & $6.1\times10^{-5}$  & $2.5\times10^{-7}$  \\
       RK4          & $5.0\times10^{-31}$  & $6.2\times10^{-3}$   & $6.1\times10^{-5}$  & $1.6\times10^{-3}$  \\
       LTSP4        & $5.6\times10^{-11}$  & $9.1\times10^{-11}$  & $2.8\times10^{-10}$ & $1.7\times10^{-7}$  \\
       \bottomrule
    \caption{$p$-value for STSP4 has lower average LPIPS scores compare to other methods at approximately the same sampling time.}
    \label{tab:p-value}
       \end{tabular}
\end{table}

\section{STSP4 vs. More Numerical Methods Combination}\label{str_aba}
\begin{figure}
    \centering
    \shortstack{ \includegraphics[width = 0.48\textwidth]{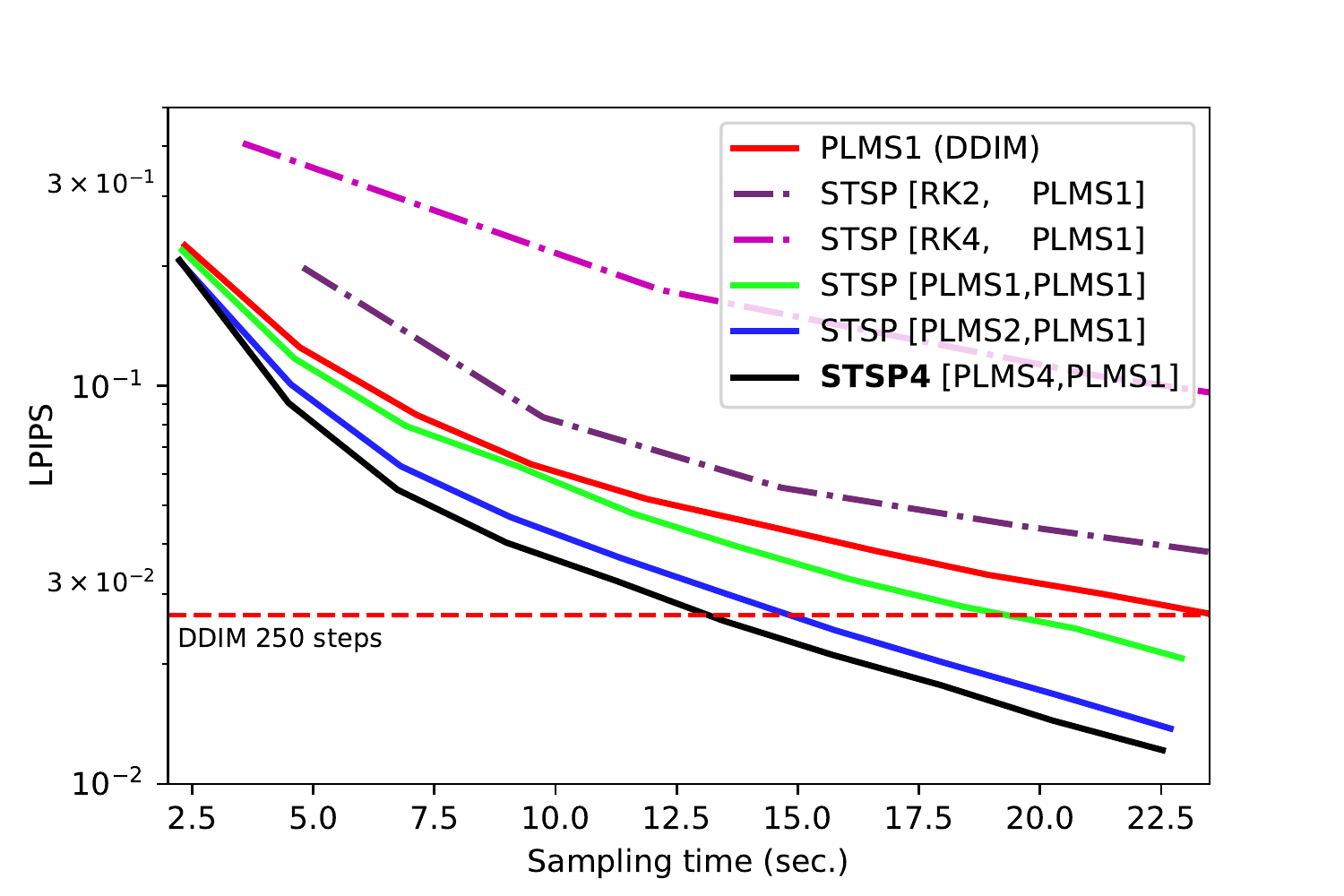}}
    \shortstack{ \includegraphics[width = 0.48\textwidth]{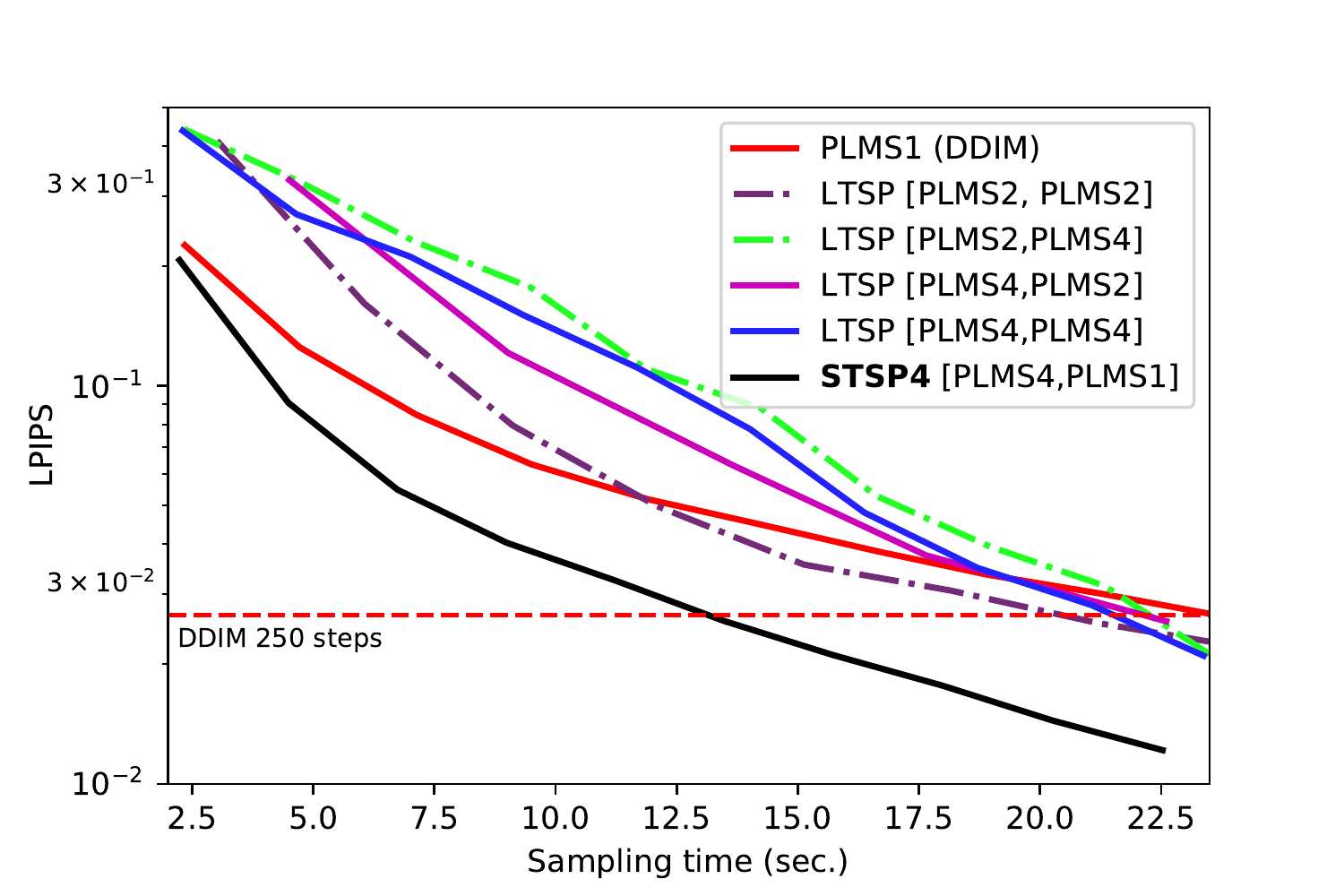}}
    \caption{We compare using Strang splitting with different numerical methods on the diffusion subproblem.}
    \label{fig:exp_stsp}
\end{figure}

We extend the experiment from Section \ref{exp2} and compare our STSP4 and DDIM baselines with more numerical method combinations.
Here we report LPIPS vs sampling time in Figure \ref{fig:exp_stsp}. 
The results show that Strang splitting method with [PLMS4, PLMS1] (our STSP4) is still the best combination, similar to the finding in Section \ref{exp2}.

\section{Comparison with DEIS and DPM-solver} 
We compare our splitting method to DPM-solver \cite{lu2022dpm} and DEIS \cite{zhang2022fast}. Since both methods have different implementation details, such as using different noise schedules, which affect the ODE solution, these methods do not converge to the same exact solution. Thus, it is not sensible to directly compare them, and we instead implement our method in their official implementations of DPM-solver \cite{lu2022dpm} and DEIS \cite{zhang2022fast}. We then compare the results with LPIPS on classifier-guided diffusion pretrained on ImageNet256 similar to Table \ref{tab:exp2} in Section \ref{fig:exp2}.
The results are shown in Table \ref{tab:dpm} and \ref{tab:deis}.
Our splitting methods can perform better than both DPM-solver \cite{lu2022dpm} and DEIS \cite{zhang2022fast}.

\begin{table}
    \centering
      \begin{tabular}{lcccc}
       \toprule
         &\multicolumn{4}{c}{Sampling time within} \\
          & 5 sec.  & 10 sec.  & 15 sec. & 20 sec. \\ 
       \midrule[0.08em]
       DPM-Solver-1 (DDIM) & 0.333 & 0.125 & 0.080  & 0.045 \\
       DPM-Solver-2        & 0.565 & 0.188 & 0.078  & 0.045 \\
       DPM-Solver-3        & 0.540 & 0.233 & 0.087  & 0.043 \\
       \textbf{LTSP4}      & 0.185 & 0.105 & 0.071  & 0.048 \\
       \textbf{STSP4}      & \textbf{0.169} & \textbf{0.062} & \textbf{0.061}  & \textbf{0.037} \\
       \bottomrule
    \caption{The comparison with DPM-solver \cite{lu2022dpm}. Average LPIPS when the sampling time is limited to be under 5-20 seconds.}
    \label{tab:dpm}
       \end{tabular}
\end{table}

\begin{table}
    \centering
      \begin{tabular}{lcccc}
       \toprule
         &\multicolumn{4}{c}{Sampling time within} \\
          & 3 sec.  & 6 sec.  & 9 sec. & 12 sec. \\ 
       \midrule[0.08em]
       0-DEIS (DDIM) & 0.333 & 0.125 & 0.080  & 0.045 \\
       1-DEIS        & 0.466 & 0.193 & 0.092  & 0.044 \\
       3-DEIS        & 0.625 & 0.511 & 0.433  & 0.345 \\
       \textbf{LTSP4}& 0.321 & 0.120 & 0.080  & 0.048 \\
       \textbf{STSP4}& \textbf{0.212} & \textbf{0.080} & \textbf{0.046}  & \textbf{0.031} \\
       \bottomrule
    \caption{Comparison with DEIS \cite{zhang2022fast}. We report the average LPIPS scores when the sampling time is limited to be under 3-12 seconds.}
    \label{tab:deis}
       \end{tabular}
\end{table}

\section{Experiment on FID vs. Sampling Time} 
\begin{figure}
    \centering
    \shortstack{ \includegraphics[width = 0.48\textwidth]{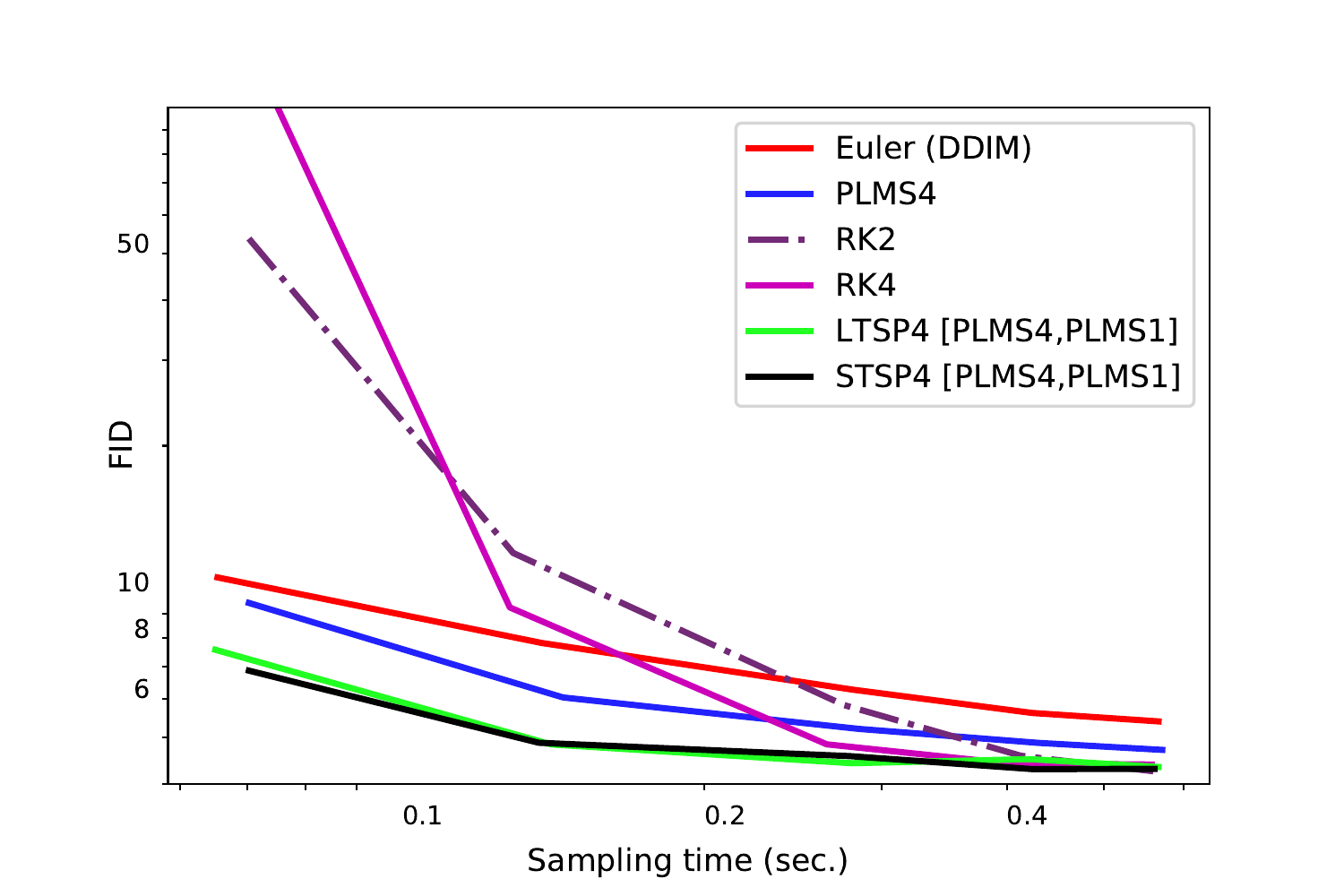}}
    \caption{FID vs. Sampling time measured on ImageNet128}
    \label{fig:fid_time}
\end{figure}

In this section, we demonstrate how splitting methods can accelerate guided diffusion sampling using FID scores. We vary the number of sampling steps, generate 20k samples, and compare the FID of the splitting methods (LTSP4 and STSP4) to many non-splitting methods, including DDIM, PLMS4, RK2, and RK4. Figure \ref{fig:fid_time} shows a comparison of FID vs. sampling time for each method. This experiment uses a classifier-guided diffusion model that was pretrained on the ImageNet128 dataset by \cite{dhariwal2021diffusion}. 
Our method can generate good sample quality, especially when the average sampling is limited to under 0.3 seconds (or about 50 steps of DDIM for 128$\times$128 resolution), while other methods show large FID scores at lower sampling steps or require more time to generate similarly high-quality samples. 

\section{Text-guided image generation} \label{txt2im}
For text-to-image generation or text-guided image generation, our implementation is based on Disco-Diffusion v3.1\footnote{https://colab.research.google.com/drive/1bItz4NdhAPHg5-u87KcH-MmJZjK-XqHN}, which also relies on \cite{crowson2021clip}.
We use v3.1 that does not contain unrelated features; however, our method can be applied to any of the versions.
We use the fine-tuned 512x512 diffusion model from Katherine Crowson \footnote{http://batbot.tv/ai/models/guided-diffusion/512x512\_diffusion\_uncond\_finetune\_008100.pt} and use pre-trained OpenAI’s CLIP models \footnote{https://github.com/openai/CLIP} including RN50, ViTB32, and ViTB16. 
For other model and conditional function configurations, we use Disco-Diffusion v3.1’s default settings.

\section{Controllable generation}
\label{con_gen}
We provide implementation details of our sampling algorithm for other conditional generation tasks.

\textbf{Image inpainting}:
Given a target masked image, $y_0$, and a mask $P$ which is a matrix of $0,1$ values, we want to sample $x_0$ so that $y_0 = Px_0$. The key concept from \cite{song2020score} is to use reverse diffusion in the unmasked area and forward diffusion in the masked area through an additional step called the impose step.
Let $x'_{t-1}$ be an unconditional diffusion sample from $x_t$ and $y_{t-1}$ be a forward diffusion sample from $y_0$. The impose step can be summarized as follows:
\begin{equation}
    x_{t-1}  =  (I-P^TP)x'_{t-1} + P^T y_{t-1}.
\end{equation}
To sample a corrupted target image $y_{t-1}$, we use the sampling formulation from \cite{song2020denoising} to sample from $y_0$ by
\begin{equation} \label{impu_forward}
    y_t \sim \mathcal{N}(\sqrt{\bar{\alpha}_t}y_0 + \sqrt{1-\bar{\alpha}_t-\eta^2_t} \epsilon_\theta(x_t, t), \eta^2_t I )
\end{equation}
where $\eta_t = \sqrt{(1-\bar{\alpha}_{t-1})/(1-\bar{\alpha}_t}) \sqrt{1-\bar{\alpha}_{t}/\bar{\alpha}_{t-1}}$.
This formulation works more effectively with numerical methods than the original sampling formulation from \cite{song2020score}, which is $y_t \sim \mathcal{N}(\sqrt{\bar{\alpha}_t}y_0, (1-\bar{\alpha}_t)  I )$.

Rather than sampling $x'_{t-1}$ unconditionally, we follow \cite{chung2022improving} and use guided diffusion sampling with the conditional function defined by:
\begin{equation} \label{impu_condi}
    f(x_t) = \frac{1}{2\gamma} || y_0 - P \hat{x}_0(x_t) ||^2_2,  \quad \hat{x}_0(x_t) = \frac{x_t - \sqrt{ 1 - \bar{\alpha}_t} \epsilon_\theta (x_t, t)}{\sqrt{\bar{\alpha}_t}},
\end{equation}
where $\gamma$ is a control parameter. 
Since directly computing $\epsilon_\theta (x_t, t)$ with the diffusion network in Equation \ref{impu_forward} and \ref{impu_condi} is time consuming, we use a secondary-model method from Katherine Crowson\footnote{https://colab.research.google.com/drive/1mpkrhOjoyzPeSWy2r7T8EYRaU7amYOOi} to speed it up.

\textbf{Image colorization}: The idea behind colorization is very similar to inpainting.
We convert images from the RGB format to the HSV format, with the grayscale value in the first channel, using an orthogonal matrix:
$$C = \begin{bmatrix}
0.577 & -0.816 & 0\\
0.577 & 0.408 & 0.707 \\
0.577 & 0.408 & -0.707 \\
\end{bmatrix}.$$
To mask out other channels and keep only the first grayscale channel, we define the mask matrix $P$ as 1 in the grayscale channel and 0 in the other two channels.
The impose step and conditional function can be defined as 
$$  x_{t-1}  = (I-C^TP^TPC)x'_{t-1} + C^TP^T PC y_{t-1}, \quad f(x_t) = \frac{1}{2\gamma} || y_0 - PC \hat{x}_0 (x_t) ||^2. $$

\textbf{Image super-resolution}: Let us denote $D$ as a down-sampling matrix.
The impose step and conditional function is given by 
$$  x_{t-1}  = (I-D^TD)x'_{t-1} + D^T y_{t-1}, \quad f(x_t) = \frac{1}{2\gamma} || y_0 - D \hat{x}_0 (x_t) ||^2. $$

In our implementation, we replace the $D$ and $D^T$ operations with the ILVR \citep{choi2021ilvr} down-sampling and up-sampling functions.

\begin{figure*}
    \centering
    \setlength\tabcolsep{1.5pt}
    \begin{tabular}{c|c|c@{}c@{}c|c@{}c@{}c}
        Original & Input & &STSP4 & & & DDIM \\
        \includegraphics[width=0.1\textwidth]{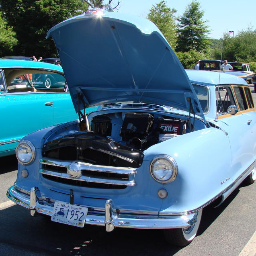} &
        \includegraphics[width=0.1\textwidth]{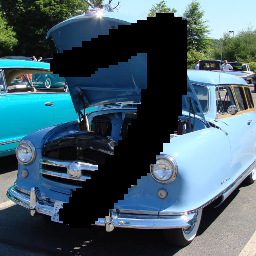}&
        \includegraphics[width=0.1\textwidth]{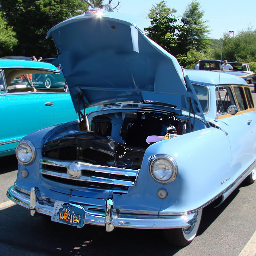}&
        \includegraphics[width=0.1\textwidth]{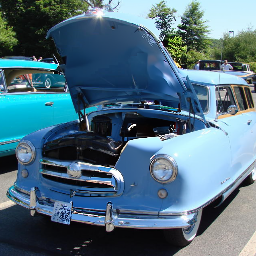}&
        \includegraphics[width=0.1\textwidth]{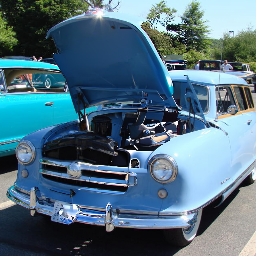}&
        \includegraphics[width=0.1\textwidth]{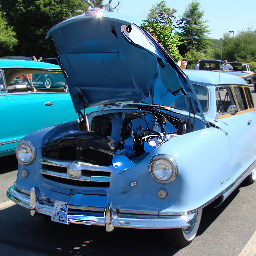}&
        \includegraphics[width=0.1\textwidth]{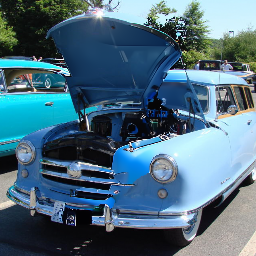}&
        \includegraphics[width=0.1\textwidth]{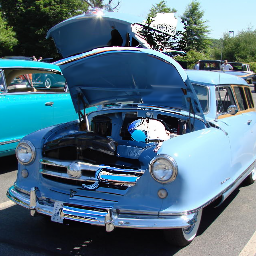}\\ 
        \includegraphics[width=0.1\textwidth]{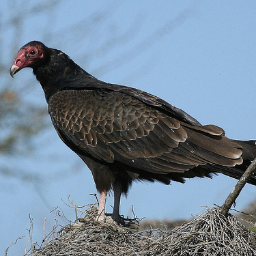} &
        \includegraphics[width=0.1\textwidth]{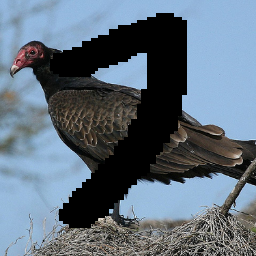}&
        \includegraphics[width=0.1\textwidth]{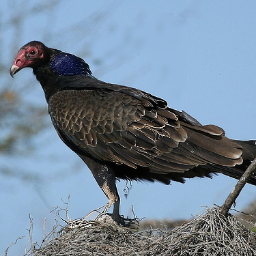}&
        \includegraphics[width=0.1\textwidth]{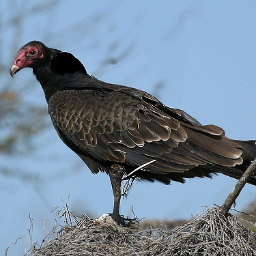}&
        \includegraphics[width=0.1\textwidth]{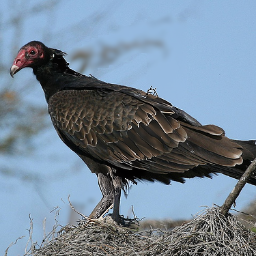}&
        \includegraphics[width=0.1\textwidth]{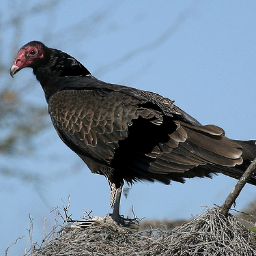}&
        \includegraphics[width=0.1\textwidth]{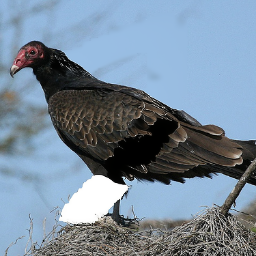}&
        \includegraphics[width=0.1\textwidth]{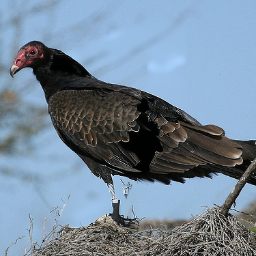}\\ 
        \includegraphics[width=0.1\textwidth]{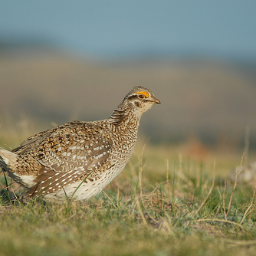} &
        \includegraphics[width=0.1\textwidth]{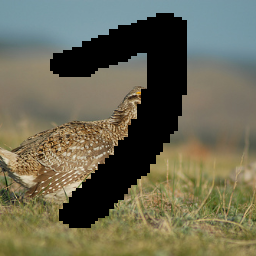}&
        \includegraphics[width=0.1\textwidth]{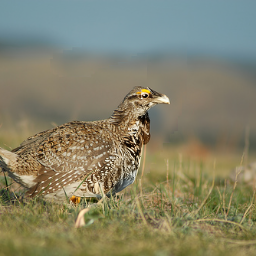}&
        \includegraphics[width=0.1\textwidth]{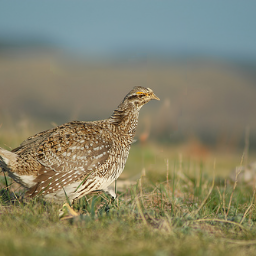}&
        \includegraphics[width=0.1\textwidth]{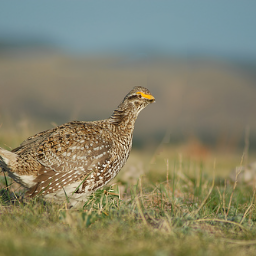}&
        \includegraphics[width=0.1\textwidth]{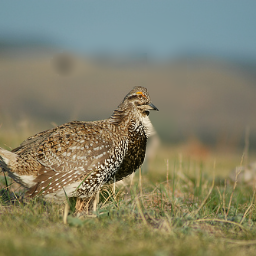}&
        \includegraphics[width=0.1\textwidth]{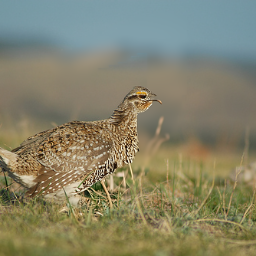}&
        \includegraphics[width=0.1\textwidth]{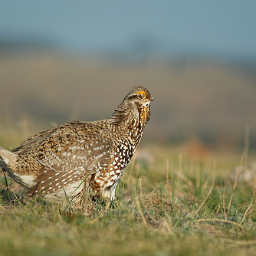}\\
        \includegraphics[width=0.1\textwidth]{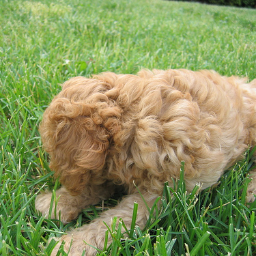} &
        \includegraphics[width=0.1\textwidth]{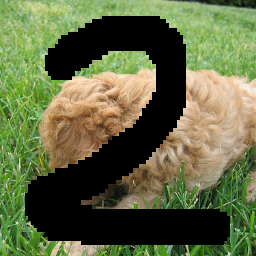}&
        \includegraphics[width=0.1\textwidth]{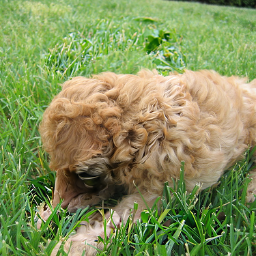}&
        \includegraphics[width=0.1\textwidth]{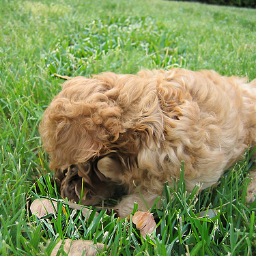}&
        \includegraphics[width=0.1\textwidth]{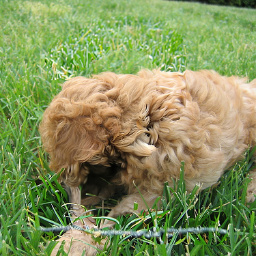}&
        \includegraphics[width=0.1\textwidth]{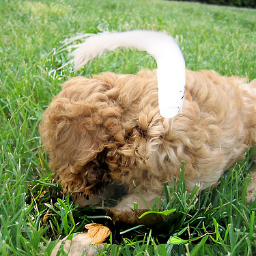}&
        \includegraphics[width=0.1\textwidth]{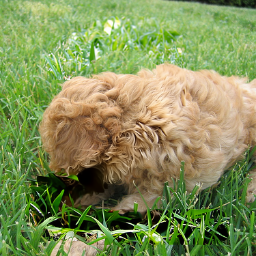}&
        \includegraphics[width=0.1\textwidth]{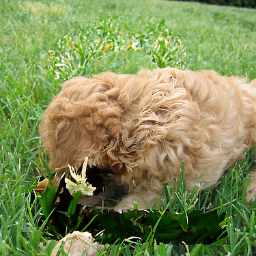}\\ 
        \includegraphics[width=0.1\textwidth]{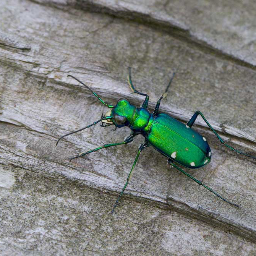} &
        \includegraphics[width=0.1\textwidth]{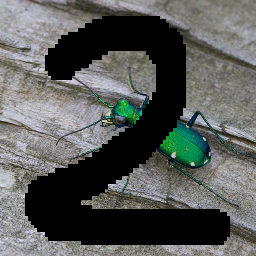}&
        \includegraphics[width=0.1\textwidth]{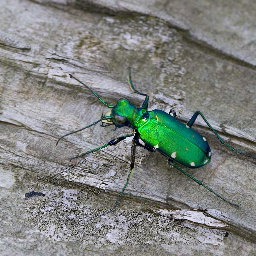}&
        \includegraphics[width=0.1\textwidth]{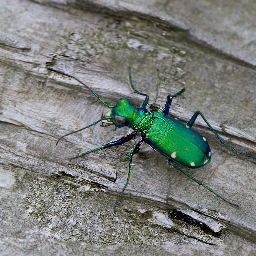}&
        \includegraphics[width=0.1\textwidth]{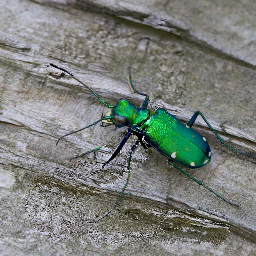}&
        \includegraphics[width=0.1\textwidth]{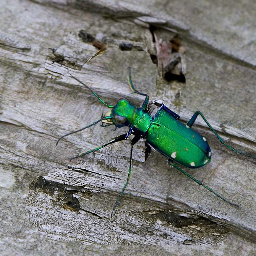}&
        \includegraphics[width=0.1\textwidth]{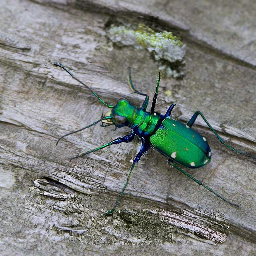}&
        \includegraphics[width=0.1\textwidth]{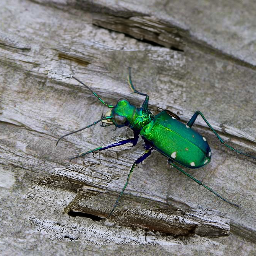}\\ 
        \includegraphics[width=0.1\textwidth]{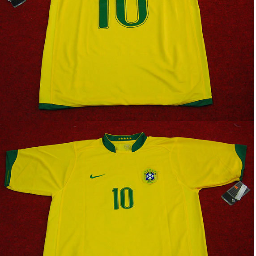} &
        \includegraphics[width=0.1\textwidth]{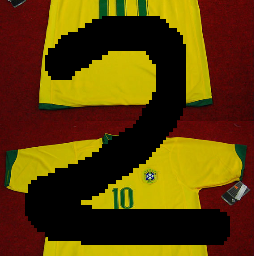}&
        \includegraphics[width=0.1\textwidth]{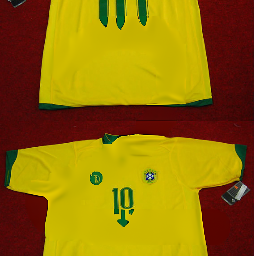}&
        \includegraphics[width=0.1\textwidth]{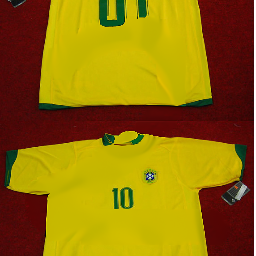}&
        \includegraphics[width=0.1\textwidth]{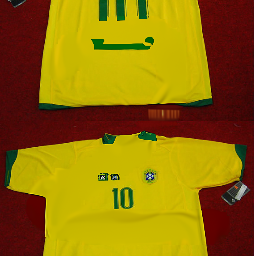}&
        \includegraphics[width=0.1\textwidth]{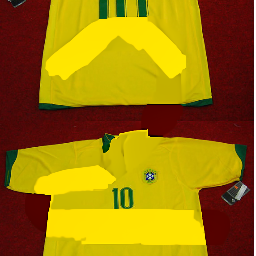}&
        \includegraphics[width=0.1\textwidth]{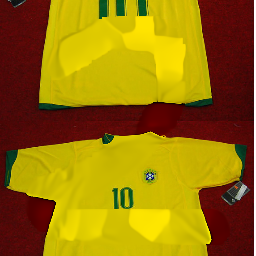}&
        \includegraphics[width=0.1\textwidth]{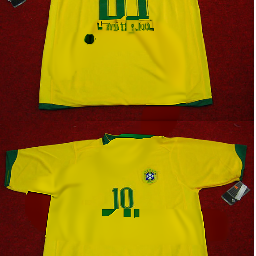}
    \end{tabular} 
    \caption{Additional inpainting results on ImageNet256.
    We show three different results generated by each method for each input. All results are generated using the same 5 second sampling time.
    }
    \label{apx_fig_inpaint}
    \vspace{-3.5mm}
\end{figure*}

\begin{figure*}
    \centering
    \setlength\tabcolsep{1.5pt}
    \begin{tabular}{c|c|c@{}c@{}c|c@{}c@{}c}
        Original & Input & &STSP4 & & & DDIM \\
        \includegraphics[width=0.1\textwidth]{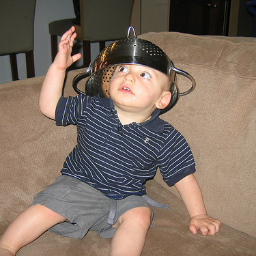} &
        \includegraphics[width=0.1\textwidth]{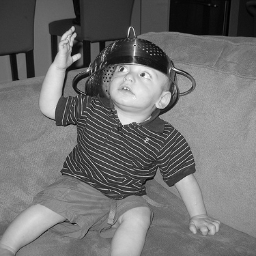}&
        \includegraphics[width=0.1\textwidth]{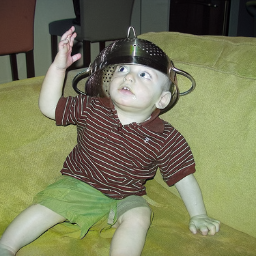}&
        \includegraphics[width=0.1\textwidth]{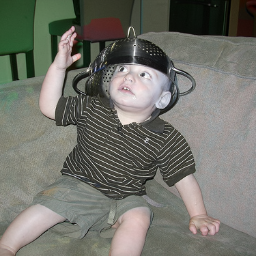}&
        \includegraphics[width=0.1\textwidth]{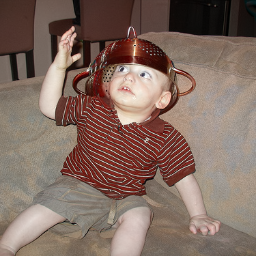}&
        \includegraphics[width=0.1\textwidth]{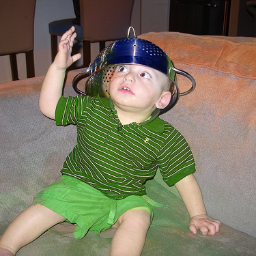}&
        \includegraphics[width=0.1\textwidth]{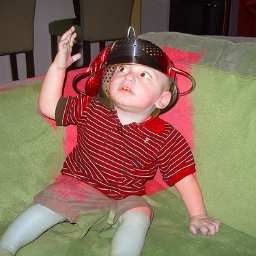}&
        \includegraphics[width=0.1\textwidth]{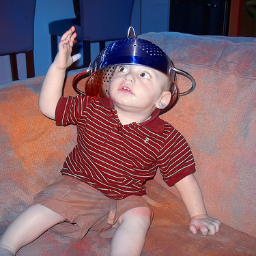}\\ 
        \includegraphics[width=0.1\textwidth]{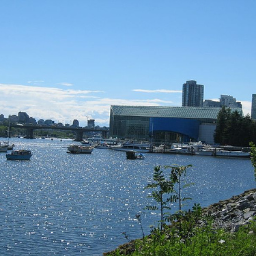} &
        \includegraphics[width=0.1\textwidth]{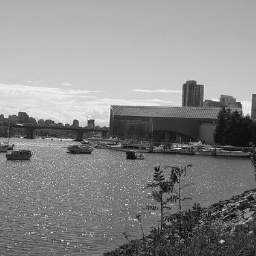}&
        \includegraphics[width=0.1\textwidth]{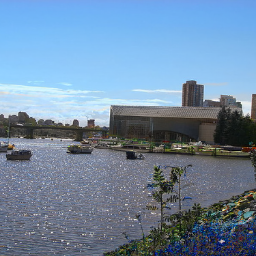}&
        \includegraphics[width=0.1\textwidth]{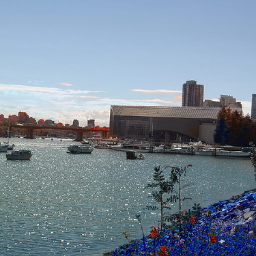}&
        \includegraphics[width=0.1\textwidth]{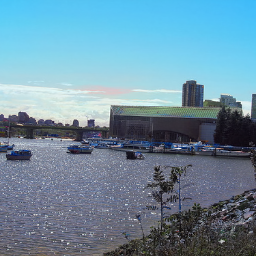}&
        \includegraphics[width=0.1\textwidth]{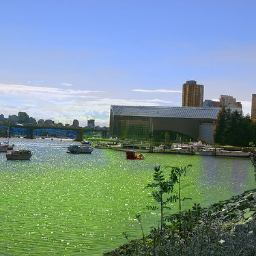}&
        \includegraphics[width=0.1\textwidth]{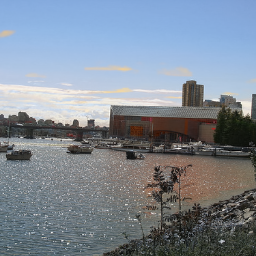}&
        \includegraphics[width=0.1\textwidth]{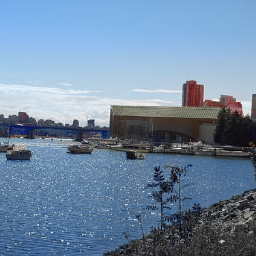}\\ 
        \includegraphics[width=0.1\textwidth]{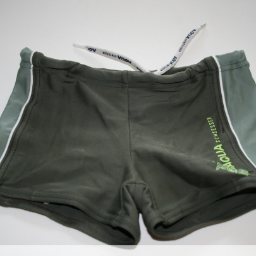} &
        \includegraphics[width=0.1\textwidth]{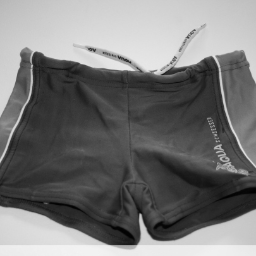}&
        \includegraphics[width=0.1\textwidth]{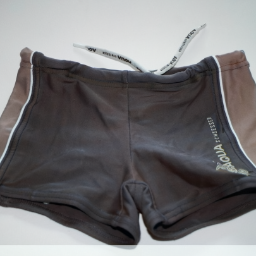}&
        \includegraphics[width=0.1\textwidth]{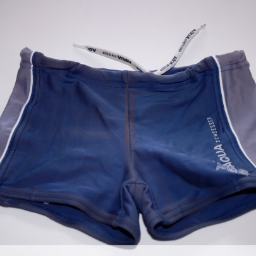}&
        \includegraphics[width=0.1\textwidth]{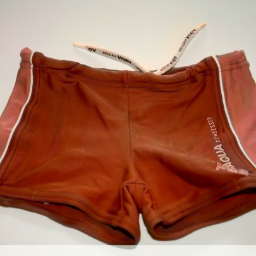}&
        \includegraphics[width=0.1\textwidth]{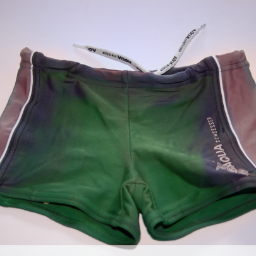}&
        \includegraphics[width=0.1\textwidth]{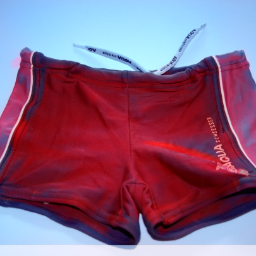}&
        \includegraphics[width=0.1\textwidth]{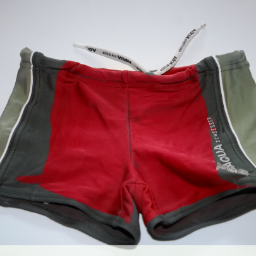}\\  
        \includegraphics[width=0.1\textwidth]{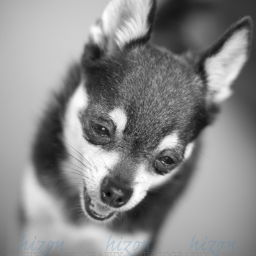} &
        \includegraphics[width=0.1\textwidth]{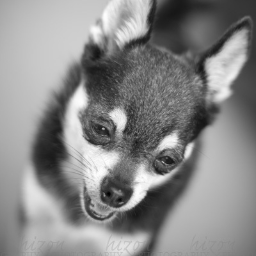}&
        \includegraphics[width=0.1\textwidth]{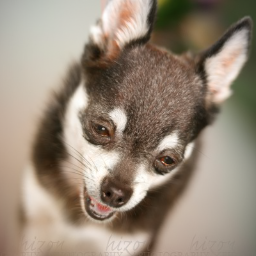}&
        \includegraphics[width=0.1\textwidth]{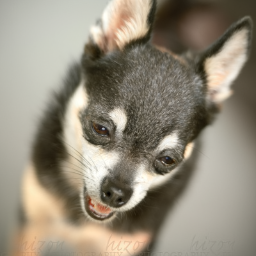}&
        \includegraphics[width=0.1\textwidth]{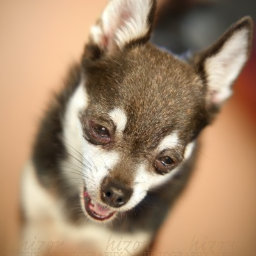}&
        \includegraphics[width=0.1\textwidth]{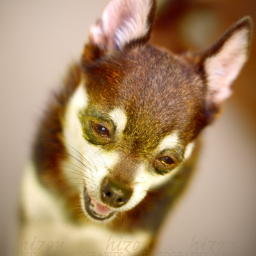}&
        \includegraphics[width=0.1\textwidth]{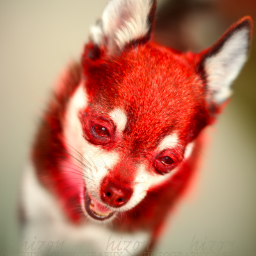}&
        \includegraphics[width=0.1\textwidth]{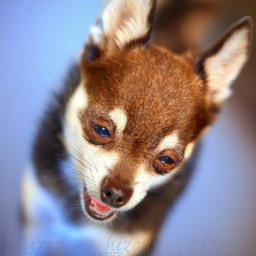}\\ 
        \includegraphics[width=0.1\textwidth]{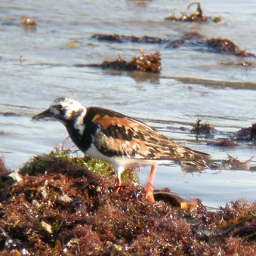} &
        \includegraphics[width=0.1\textwidth]{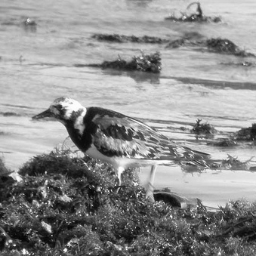}&
        \includegraphics[width=0.1\textwidth]{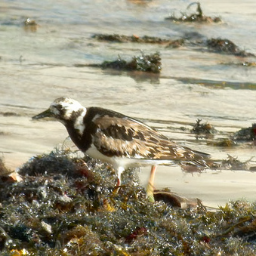}&
        \includegraphics[width=0.1\textwidth]{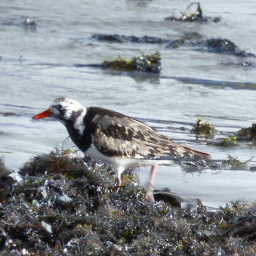}&
        \includegraphics[width=0.1\textwidth]{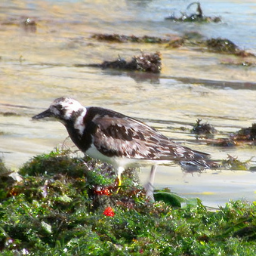}&
        \includegraphics[width=0.1\textwidth]{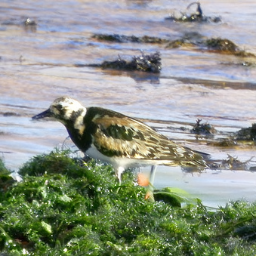}&
        \includegraphics[width=0.1\textwidth]{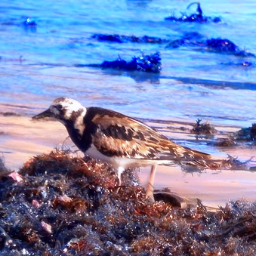}&
        \includegraphics[width=0.1\textwidth]{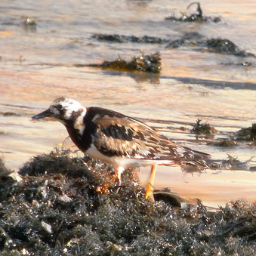}\\ 
        \includegraphics[width=0.1\textwidth]{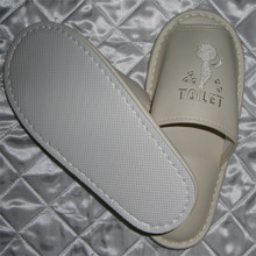} &
        \includegraphics[width=0.1\textwidth]{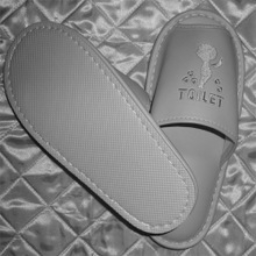}&
        \includegraphics[width=0.1\textwidth]{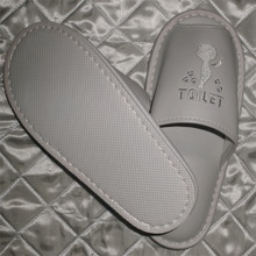}&
        \includegraphics[width=0.1\textwidth]{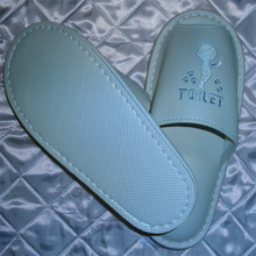}&
        \includegraphics[width=0.1\textwidth]{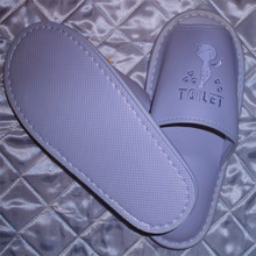}&
        \includegraphics[width=0.1\textwidth]{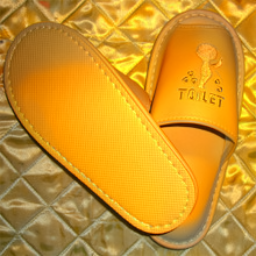}&
        \includegraphics[width=0.1\textwidth]{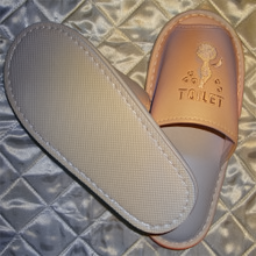}&
        \includegraphics[width=0.1\textwidth]{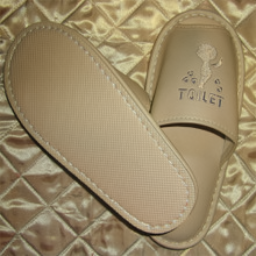}
    \end{tabular} 
    \caption{Additional colorization results on ImageNet256.
    We show three different results generated by each method for each input. All results are generated using the same 5 second sampling time.
    }
    \label{apx_fig_color}
    \vspace{-2.5mm}
\end{figure*}

\begin{figure*}
    \centering
    \setlength\tabcolsep{1.5pt}
    \begin{tabular}{c|c|c@{}c@{}c|c@{}c@{}c}
        Original & Input & &STSP4 & & & DDIM \\
        \includegraphics[width=0.1\textwidth]{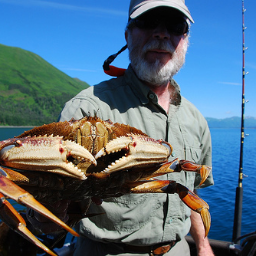} &
        \includegraphics[width=0.1\textwidth]{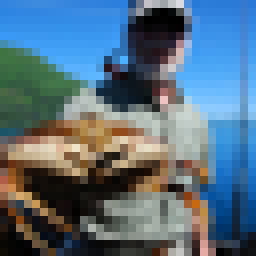}&
        \includegraphics[width=0.1\textwidth]{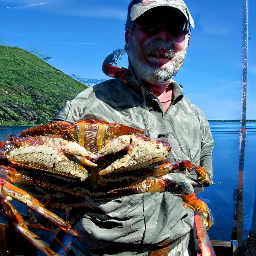}&
        \includegraphics[width=0.1\textwidth]{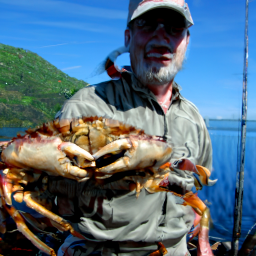}&
        \includegraphics[width=0.1\textwidth]{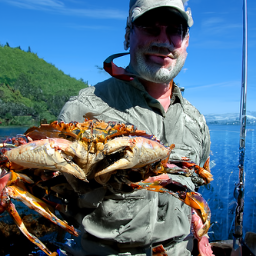}&
        \includegraphics[width=0.1\textwidth]{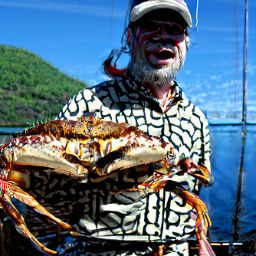}&
        \includegraphics[width=0.1\textwidth]{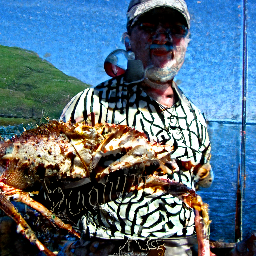}&
        \includegraphics[width=0.1\textwidth]{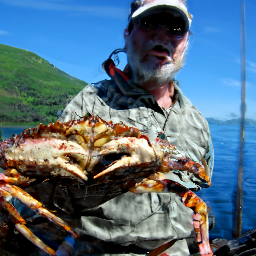}\\ 
        \includegraphics[width=0.1\textwidth]{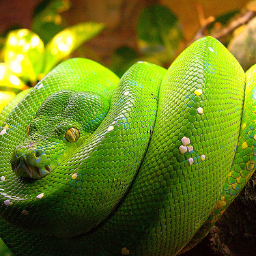} &
        \includegraphics[width=0.1\textwidth]{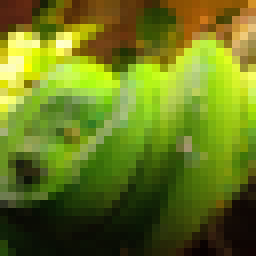}&
        \includegraphics[width=0.1\textwidth]{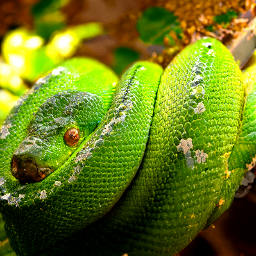}&
        \includegraphics[width=0.1\textwidth]{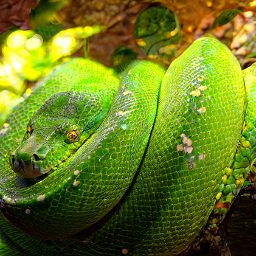}&
        \includegraphics[width=0.1\textwidth]{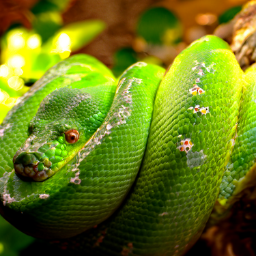}&
        \includegraphics[width=0.1\textwidth]{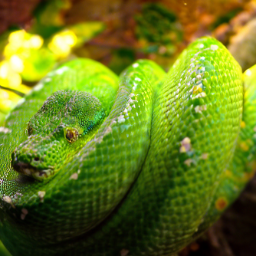}&
        \includegraphics[width=0.1\textwidth]{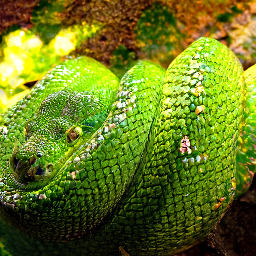}&
        \includegraphics[width=0.1\textwidth]{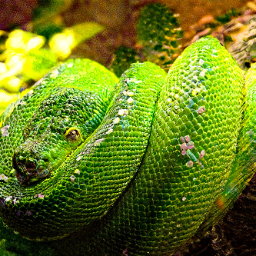}\\ 
        \includegraphics[width=0.1\textwidth]{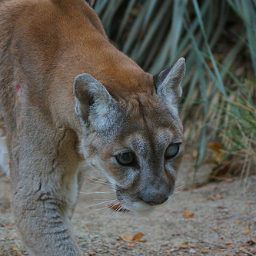} &
        \includegraphics[width=0.1\textwidth]{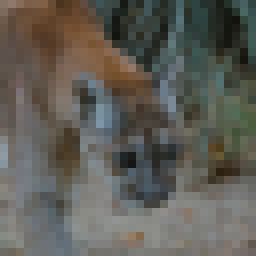}&
        \includegraphics[width=0.1\textwidth]{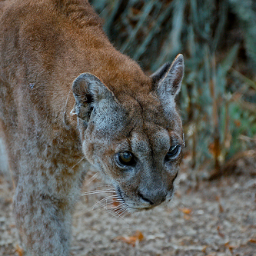}&
        \includegraphics[width=0.1\textwidth]{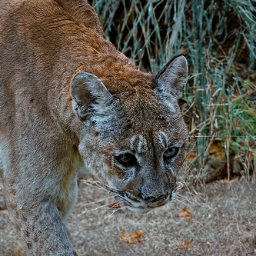}&
        \includegraphics[width=0.1\textwidth]{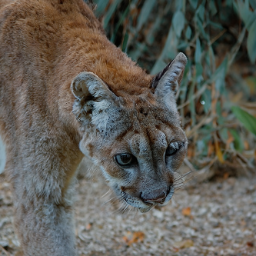}&
        \includegraphics[width=0.1\textwidth]{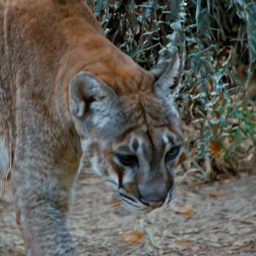}&
        \includegraphics[width=0.1\textwidth]{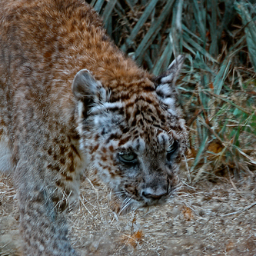}&
        \includegraphics[width=0.1\textwidth]{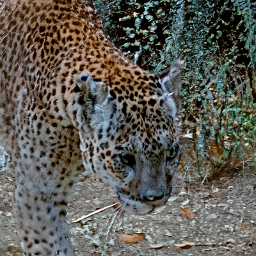}\\  
        \includegraphics[width=0.1\textwidth]{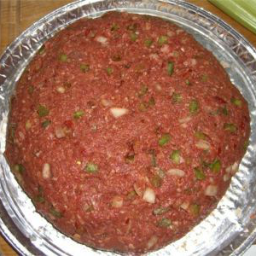} &
        \includegraphics[width=0.1\textwidth]{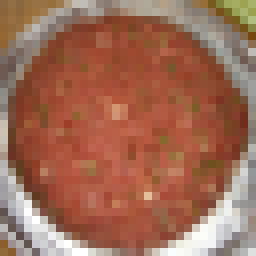}&
        \includegraphics[width=0.1\textwidth]{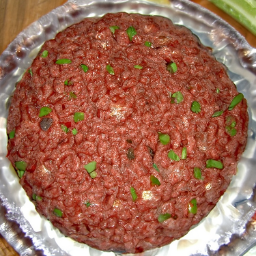}&
        \includegraphics[width=0.1\textwidth]{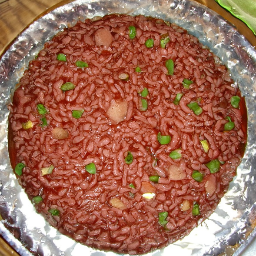}&
        \includegraphics[width=0.1\textwidth]{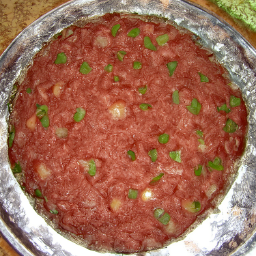}&
        \includegraphics[width=0.1\textwidth]{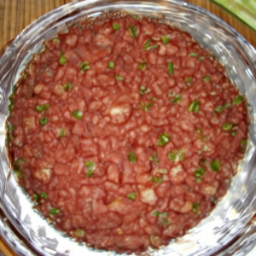}&
        \includegraphics[width=0.1\textwidth]{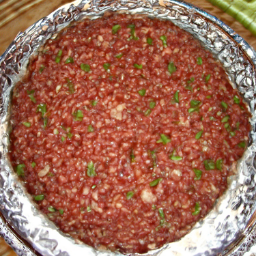}&
        \includegraphics[width=0.1\textwidth]{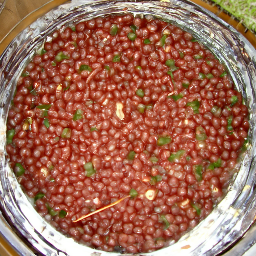}\\ 
        \includegraphics[width=0.1\textwidth]{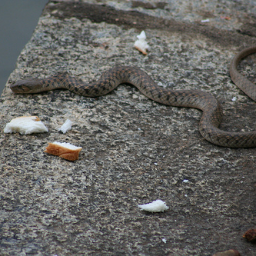} &
        \includegraphics[width=0.1\textwidth]{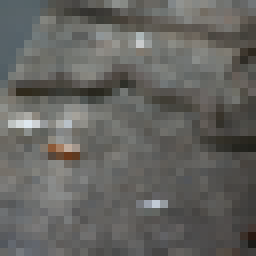}&
        \includegraphics[width=0.1\textwidth]{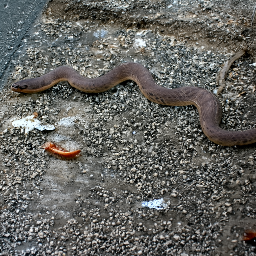}&
        \includegraphics[width=0.1\textwidth]{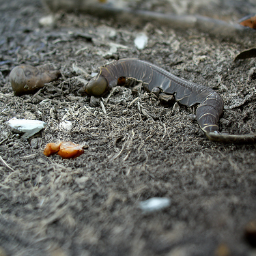}&
        \includegraphics[width=0.1\textwidth]{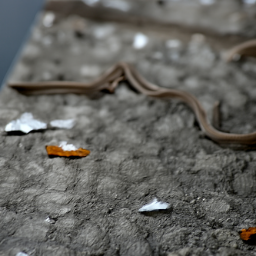}&
        \includegraphics[width=0.1\textwidth]{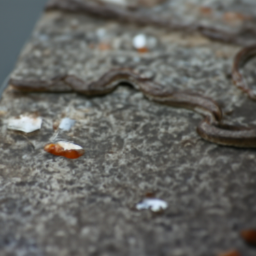}&
        \includegraphics[width=0.1\textwidth]{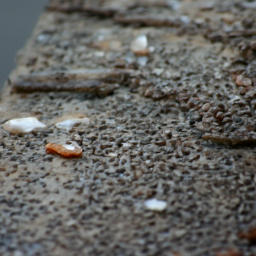}&
        \includegraphics[width=0.1\textwidth]{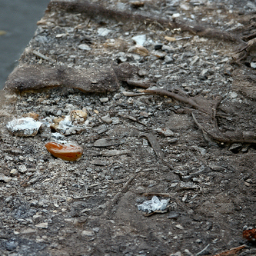}\\ 
        \includegraphics[width=0.1\textwidth]{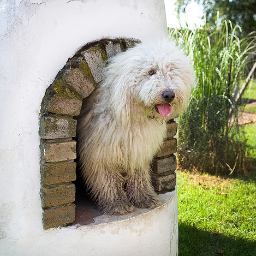} &
        \includegraphics[width=0.1\textwidth]{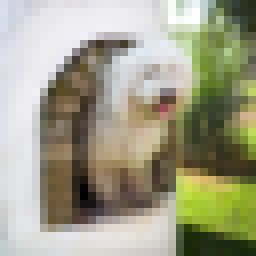}&
        \includegraphics[width=0.1\textwidth]{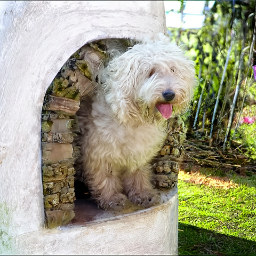}&
        \includegraphics[width=0.1\textwidth]{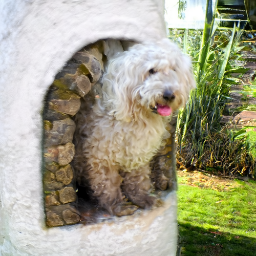}&
        \includegraphics[width=0.1\textwidth]{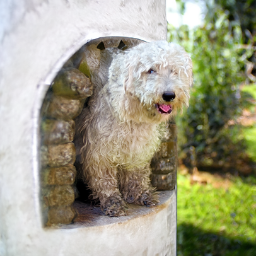}&
        \includegraphics[width=0.1\textwidth]{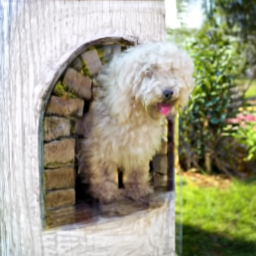}&
        \includegraphics[width=0.1\textwidth]{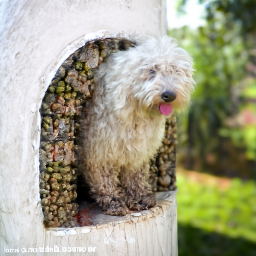}&
        \includegraphics[width=0.1\textwidth]{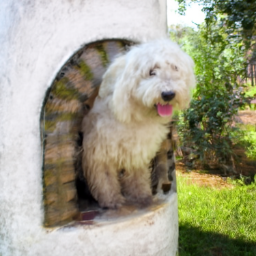}
    \end{tabular} 
    \caption{
    Additional 8x super-resolution results on ImageNet256. We show three different results generated by each method for each input. All results are generated using the same 5 second sampling time.
    }
    \label{apx_fig_supres}
    \vspace{-2.5mm}
\end{figure*}

Figure \ref{apx_fig_inpaint}, \ref{apx_fig_color}, and \ref{apx_fig_supres} show additional qualitative results on the three tasks.
We evaluate LPIPS and PSNR between the generated images and their original images for the three tasks in Table \ref{tab:imputation}. Our test set consists of 200 input images, and each test image will be used to produce 6 samples for each task for the evaluation.

Note that LPIPS and PSNR are not the ideal measurements for this evaluation, but they are a de facto standard for benchmarking. Our goal here is to demonstrate how our technique can generalize to other conditional generation tasks. 
Non-diffusion techniques can not be compared the acceleration effect with ours since they are based on completely different things.  Some comparison of the diffusion technique to other techniques can also be found in \cite{chung2022improving}. 
To achieve state-of-the-art performance on these tasks in terms of quality and speed, other recent techniques may be more suitable, such as improved initialization \citep{chung2022come} and forward-backward repetition \citep{meng2021sdedit}. However, our contributions are orthogonal to these investigations.


\begin{table}[h]
    \centering
    \begin{tabular}{l|cc|cc|cc}
        \toprule
        &\multicolumn{2}{c|}{Inpainting} & \multicolumn{2}{c|}{Colorization} & \multicolumn{2}{c}{Super-resolution}\\
        Methods & LPIPS & PSNR & LPIPS & PSNR & LPIPS & PSNR \\
        \midrule
       DDIM       & 0.17     & 19.52    & 0.28    & 20.40    & 0.46   & 17.88    \\
       PLMS4      & 0.25     & 15.01    & 0.47    & 13.50    & 0.73   & 7.85    \\
       \textbf{LTSP4}      & 0.17     & 19.61    & 0.31    & 19.40    & 0.51   & 16.07    \\
       \textbf{STSP4}      & \textbf{0.16} & \textbf{20.03} & \textbf{0.26} & \textbf{21.27} & \textbf{0.42} & \textbf{19.34}    \\
       \bottomrule
    \end{tabular}
    \caption{Average LPIPS and PSNR scores from different methods on three different tasks.}
    \label{tab:imputation}
\end{table}

\section{Dreambooth Stable Diffusion} 
\label{dreambooth}
Dreambooth \citep{ruiz2022dreambooth} is a technique for fine-tuning a pretrained text-to-image diffusion model on a given set of images.
We discover that, similar to guided diffusion models, Dreambooth on Stable Diffusion (a pretrained text-guided latent-space diffusion) sometimes cannot be used with high-order methods but can be accelerated by our proposed method.
This example demonstrates that our splitting method is effective not only on classifier-guided models but also classifier-free diffusion models.

The guided ODE of a classifier-free model is given by
\begin{equation} \label{eq:dream_guide}
    \frac{ d \bar{x}}{d \sigma} = \bar{\epsilon}_\sigma (\bar{x}|\phi) + s (\bar{\epsilon}_\sigma (\bar{x}|c) - \bar{\epsilon}_\sigma (\bar{x}|\phi) ),
\end{equation}
where $c$ is the input prompt, $\bar{\epsilon}_\sigma (\bar{x}|c)$ is the network output conditioned on the input prompt, and $\bar{\epsilon}_\sigma (\bar{x}|\phi)$ is the network output conditioned on a null label $\phi$.
We can split the guided ODE into two subproblems as follows:
\begin{equation} 
    \frac{dy}{d\sigma} = \bar{\epsilon}_\sigma (y|\phi), \quad 
    \frac{dz}{d\sigma} = s (\bar{\epsilon}_\sigma (z|c) - \bar{\epsilon}_\sigma (z|\phi) ). 
\end{equation}

\begin{figure*}
    \centering
    \setlength\tabcolsep{1.5pt}
    \begin{tabular}{cl}
    \shortstack{\small Number \\of steps} & \qquad\quad\: 20 \qquad\qquad\qquad\: 40 \qquad\qquad\qquad\quad 80 \qquad\qquad\qquad 160  \\
    \shortstack{PLMS4\\\citep{liu2022pseudo}\vspace{0.8cm}} &
    \includegraphics[width=0.8\textwidth]{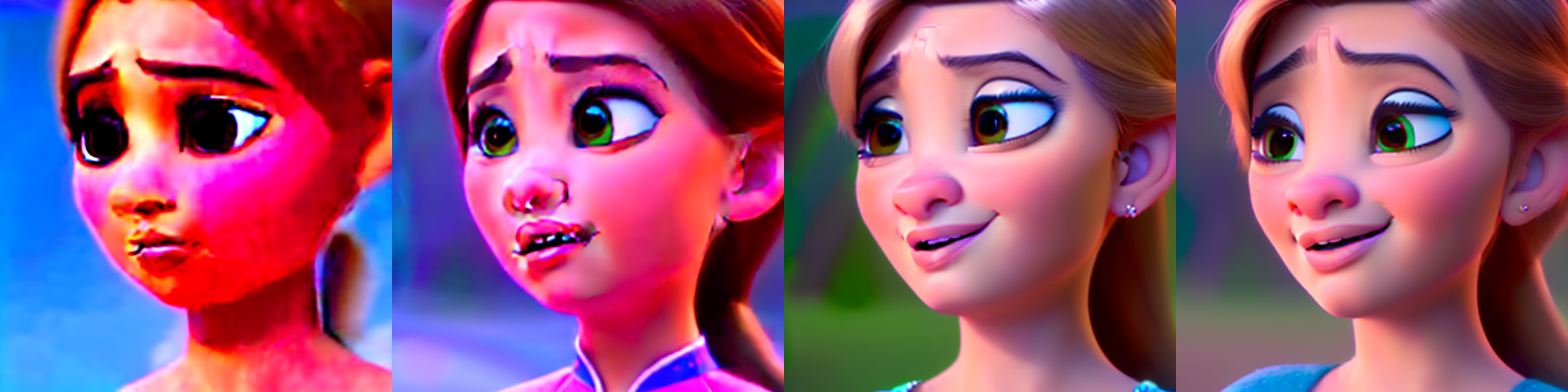} \\
    \shortstack{\textbf{LTSP4}\\(Ours)\vspace{0.8cm}} &
    \includegraphics[width=0.8\textwidth]{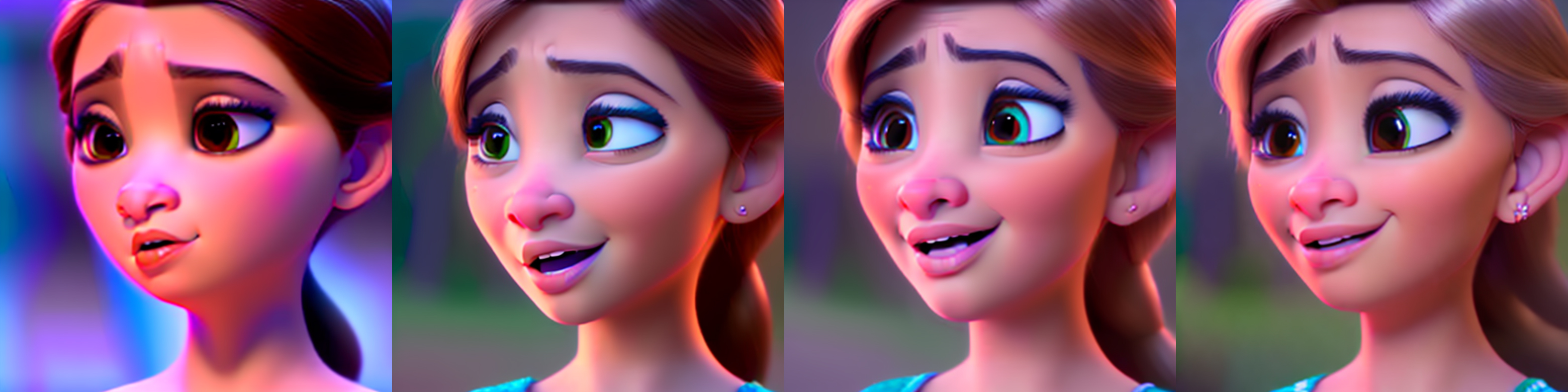}\\
    \shortstack{\textbf{STSP4}\\(Ours)\vspace{0.8cm}} &
    \includegraphics[width=0.8\textwidth]{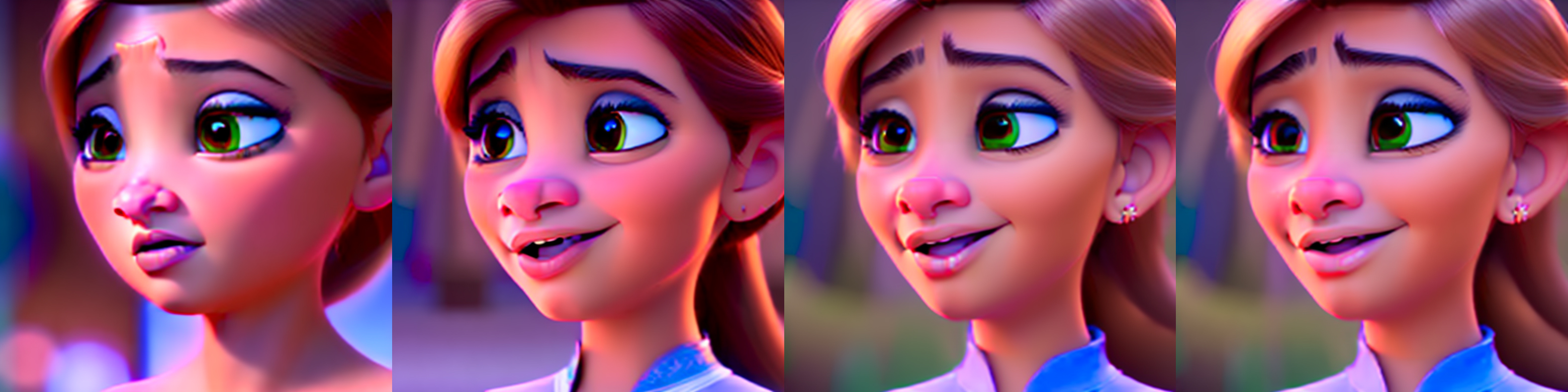} \\
    \end{tabular}
    \caption{
Generated samples from a text-guided Stable Diffusion model fine-tuned on a dataset of screenshots from Disney studio using 20-160 sampling steps. Our splitting technique produces high-quality results in fewer sampling steps. prompt: ``a girl face in modern Disney style''
    }
    \label{fig:sd}
\end{figure*}

We test on ``mo-di-diffusion\footnote{https://huggingface.co/nitrosocke/mo-di-diffusion}'', a Dreambooth Stable Diffusion model that was fine-tuned on a dataset of screenshots from Disney studio. We use the prompt `` a girl face in modern Disney style.'' The result is shown in Figure \ref{fig:sd}.
However, we believe that Dreambooth diffusion models may have different problems from classifier-guided diffusion models and deserve their own in-depth study.

\section{CLIP-Guided Stable Diffusion} 
\label{clip_sd}
\begin{figure}
    \centering
    \setlength\tabcolsep{1.5pt}
    \begin{tabular}{c|ccc}
        \shortstack{\small Stable Diffusion's prompt: ``a ball'' \\
        \small CLIP's prompt: ``a metal sphere''  \vspace{1cm}}
       &
        \includegraphics[width=0.2\textwidth]{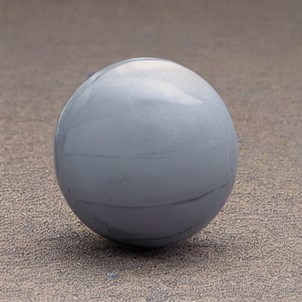}&
        \includegraphics[width=0.2\textwidth]{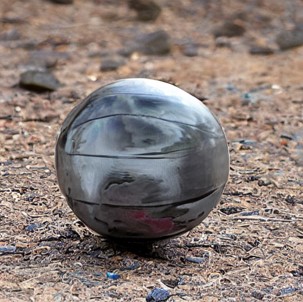}&
        \includegraphics[width=0.2\textwidth]{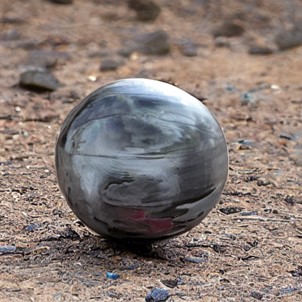}
        \\
        \shortstack{\small Stable Diffusion's prompt: ``a ball'' \\
        \small CLIP's prompt: ``a red apple''  \vspace{1cm}}
       &
        \includegraphics[width=0.2\textwidth]{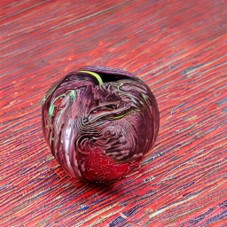}&
        \includegraphics[width=0.2\textwidth]{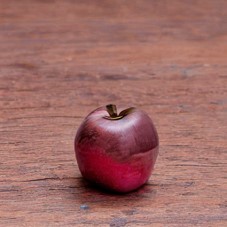}&
        \includegraphics[width=0.2\textwidth]{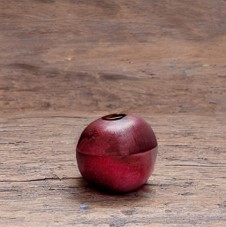}
        \\
          & PLMS4 & \textbf{LTSP4}  & \textbf{STSP4} \\
          & \citep{liu2022pseudo} & (Ours) & (Ours)\\
        &\small (100 steps) & \small  (100 steps) & \small  (55 steps) \\
    \end{tabular}
        \caption{ Text-to-image generation using CLIP-guided Stable Diffusion from different high-order sampling methods with approximately the same sampling time.}
    \label{fig:clip_sd}
\end{figure}

UPainting \citep{li2022upainting} suggest that text-to-image Stable Diffusion quality can be improve using gradient from CLIP model.
This is an example of combining classifier-free and gradient-guided technique to obtain better result.
As we have demonstrated in our paper, adding a gradient term to the diffusion model can cause high-order methods fail to accelerate sampling.
The guided ODE \ref{eq:guide} can be modified as follows:

\begin{equation} \label{eq:clip_guide}
    \frac{ d \bar{x}}{d \sigma} = \bar{\epsilon}_\sigma (\bar{x}|\phi) + s (\bar{\epsilon}_\sigma (\bar{x}|c) - \bar{\epsilon}_\sigma (\bar{x}|\phi) ) - \lambda \nabla_{\bar{x}}  (f_\text{img} (\bar{x}) \cdot f_\text{txt} (a)),
\end{equation}
where $f_\text{img} (\bar{x})$ is CLIP's image encoder output and $f_\text{txt} (a)$ is the output of CLIP's text encoder.
We can apply our method by splitting the Equation \ref{eq:clip_guide} into two subproblems by
\begin{equation} 
    \frac{dy}{d\sigma} = \bar{\epsilon}_\sigma (y|\phi) + s (\bar{\epsilon}_\sigma (y|c) - \bar{\epsilon}_\sigma (y|\phi) ), \quad 
    \frac{dz}{d\sigma} = - \lambda \nabla_{z}  (f_\text{img} (z) \cdot f_\text{txt} (a)). 
\end{equation}
We show sample images using different numerical methods in Figure \ref{fig:clip_sd}. Our method produces high-quality results in fewer sampling steps.

\section{Convergence Orders of Methods} 
In this section, we show the convergence order of Lie-Trotter and Strang splitting methods. Suppose the differential equation we want to solve is defined by
\begin{align}
    \frac{dx}{dt} = f_0 (x) = f_1 (x) + f_2(x), 
\end{align}
where $f_1, f_2$ are assumed to be differentiable.
We define $\Phi_{\Delta t,f_i}$ as a mapping solution of an ODE $\frac{dx}{dt} = f_i (x,t)$ in an interval $[t_0, t_0 + \Delta t]$.
Noting that $\Phi_{0,f_i}(x) = x$ and $\frac{d}{dt} \Phi_{\Delta t,f_i}(x) = f_i(x)$.

By perform a Taylor expansion, we have
\begin{align} \label{expan}
    \Phi_{\Delta t,f_i}(x) = x + \Delta t f_i (x) + \frac{(\Delta t)^2}{2} f'_i(x) f_i(x) +  \mathcal{O}(\Delta t^3).
\end{align}

\textbf{Lie-Trotter splitting}

A single step of the Lie-Trotter splitting method can be expressed as $\Phi_{\Delta t,f_2}(\Phi_{\Delta t,f_1}(x))$. Applying the expansion of Equation \ref{expan} to this formulation gives:
\begin{align}
    \Phi_{\Delta t,f_2}(\Phi_{\Delta t,f_1}(x)) &= [x + \Delta t f_1 (x) +  \mathcal{O}(\Delta t^2)] + \Delta t f_2 [x + \Delta t f_1 (x) +  \mathcal{O}(\Delta t^2)] +  \mathcal{O}(\Delta t^2) \nonumber\\
    &= x + \Delta t f_1 (x) + \Delta t f_2 (x) + \mathcal{O}(\Delta t^2)  \nonumber\\
    &= \Phi_{\Delta t,f_0}(x) + \mathcal{O}(\Delta t^2).
\end{align}
Hence, the Lie-Trotter splitting method is of first-order.

\textbf{Strang splitting}

Consider a single step of the Strang splitting, which is $\Phi_{\Delta t/2,f_2}(\Phi_{\Delta t,f_1}(\Phi_{\Delta t/2,f_2}(x)))$. To show the second-order accuracy of the Strange splitting, we first expand the two inner operators.
\begin{align}
    \Phi_{\Delta t,f_1}(\Phi_{\Delta t/2,f_2}(x)) &= [x + \frac{\Delta t}{2} f_2 (x) + \frac{(\Delta t)^2}{2^2 2!} f'_2(x) f_2(x)  +  \mathcal{O}(\Delta t^3)] \nonumber \\
     &+ (\Delta t) f_1 [x + \frac{\Delta t}{2} f_2 (x)  +  \mathcal{O}(\Delta t^2)]   \nonumber \\
     &+ \frac{(\Delta t)^2}{2} f'_1 [x +   \mathcal{O}(\Delta t)] f_1 [x +   \mathcal{O}(\Delta t)] +  \mathcal{O}(\Delta t^3) \\
    &=x + \frac{\Delta t}{2} f_2 (x) + \frac{(\Delta t)^2}{2^2 2!}f'_2(x)f_2(x) \nonumber \\
    &+ (\Delta t) f_1(x) + \frac{(\Delta t)^2}{2} f'_1 (x)f_2(x) \nonumber \\
    & + \frac{(\Delta t)^2}{2} f'_1(x) f_1(x) +\mathcal{O}(\Delta t^3)
\end{align}
Consequently, 
\begin{align}
    \Phi_{\frac{\Delta t}{2},f_2}(\Phi_{\Delta t,f_1}(\Phi_{\frac{\Delta t}{2},f_2}(x))) 
    &= x + (\Delta t) f_1 (x) + \frac{\Delta t}{2}f_2(x) \nonumber \\
    &+ \frac{(\Delta t)^2}{2^2 2!} f'_2(x)f_2(x) + \frac{(\Delta t)^2}{2} f'_1(x)f_2(x) + \frac{(\Delta t)^2}{2!}f'_1(x)f_1(x) \nonumber \\
    & + \frac{\Delta t}{2} f_2 [x + (\Delta t) f_1 (x) + \frac{\Delta t}{2}f_2(x) + \mathcal{O}(\Delta t^2)] \nonumber \\
    & + \frac{(\Delta t)^2}{2^2 2!}f'_2[x+\mathcal{O}(\Delta t)]f_2[x+\mathcal{O}(\Delta t)] + \mathcal{O}(\Delta t^3)
\end{align}

\begin{align}
    \Phi_{\frac{\Delta t}{2},f_2}(\Phi_{\Delta t,f_1}(\Phi_{\frac{\Delta t}{2},f_2}(x))) 
    &= x + (\Delta t) f_1 (x) + (\Delta t) f_2 (x) \nonumber \\
    &+ \frac{(\Delta t)^2}{2!}[f'_1(x)f_1(x) + f'_1(x)f_2(x)+f'_2(x)f_1(x)+f'_2(x)f_2(x)]  \nonumber \\
    &+ \mathcal{O}(\Delta t^3) \\
    &=x + (\Delta t) f_0(x) +\frac{(\Delta t)^2}{2!} f'_0(x) f_0(x) + \mathcal{O}(\Delta t^3) \\
    &= \Phi_{\Delta t, f_0}(x) + \mathcal{O}(\Delta t^3)
\end{align}
Therefore, the Strang splitting method's convergence rate is of second-order.

\section{Toy Example} 
In this section, we create a toy example to demonstrate how high-order numerical methods can become unstable on a certain class of ODE problems (Stiff equation) despite involving no neural networks. Let us define the following ODE:
\begin{align} \label{eq:toy_ode}
    \frac{dx}{dt} = \epsilon(x) + s \cdot g(x), \qquad x(0) = \begin{bmatrix} 1 \\ 0\end{bmatrix},
\end{align}
where $s$ is a scaling parameter and

\begin{align}
    \epsilon \left( \begin{bmatrix} x_1 \\ x_2\end{bmatrix} \right) = \begin{bmatrix} 0 & 1 \\ -1 & -2\end{bmatrix}\begin{bmatrix} x_1 \\ x_2\end{bmatrix}, \quad
    g \left( \begin{bmatrix} x_1 \\ x_2\end{bmatrix} \right)        = \begin{bmatrix} 0 & 0 \\ -1 & -1\end{bmatrix}\begin{bmatrix} x_1 \\ x_2\end{bmatrix}.
\end{align}

\begin{figure}
    \centering
    \setlength\tabcolsep{1.5pt}
    \begin{tabular}{cccc}
        
        \shortstack{$s=3$\vspace{1.8cm}}
       &
        \includegraphics[width=0.30\textwidth]{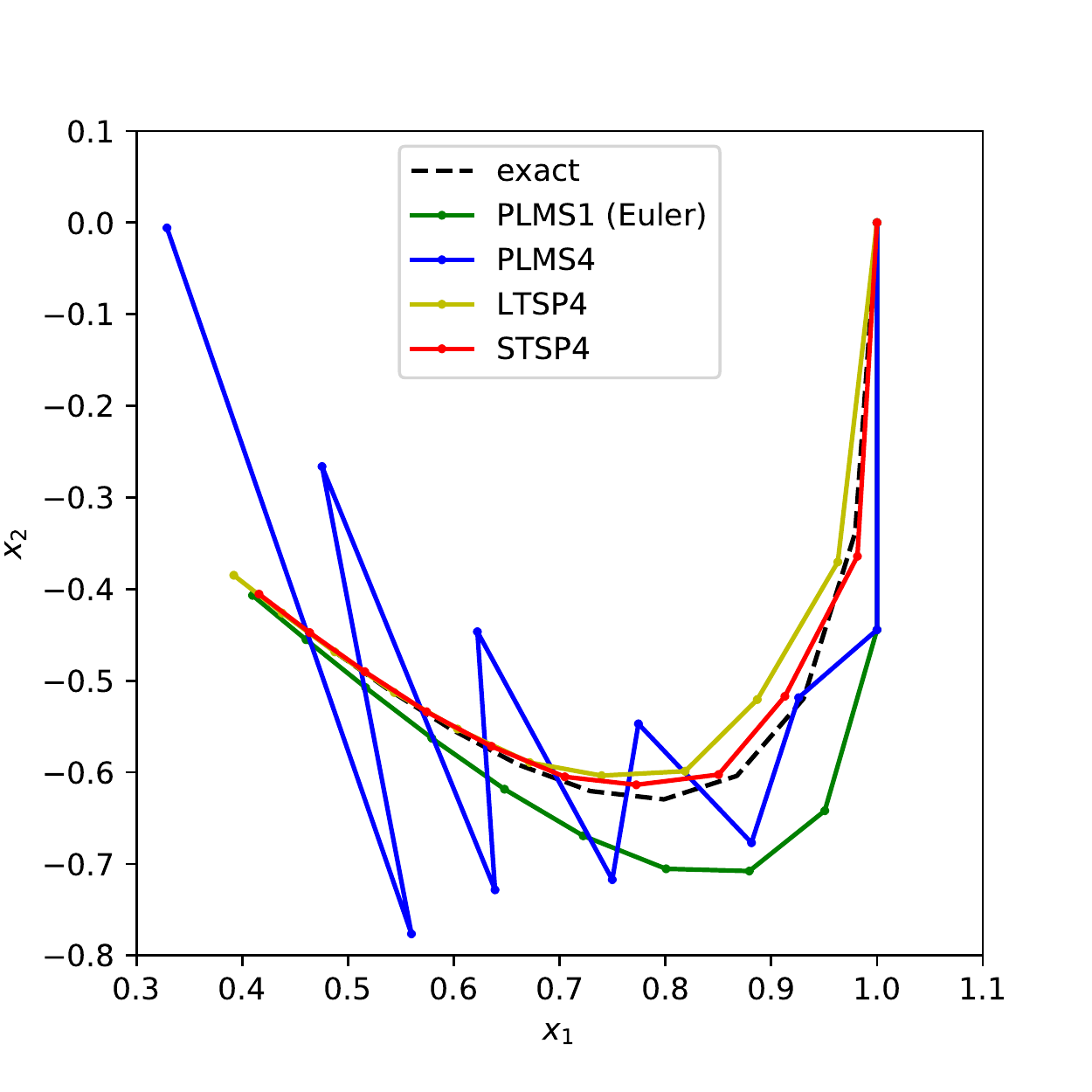}&
        \includegraphics[width=0.30\textwidth]{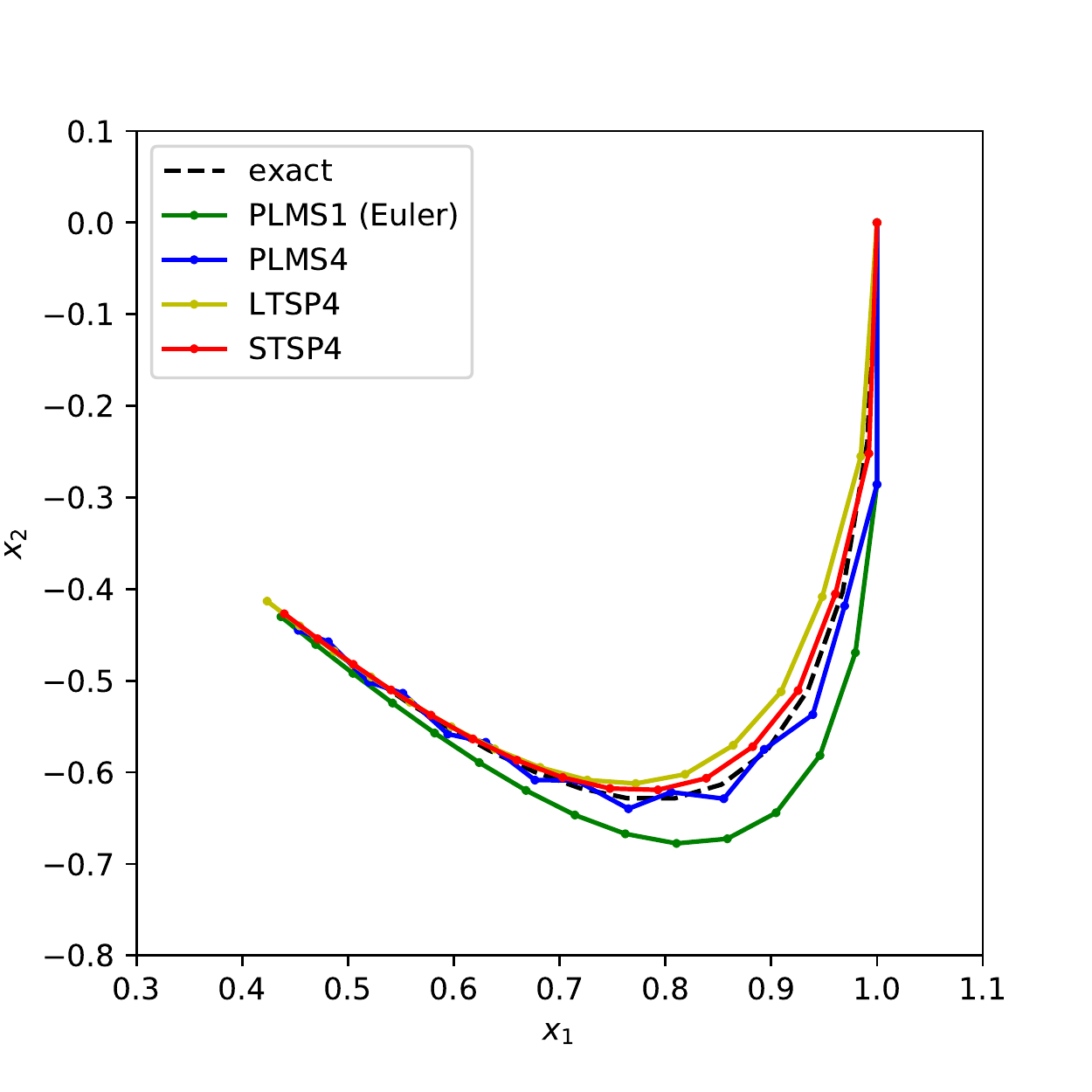}&
        \includegraphics[width=0.30\textwidth]{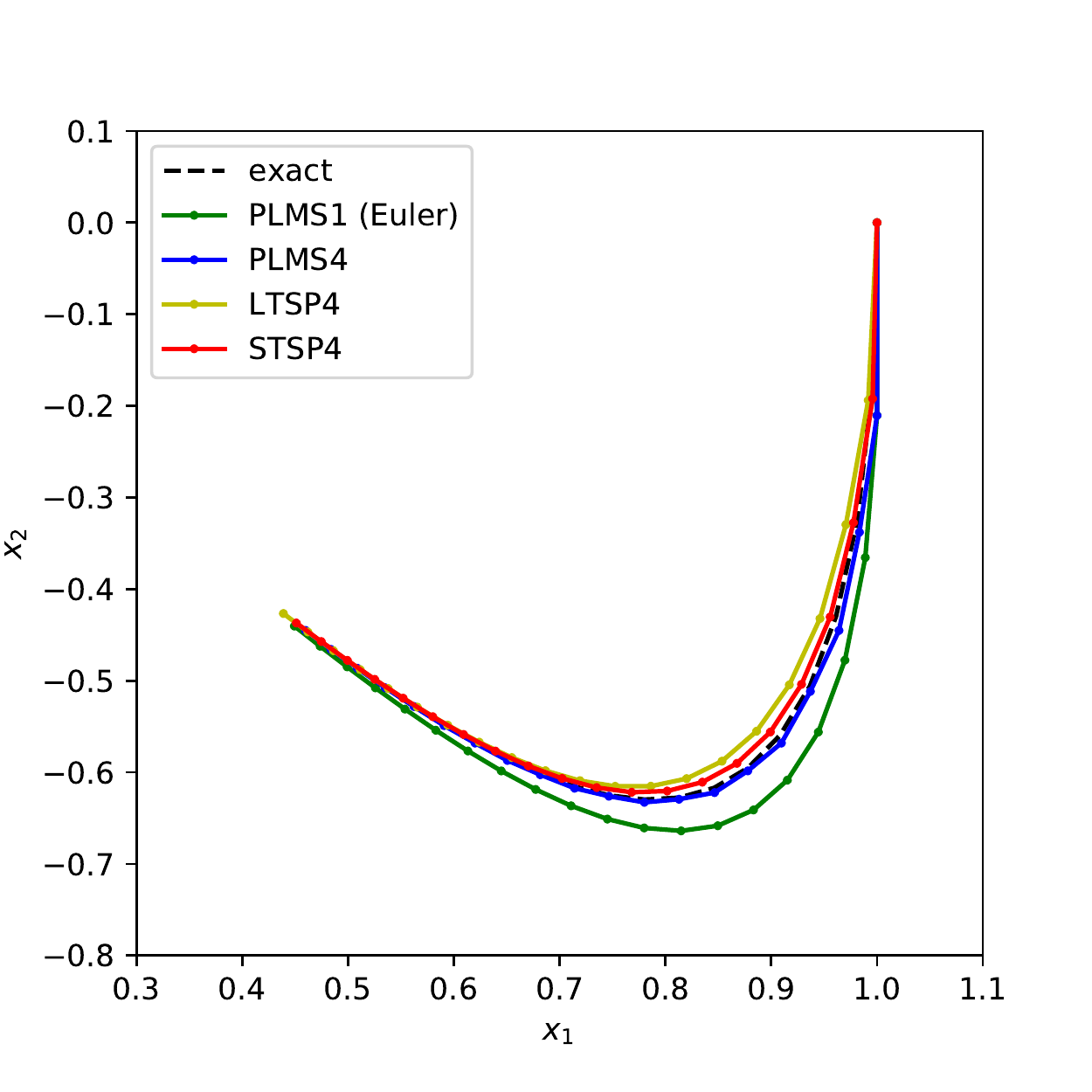}
        \\
        \shortstack{$s=5$\vspace{1.8cm}}
       &
        \includegraphics[width=0.30\textwidth]{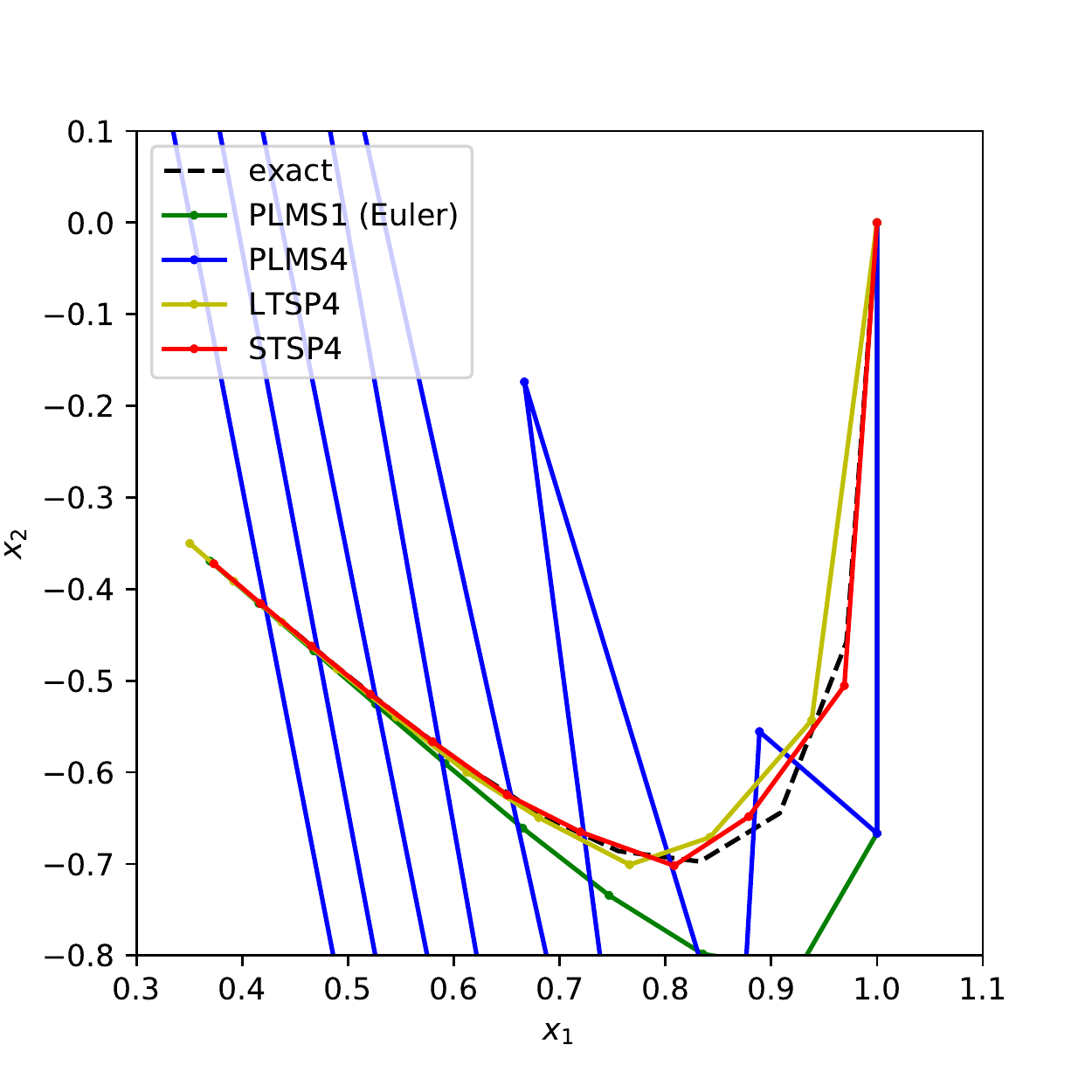}&
        \includegraphics[width=0.30\textwidth]{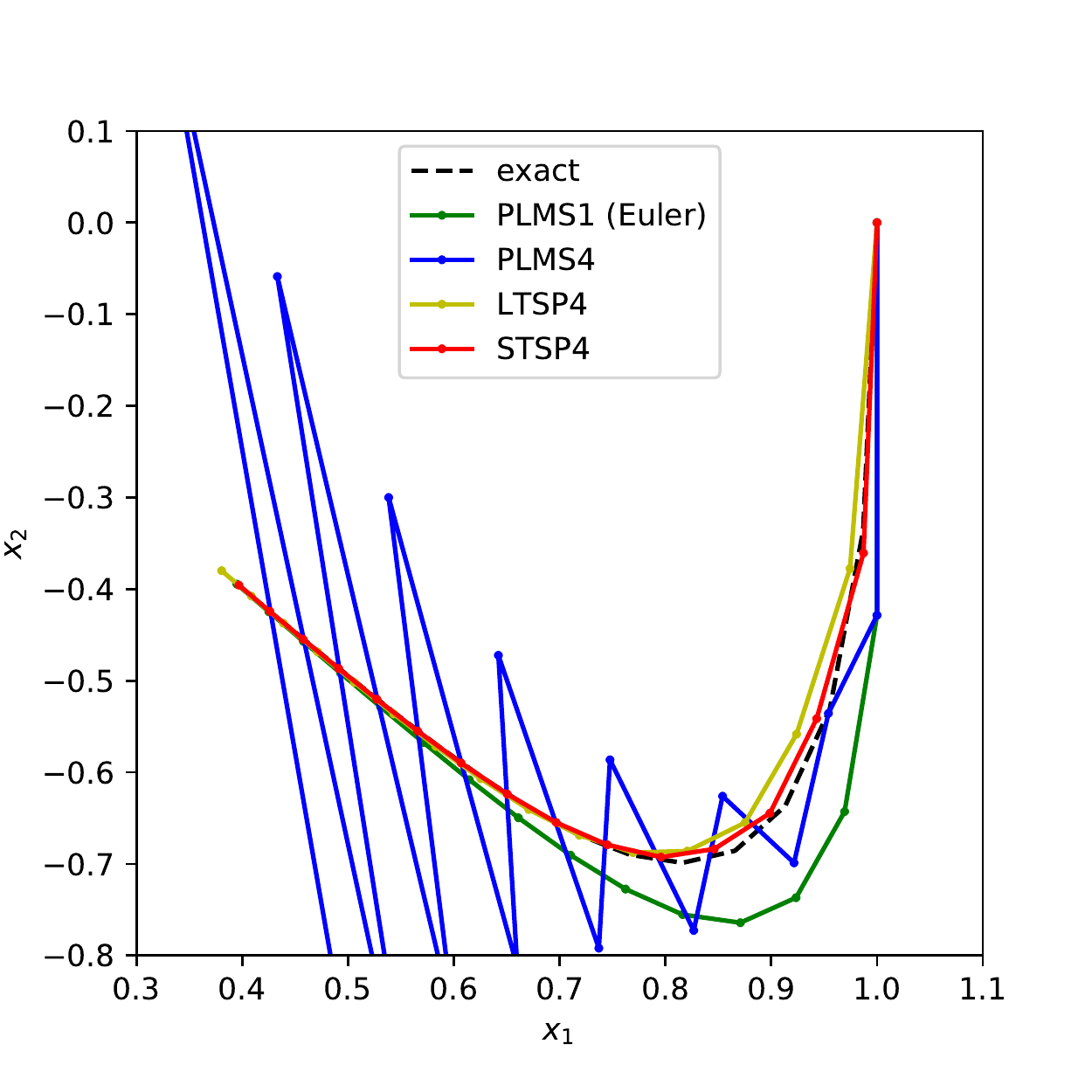}&
        \includegraphics[width=0.30\textwidth]{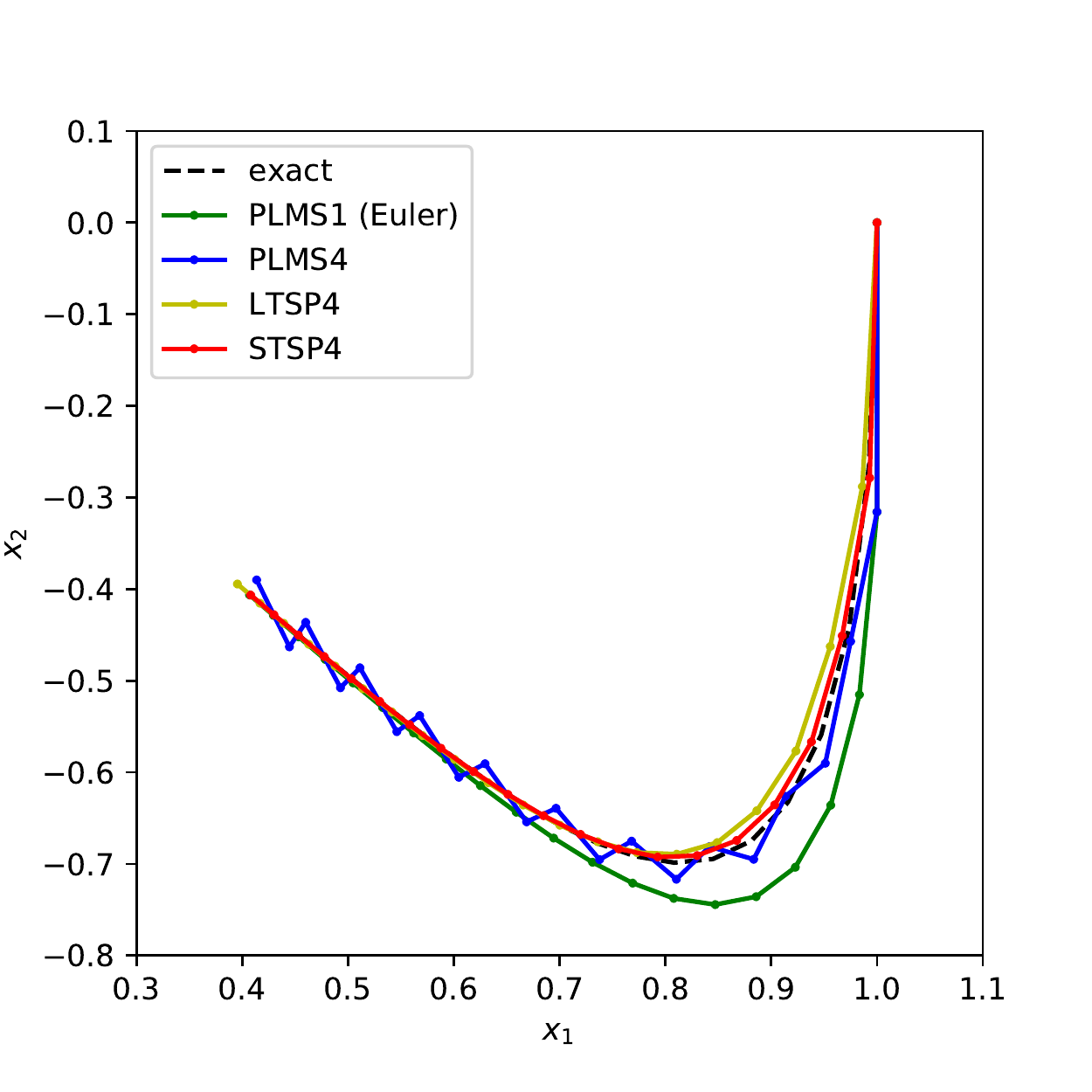}
        \\
        & 10 steps &  15 steps &   20 steps \\
    \end{tabular}
        \caption{ 
        Solution trajectories of different numerical methods on a toy ODE problem using different numbers of steps. Non-splitting methods especially  PLMS4  are more likely to fail to converge to the exact solution when the number of steps is reduced. 
        }
    \label{fig:toy}
\end{figure}

Figure \ref{fig:toy} depicts solution trajectories of various numerical methods
We can observe that the PLMS4's trajectory can run far from the exact solution than the Euler's method or the splitting methods, LSTP4 and STSP4. The exact solution of Equation \ref{eq:toy_ode} is

\begin{align}
    x(t) = \frac{1}{s}\begin{bmatrix} -1 \\ s+1 \end{bmatrix} e^{-(s+1)t} + \frac{1}{s}\begin{bmatrix} s+1 \\ -s-1 \end{bmatrix} e^{-t}.
\end{align}

When $s$ increases, the term $e^{-(s+1)t}$ decays to zero more rapidly than the term $e^{-t}$. When the two terms behave differently, classical high-order numerical methods tend to perform poorly unless they employ a very small step size.

\section{Stability Analysis} 
In this section, we further analyze numerical solutions using stability analysis. We compute the lowest number of steps before the numerical solution is guaranteed to diverge by theory and visualize the solution trajectories using different numbers of steps to empirically support the theory.
We only focus on Euler and other second-order numerical methods in this part.

One way to analyze numerical methods that solve Equation \ref{eq:toy_ode} is to evaluate them with a test equation that leads to the solution $y(t)=e^{-(s+1)t}$, such as 

\begin{equation} \label{test_eq}
    y'=-(s+1)y.
\end{equation}
Note that the solution $y(t) \rightarrow 0$ as $t \rightarrow \infty$.

\textbf{Euler's Method:} Applying the Euler's method to Equation \ref{test_eq} yields:
\begin{align*}
    y_{n+1} = y_n - \Delta t (s+1) y_n = (1-\Delta t (s+1))y_n.
\end{align*}
After solving this recurrence relation, we have $y_n = (1-\Delta t (s+1))^n y_0$. The condition for the numerical solution $y_n \rightarrow 0$ as $n \rightarrow \infty$ is equivalent to $|1-\Delta t (s+1) |< 1$ or 

\begin{align*}
    -1 < 1 - &\Delta t (s+1) < 1 ,\\
    2 > \Delta t& (s+1) > 0 .
\end{align*}

Let us substitute $\Delta t  =  1/N$, where $N$ is the number of steps. Now, we can conclude that if $N$ is lower than $\frac{s+1}{2}$, the solution of Euler's method in Equation \ref{test_eq} diverges from the exact solution.

\textbf{PLMS2:} Consider a second-order linear multistep method on the same test Equation \ref{test_eq}:
\begin{align}
    y_{n+1} &= y_n +\Delta t \left(-\frac{3}{2}(s+1)y_n + \frac{1}{2}y_{n-1}(s+1)\right) \\
            &= \left(1-\Delta t \frac{3}{2} (s+1)\right)y_n +\Delta t\frac{1}{2} (s+1).
\end{align}
After solving the linear recurrence relation, we obtain
\begin{gather}
    y_{n} = a_1 r_1^n + a_2 r_2^n,\\
\text{where } r_1 = \frac{1}{2}\left(1-\frac{3}{2}\Delta t (s+1) + \sqrt{1 - \Delta t (s+1) + \frac{9}{4}(\Delta t)^2 (s+1)^2} \right),\\  \text{and } r_2 = \frac{1}{2}\left(1-\frac{3}{2}\Delta t (s+1) - \sqrt{1 - \Delta t (s+1) + \frac{9}{4}(\Delta t)^2 (s+1)^2} \right).
\end{gather}

The numerical solution $y_n \rightarrow 0$ as $n \rightarrow \infty$ when both $|r_1|<1$ and $|r_2|<1$, which is equivalent to
\begin{align} \label{condi_plms2}
    \left|\frac{1}{2}\left(1-\frac{3}{2} \frac{(s+1)}{N} \pm \sqrt{1 - \frac{(s+1)}{N}  + \frac{9}{4}\left(\frac{(s+1)}{N}\right)^2 } \right)\right| < 1.
\end{align}

In Table \ref{tab:theory}, we report the lowest number $N$ for each $s$ before Inequality \ref{condi_plms2} is not satisfied. In other words, if the number of steps is below the lowest number $N$ in the table, the solution of the method in Equation \ref{test_eq} is guaranteed to diverge from the exact solution. The analysis of the higher-order methods can be done in a similar fashion.

\textbf{LTSP2:} We analyze the Lie-Trotter splitting method similarly. In this case, the test Equation \ref{test_eq} needs to also be split into
\begin{align}
    \hat{y}' =& - \hat{y}, \label{test_sp1} \\ 
    \tilde{y}' =& - s \tilde{y}. \label{test_sp2}
\end{align}
Let us apply the second order linear multistep method (PLMS2) in Equaiton \ref{test_sp1} and Euler's method (PLMS1) on Equation \ref{test_sp2}. We have
\begin{align}
    \hat{y}_{n+1} = \hat{y}_n -\Delta t \left( \frac{3}{2}\hat{y}_n - \frac{1}{2} \hat{y}_{n-1}\right), \qquad \tilde{y}_{n+1} = \tilde{y}_n -\Delta t s \tilde{y}_n.
\end{align}

Thus, a single combining step of LTSP2 can be formulated by
\begin{align}
    y_{n+1} = (1-s\Delta t) \left(\left(1-\frac{3}{2}\Delta t\right)y_n + \frac{\Delta t}{2} y_{n-1}\right).
\end{align}
Similar to the above, we solve the linear recurrence relation and obtain the following condition
{\small
\begin{align} \label{condi_ltsp2}
    \left|\frac{1}{2}\left( \left(1-\frac{s}{N}\right) \left(1-\frac{3}{2}\frac{s}{N}\right) \pm \sqrt{\left(1-\frac{s}{N}\right)^2 \left(1-\frac{3}{2}\frac{s}{N}\right)^2  + \frac{2}{N}\left(1-\frac{s}{N}\right) } \right)\right| < 1.
\end{align}
}%

 We report the lowest integer number $N$ for each $s$ before Inequality \ref{condi_ltsp2} is not satisfied in Table \ref{tab:theory}.
 
 \textbf{STSP2:} We analyze the Strang splitting method by splitting the test Equation \ref{test_eq} into
 \begin{align}
    \bar{y}' =& - s \bar{y} \label{test_st1} \\
    \hat{y}' =& - \hat{y} \label{test_st2} \\ 
    \tilde{y}' =& - s \tilde{y} \label{test_st3}
\end{align}
We apply the second-order linear multistep method (PLMS2) in Equaiton \ref{test_st2} and Euler's method on Equation \ref{test_st1} and \ref{test_st3}.
 \begin{align}
    \bar{y}_{n+1} =& \left(1 - \frac{\Delta t}{2} s\right) \bar{y}_n  \label{numer_st1}\\
    \hat{y}_{n+1} =& \left(1 - \frac{3}{2} \Delta t \right) \hat{y}_n + \frac{\Delta t}{2}  \hat{y}_{n-1} \label{numer_st2}\\ 
    \tilde{y}_{n+1} =& \left(1 - \frac{\Delta t}{2} s\right) \tilde{y}_n  \label{numer_st3}
\end{align}
We combine Equation \ref{numer_st1}-\ref{numer_st3} into
\begin{align}
    y_{n+1} = \left(1 - \frac{s}{2N}\right)^2\left(1 - \frac{3}{2N}\right)y_n + \frac{1}{2N}\left(1 - \frac{s}{2N}\right)^2y_{n-1}.
\end{align}

After solving the linear recurrence relation, we obtain the following condition
\begin{align} \label{condi_stsp2}
    \left|\frac{1}{2} \left(b \pm \sqrt{b^2 + \frac{2}{N}c}\right)\right| < 1,
\end{align}
where $b=\left(1-\frac{s}{2N}\right)^2\left(1-\frac{3}{2N}\right)$ and $c = \left(1-\frac{s}{2N}\right)^2$. In Table \ref{tab:theory}, we report the lowest number of steps $N$ for each $s$ before Inequality \ref{condi_stsp2} is not satisfied.

\begin{figure}
    \centering
    \setlength\tabcolsep{1.5pt}
    \begin{tabular}{cccc}
        
        \shortstack{$s=10$\vspace{1.8cm}}
       &
        \includegraphics[width=0.30\textwidth]{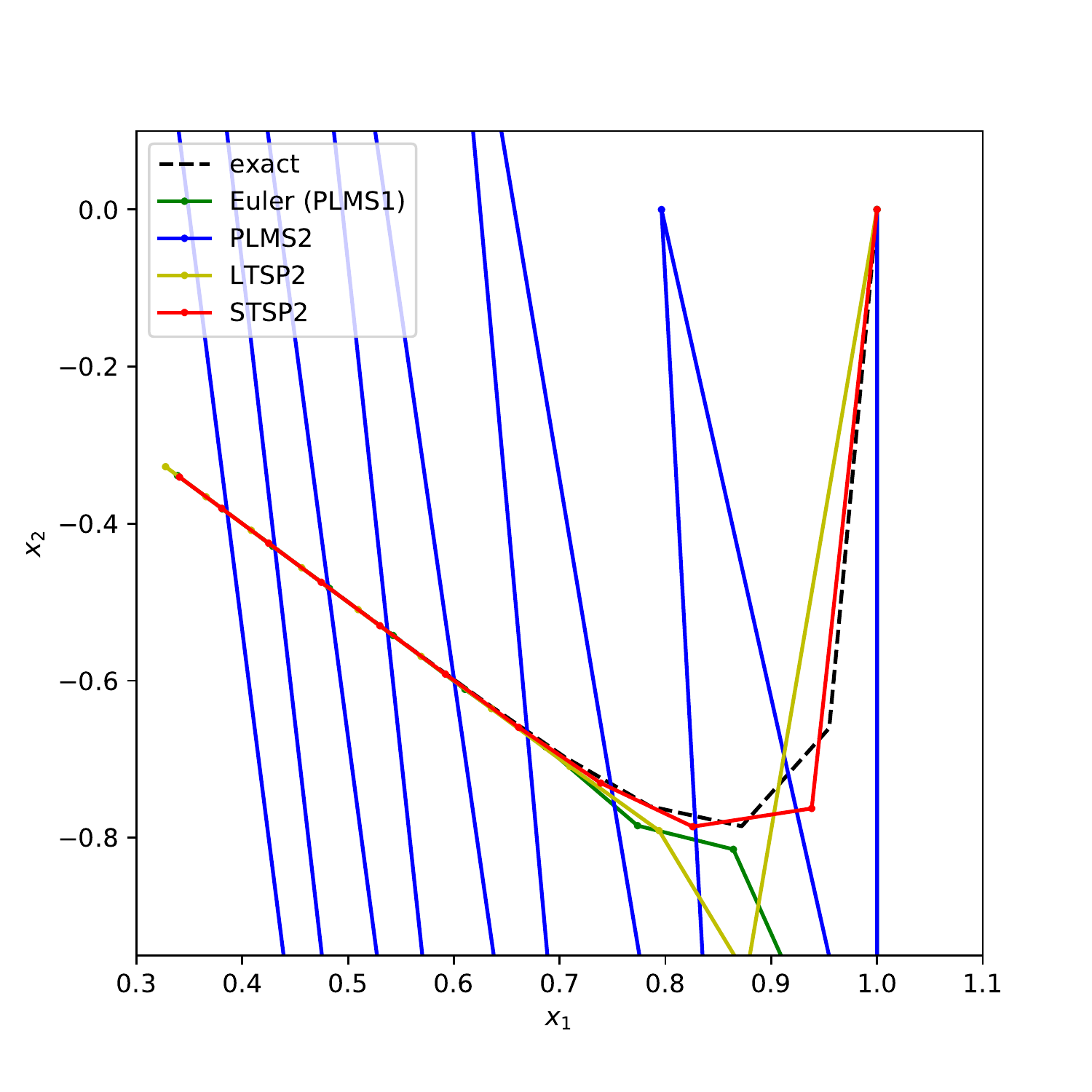}&
        \includegraphics[width=0.30\textwidth]{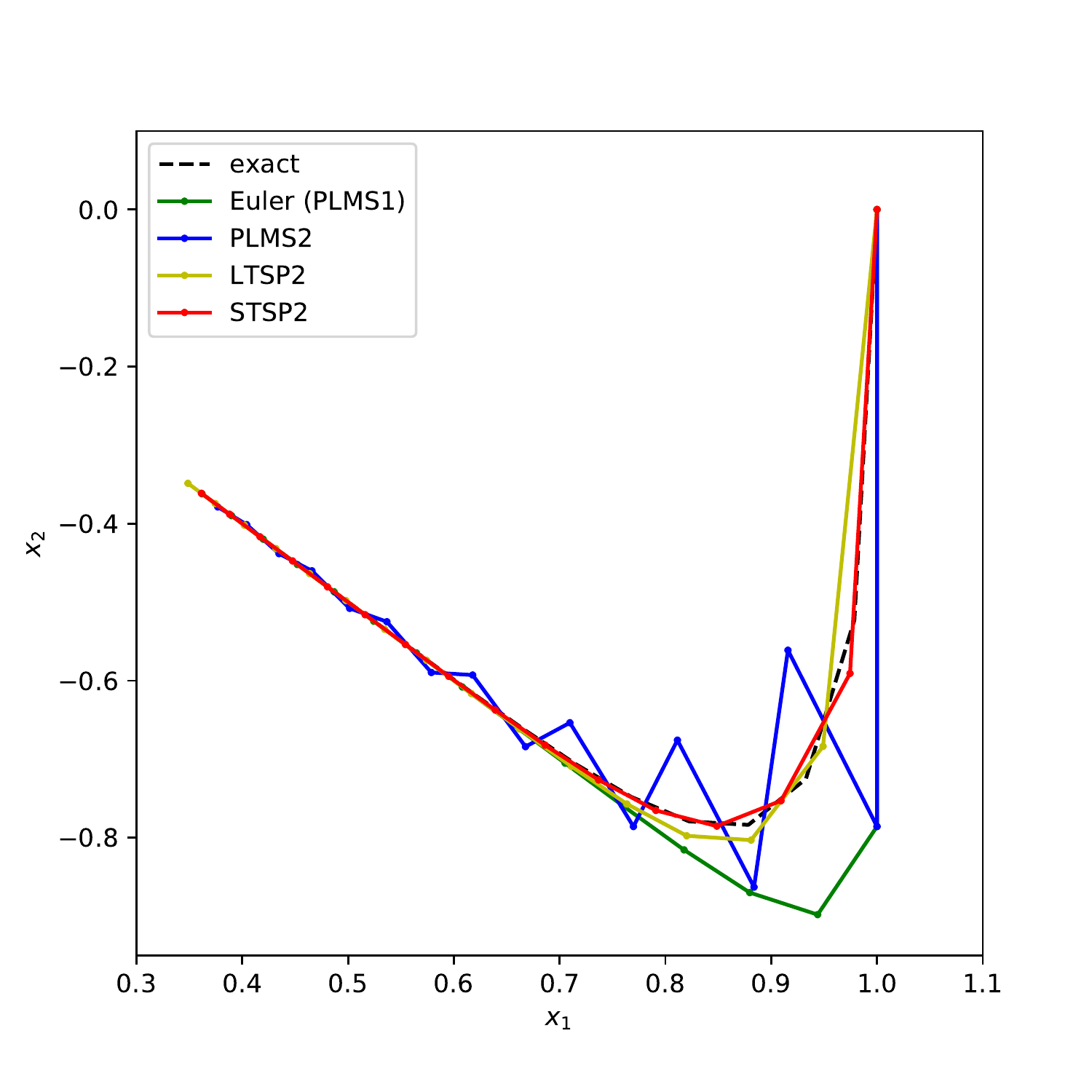}&
        \includegraphics[width=0.30\textwidth]{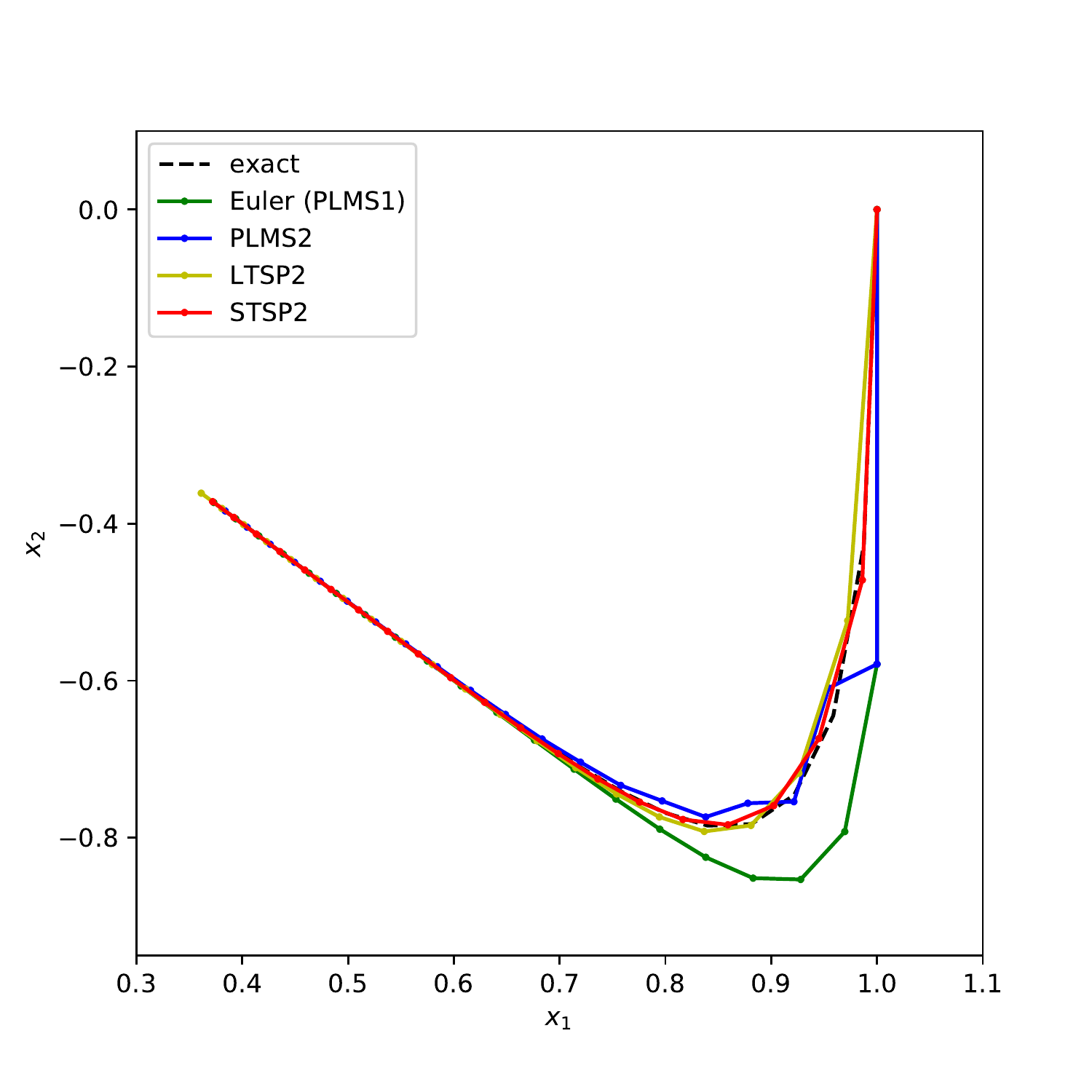}  \\
        
        \shortstack{$s=15$\vspace{1.8cm}}
       &
        \includegraphics[width=0.30\textwidth]{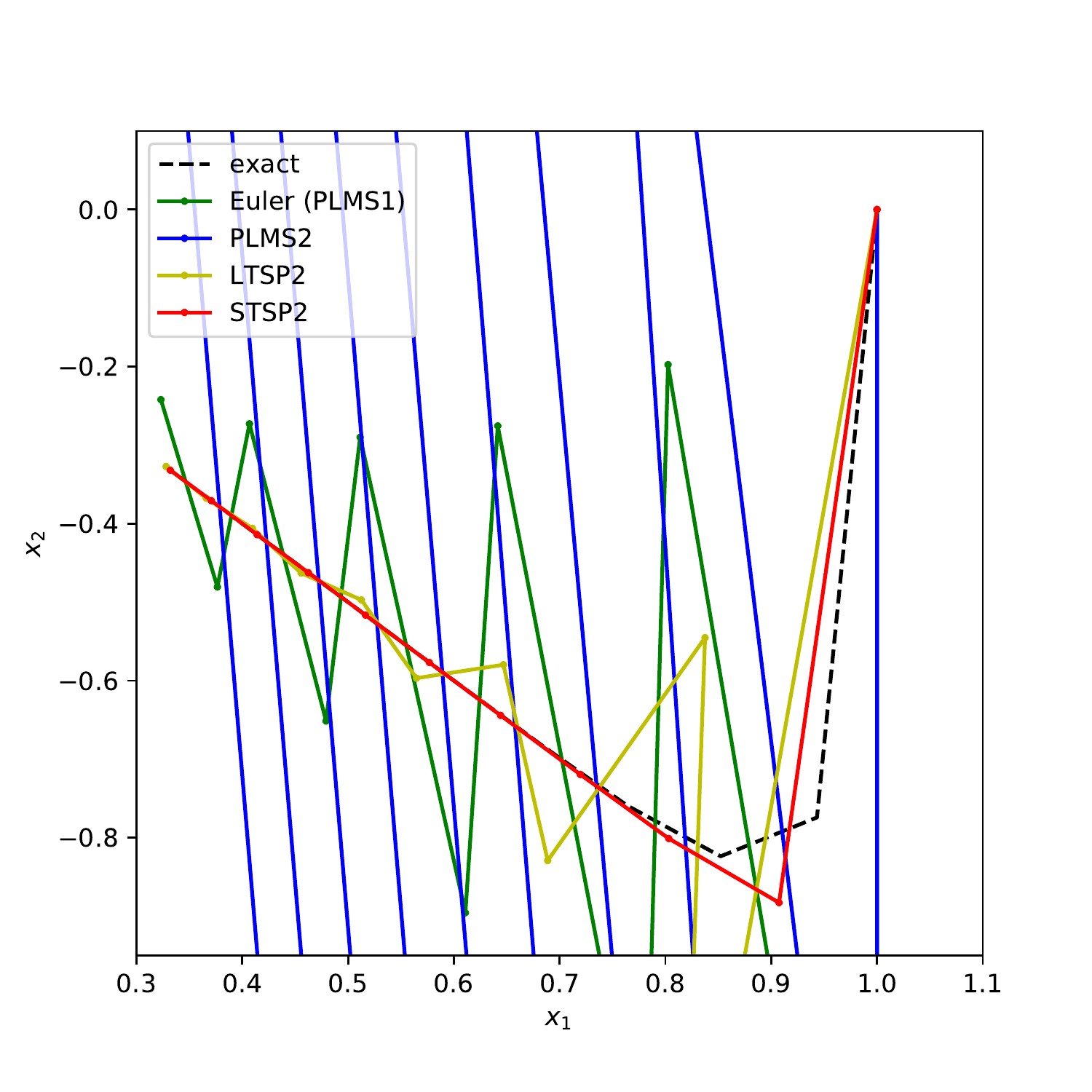}&
        \includegraphics[width=0.30\textwidth]{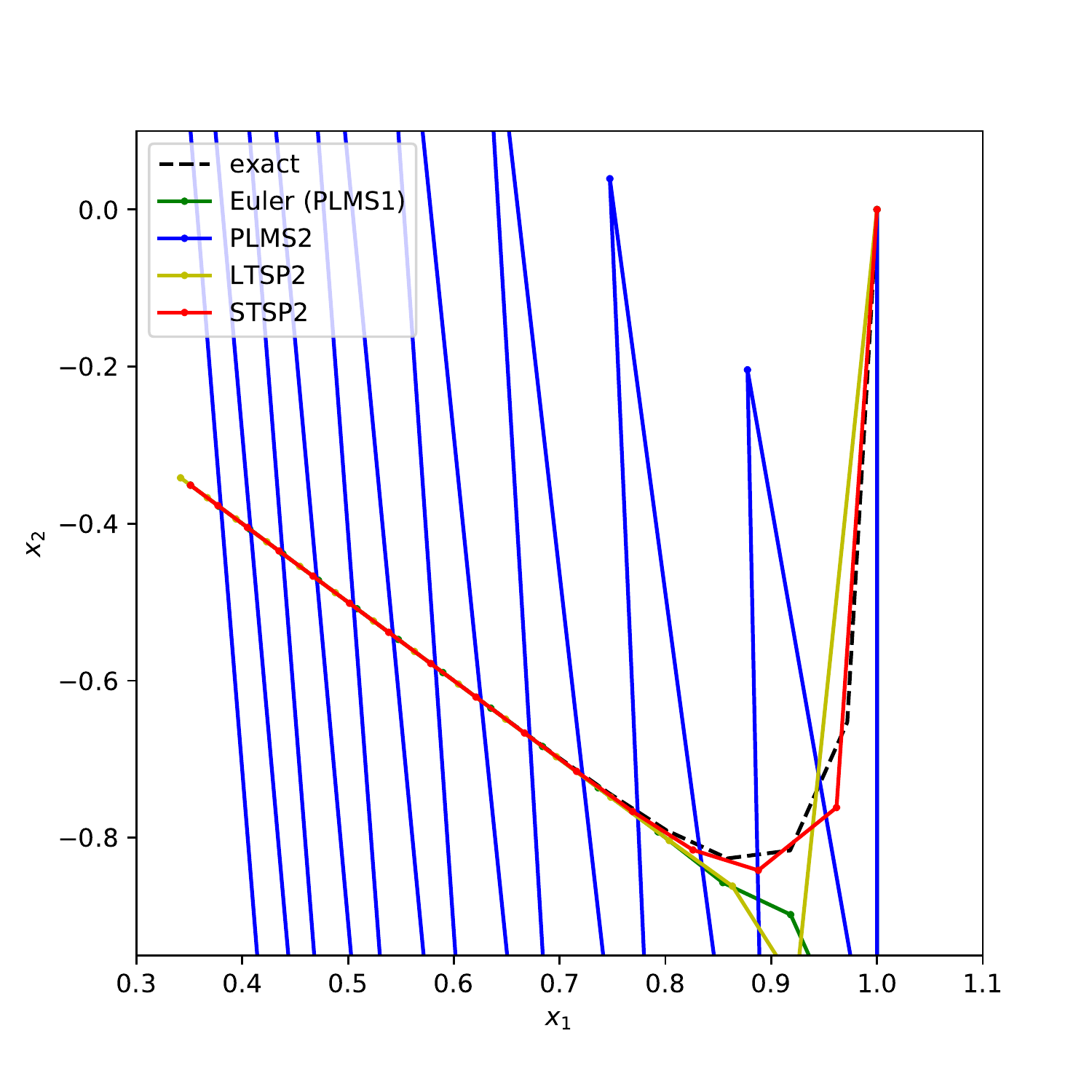}&
        \includegraphics[width=0.30\textwidth]{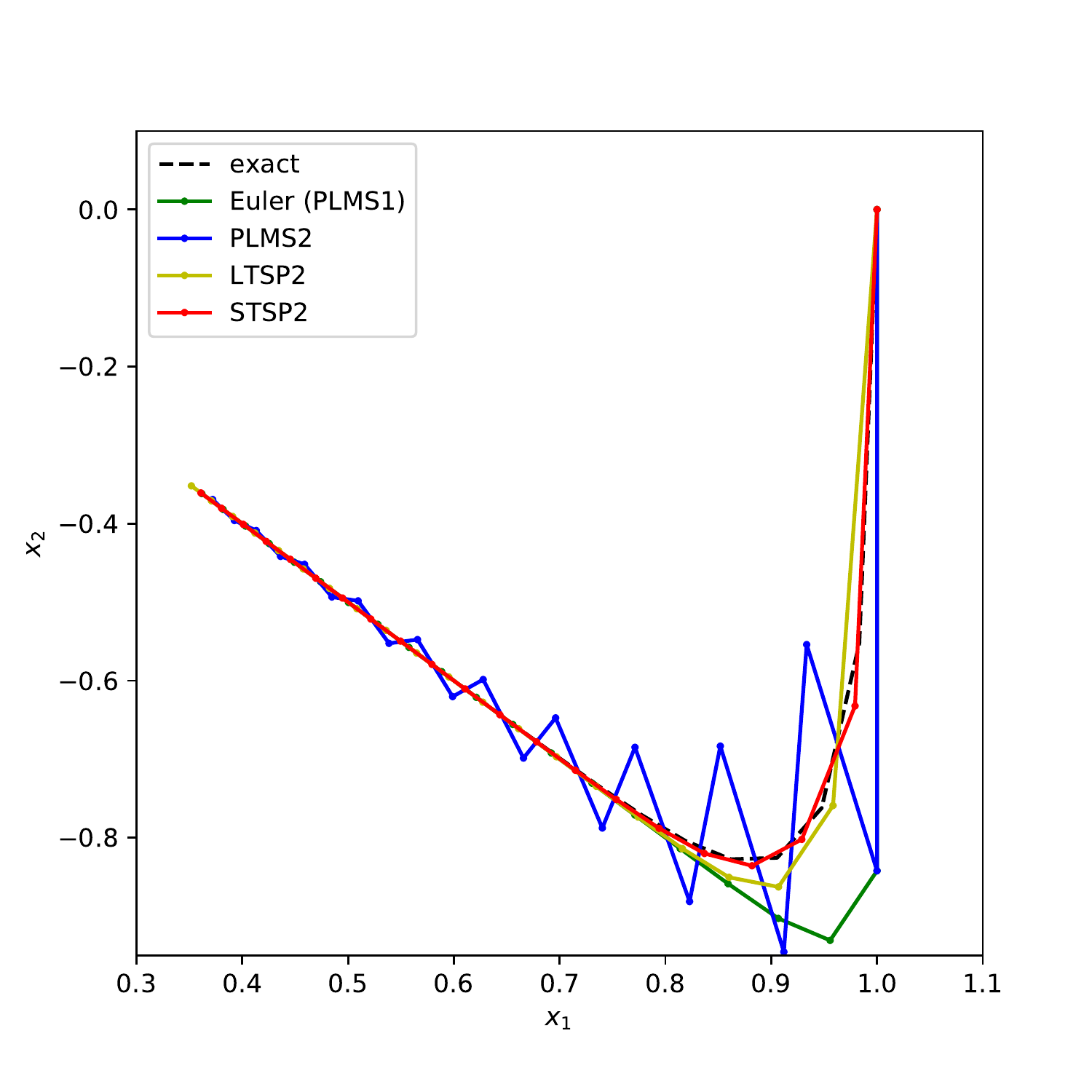}  \\
        
        \shortstack{$s=20$\vspace{1.8cm}}
       &
        \includegraphics[width=0.30\textwidth]{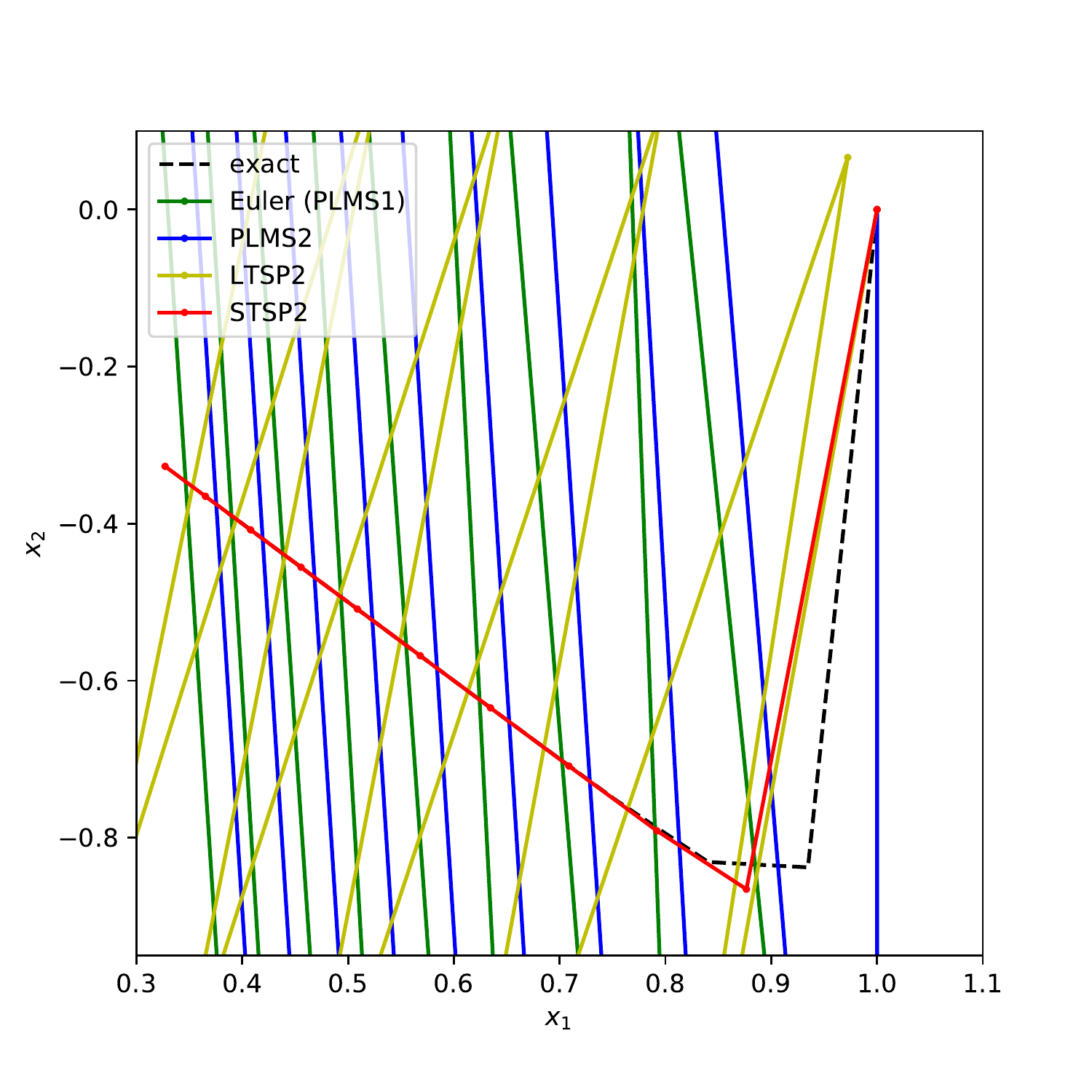}&
        \includegraphics[width=0.30\textwidth]{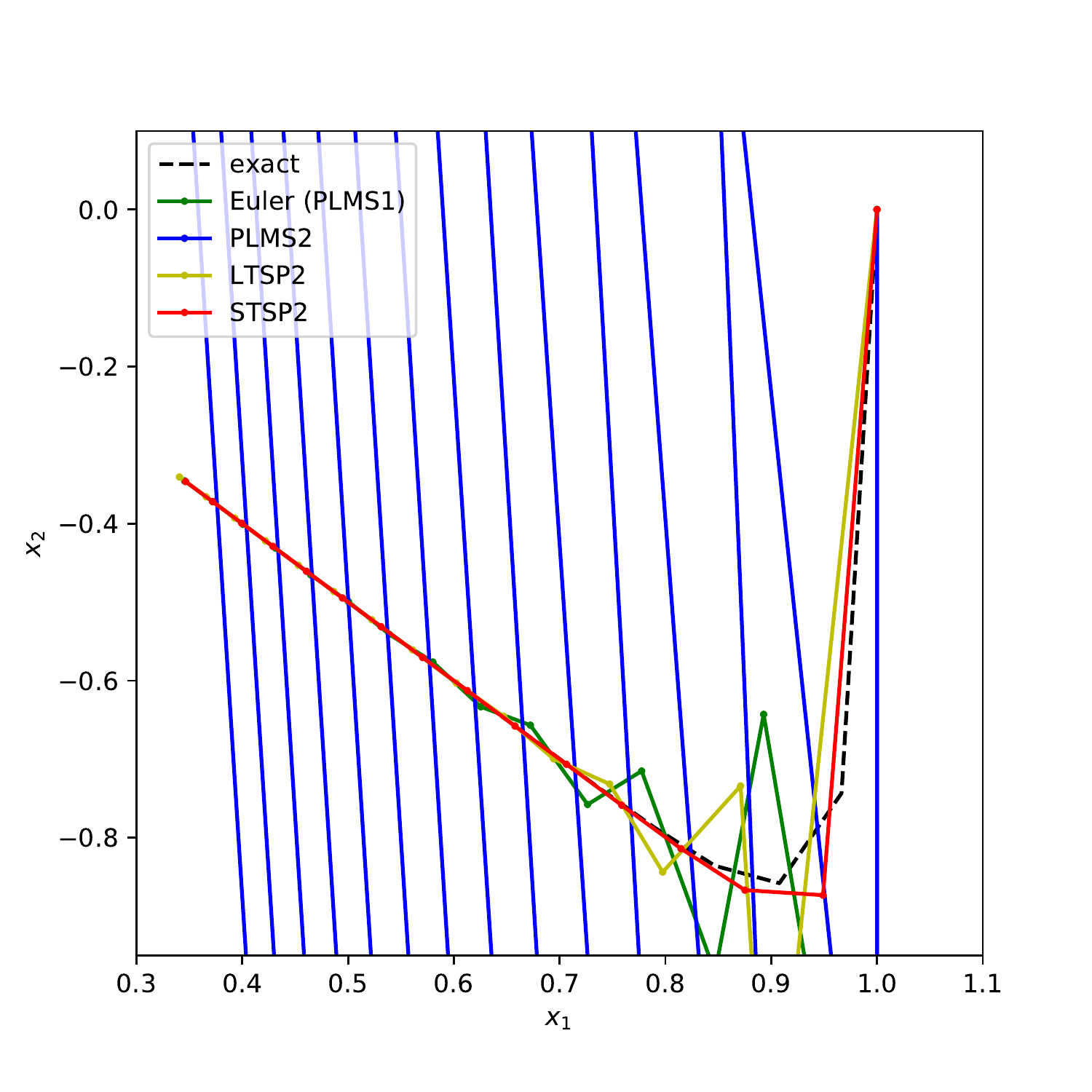}&
        \includegraphics[width=0.30\textwidth]{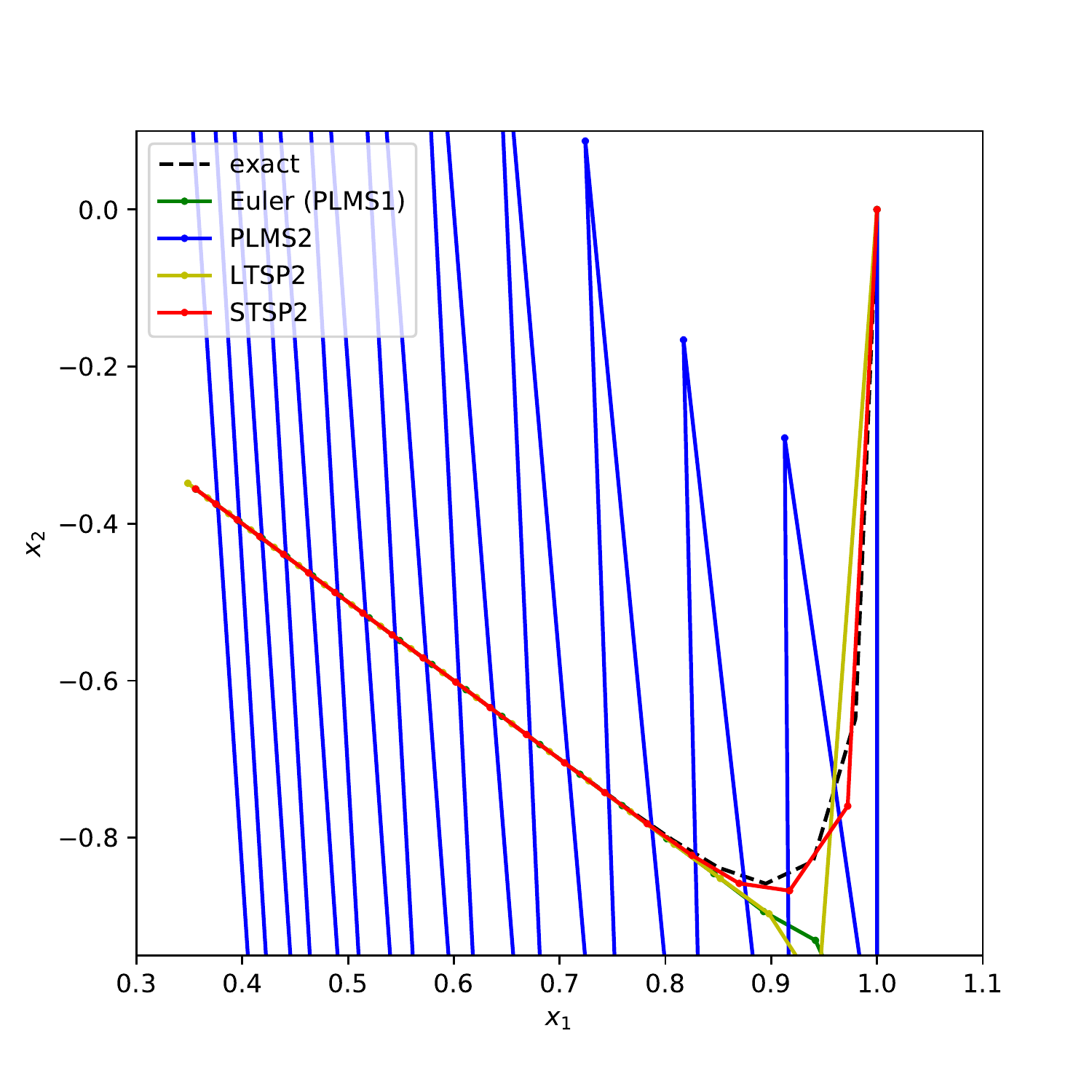}
        \\
        & 10 steps &  15 steps &   20 steps \\
    \end{tabular}
        \caption{This figure show how numerical solutions look like when the number of steps are close to the number in Table \ref{tab:theory}}.
    \label{fig:toy2}
\end{figure}

In Table \ref{tab:theory}, we compare the lowest number of steps $N$ before each method is guaranteed to diverge from our analysis. We also show numerical solutions of our toy example in Figure \ref{fig:toy2} to compare to our analysis. It is important to note that if the number of steps exceeds Table \ref{tab:theory}, we cannot presume that the numerical solution will function properly. 


\begin{table}
    \centering
      \begin{tabular}{l ccc ccc cc}
       \toprule
           & $s=5$  & $s=10$  & $s=15$ & $s=20$ & $s=30$ & $s=40$ & $s=60$ & $s=80$\\ 
       \midrule[0.08em]
       Euler       & 4  & 6  & 9  & 11 & 16 & 21 & 31 & 41\\
       PLMS2       & 6  & 11 & 16 & 22 & 32 & 42 & 63 & 83\\
       \textbf{LTSP2}       & 2  & 3  & 7  & 9  & 14 & 19 & 29 & 39\\
       \textbf{STSP2}       & 2  & 3  & 4  & 5  & 8  & 10  & 15 & 20\\
       \bottomrule
    \caption{The lowest number of steps before we can guarantee by theory that each numerical method will fail to solve Equation \ref{test_eq}. Notice that LTSP2 and STSP2 are having lower number which mean they are also harder to fail when the number of step are reduced than Euler and PLMS2.}
    \label{tab:theory}
       \end{tabular}
\end{table}

\end{document}